\pdfoutput=1
\documentclass[letterpaper]{article}
\usepackage{times}
\usepackage{graphicx}
\usepackage{breakcites}
\usepackage{algorithm,algorithmic,a4wide,amssymb,natbib,multicol,enumitem,url}
\usepackage[breaklinks]{hyperref}
 \usepackage[top=3.7cm, bottom=3.7cm, left=3.7cm, right=3.7cm]{geometry} 
\makeatletter
\renewcommand\hyper@natlinkbreak[2]{#1}
\makeatother
\newcommand{\mytilde}{\raise.17ex\hbox{$\scriptstyle\mathtt{\sim}$}}
\begin{document}

\title{Deep Learning in  Neural Networks: An Overview  \\
{\small Technical Report IDSIA-03-14 / arXiv:1404.7828 v4 [cs.NE] (88 pages, 888 references)}}

\date{8 October 2014}
\author{J\"{u}rgen Schmidhuber~\\
The Swiss AI Lab IDSIA \\
Istituto Dalle Molle di Studi sull'Intelligenza  Artificiale\\
University of Lugano~\& SUPSI \\
Galleria 2, 6928 Manno-Lugano~\\
Switzerland}
\maketitle

\begin{abstract}
In recent years, deep artificial neural networks (including recurrent ones) have won numerous contests in pattern recognition and machine learning.
This historical survey compactly summarises relevant work, much of it from the previous millennium. Shallow and deep learners are distinguished by the depth of their {\em credit assignment paths}, which are chains of possibly learnable, causal links between 
actions and effects. I review deep supervised learning (also recapitulating the history of backpropagation), unsupervised learning, reinforcement learning \& evolutionary computation, and indirect search for short programs encoding deep and large networks.

\vspace{7mm}
\noindent
LATEX source: \url{http://www.idsia.ch/~juergen/DeepLearning8Oct2014.tex} \\
Complete BIBTEX file (888 kB): \url{http://www.idsia.ch/~juergen/deep.bib}

\end{abstract}


\vspace{1cm}
\subsubsection*{Preface}
\label{foreword}
This is the preprint of an invited {\em Deep Learning} (DL) overview. One of its goals is to assign credit to those who contributed to the present state of the art. I acknowledge the limitations of attempting to achieve this goal. The DL research community itself may be viewed as a continually evolving, deep network of scientists who have influenced each other in complex ways. Starting from recent DL results, I tried to trace back the origins of relevant ideas through the past half century and beyond, sometimes using ``local search" to follow citations of citations  backwards in time. Since not all DL publications properly acknowledge earlier relevant work, additional global search strategies were employed, aided by consulting numerous neural network experts. As a result, the present preprint mostly consists of references. Nevertheless, through an expert selection bias I may have missed important work. A related bias was surely introduced by my special familiarity with the work of my own DL research group in the past quarter-century. For these reasons, this work should be viewed as merely a snapshot of an ongoing credit assignment process. To help improve it, please do not hesitate to send corrections and  suggestions to {\em juergen@idsia.ch}.


\newpage
\tableofcontents

\subsubsection*{Abbreviations in Alphabetical Order}
\label{abb}
\vspace{-0.5cm}
\begin{multicols}{2}
\begin{itemize}[leftmargin=0cm,itemindent=0cm,labelwidth=\itemindent,labelsep=0cm,align=left,noitemsep,nolistsep]
\item[] AE: Autoencoder    
\item[] AI: Artificial Intelligence 
\item[] ANN: Artificial Neural Network    
\item[] BFGS: Broyden-Fletcher-Goldfarb-Shanno    
\item[] BNN: Biological Neural Network 
\item[] BM: Boltzmann Machine 
\item[] BP: Backpropagation 
\item[] BRNN: Bi-directional Recurrent Neural Network 
\item[] CAP: Credit Assignment Path 
\item[] CEC: Constant Error Carousel 
\item[] CFL: Context Free Language 
\item[] CMA-ES: Covariance Matrix Estimation ES 
\item[] CNN: Convolutional Neural Network 
\item[] CoSyNE: Co-Synaptic Neuro-Evolution 
\item[] CSL: Context Senistive Language
\item[] CTC : Connectionist Temporal Classification 
\item[] DBN: Deep Belief Network 
\item[] DCT: Discrete Cosine Transform
\item[] DL: Deep Learning 
\item[] DP: Dynamic Programming  
\item[] DS: Direct Policy Search
\item[] EA: Evolutionary Algorithm 
\item[] EM: Expectation Maximization 
\item[] ES: Evolution Strategy 
\item[] FMS: Flat Minimum Search
\item[] FNN: Feedforward Neural Network  
\item[] FSA: Finite State Automaton 
\item[] GMDH: Group Method of Data Handling  
\item[] GOFAI: Good Old-Fashioned AI  
\item[] GP: Genetic Programming 
\item[] GPU: Graphics Processing Unit 
\item[] GPU-MPCNN: GPU-Based MPCNN  
\item[] HMM: Hidden Markov Model 
\item[] HRL: Hierarchical Reinforcement Learning 
\item[] HTM:  Hierarchical Temporal Memory 
\item[] HMAX:  Hierarchical Model ``and X"
\item[] LSTM: Long Short-Term Memory (RNN) 
\item[] MDL: Minimum Description Length 
\item[] MDP:  Markov Decision Process 
\item[] MNIST: Mixed National Institute of Standards and Technology Database 
\item[] MP: Max-Pooling 
\item[] MPCNN: Max-Pooling CNN
\item[] NE: NeuroEvolution 
\item[] NEAT: NE of Augmenting Topologies 
\item[] NES: Natural Evolution Strategies 
\item[] NFQ: Neural Fitted Q-Learning 
\item[] NN: Neural Network 
\item[] OCR: Optical Character Recognition
\item[] PCC: Potential Causal Connection 
\item[] PDCC: Potential Direct Causal Connection 
\item[] PM: Predictability Minimization 
\item[] POMDP:  Partially Observable MDP 
\item[] RAAM: Recursive Auto-Associative Memory  
\item[] RBM: Restricted Boltzmann Machine 
\item[] ReLU: Rectified Linear Unit
\item[] RL: Reinforcement Learning 
\item[] RNN: Recurrent Neural Network 
\item[] R-prop: Resilient Backpropagation 
\item[] SL: Supervised Learning 
\item[] SLIM NN: Self-Delimiting  Neural Network  
\item[] SOTA: Self-Organising Tree Algorithm
\item[] SVM: Support Vector Machine 
\item[] TDNN: Time-Delay Neural Network 
\item[] TIMIT: TI/SRI/MIT Acoustic-Phonetic Continuous Speech Corpus
\item[] UL: Unsupervised Learning 
\item[] WTA:  Winner-Take-All   
\end{itemize}
\end{multicols}

\section{Introduction to Deep Learning (DL) in Neural Networks (NNs)}
\label{intro}

Which modifiable components of a learning system are responsible for its success or failure?
What changes to them improve performance? 
This has been called the {\em fundamental credit assignment problem}~\citep{Minsky:63}.
There are general credit assignment methods for {\em universal problem solvers} 
that are 
time-optimal in various theoretical senses
(Sec.~\ref{unirl}).
The present survey, however, will focus on the narrower, but now commercially important, subfield
of {\em Deep Learning} (DL) in {\em Artificial Neural Networks} (NNs).

A standard neural network (NN) consists of many simple, connected
processors called neurons, each producing a sequence of real-valued
activations.  Input neurons get activated through sensors perceiving the
environment, other neurons get activated through weighted 
connections from previously active neurons (details in Sec.~\ref{notation}).  
Some neurons may influence the environment
by triggering actions.  {\em Learning} or {\em credit assignment} is
about finding weights that make the NN exhibit {\em desired} behavior,
such as driving a car.  Depending on the problem and how the neurons
are connected, such behavior may require long causal chains of
computational stages (Sec.~\ref{caps}), where each stage transforms
(often in a non-linear way) the aggregate activation of the
network. Deep Learning is about accurately assigning credit across
{\em many} such stages.

{\em Shallow} NN-like models with {\em few} such stages have been around for many decades 
if not centuries  (Sec.~\ref{1940}).
Models with several successive nonlinear layers of  neurons date back at least 
to the 1960s (Sec.~\ref{1965}) and  1970s (Sec.~\ref{1970}).
An efficient gradient descent method for teacher-based {\em Supervised Learning} (SL) in discrete,
differentiable networks 
of arbitrary depth 
called {\em  backpropagation} (BP)
was developed in the 1960s and 1970s, and applied to NNs in 1981 (Sec.~\ref{1970}).  
BP-based 
training of {\em deep} NNs with {\em many} layers, however, 
had been found to be difficult in practice by the late 1980s (Sec.~\ref{1990}), and had
become an explicit research subject 
by the early 1990s (Sec.~\ref{1991a}). 
DL became 
practically feasible to some extent through the help of
{\em Unsupervised Learning} (UL), e.g., Sec.~\ref{1991b} (1991), Sec.~\ref{2006} (2006).
The 1990s and 2000s also saw many improvements of 
purely supervised DL (Sec.~\ref{super}). 
In the new millennium, deep NNs have finally attracted wide-spread attention,
mainly by outperforming alternative machine learning methods
such as kernel machines~\citep{Vapnik:95,advkernel}
in numerous important applications.
In fact, since 2009,
supervised deep NNs
have won many
official international pattern recognition competitions (e.g.,
Sec.~\ref{2009},~\ref{2011},~\ref{2012},~\ref{2013}),  
achieving the first superhuman visual pattern recognition results in
limited domains (Sec.~\ref{2011}, 2011).
Deep NNs also have become relevant for the more general field of 
{\em Reinforcement Learning} (RL) where there is no supervising teacher (Sec.~\ref{deeprl}).

Both {\em feedforward} (acyclic) NNs (FNNs)
and {\em recurrent} (cyclic) NNs (RNNs) have won contests 
(Sec. \ref{1994}, \ref{2003}, \ref{2009}, \ref{2011}, \ref{2012}, \ref{2013}). 
In a sense, RNNs are the deepest of all NNs (Sec.~\ref{caps})---they are general computers
more powerful than FNNs, 
and can in principle create and process
 memories of arbitrary sequences of input patterns~\citep[e.g.,][]{siegelmann91turing,schmidhuber1990}. 
Unlike traditional methods for automatic sequential program synthesis~\citep[e.g.,][]{waldinger69,balzer1985,soloway1986,Deville:94}, RNNs can learn programs that mix sequential and parallel information processing in a natural and efficient way, exploiting the massive parallelism viewed as crucial for sustaining the rapid decline of computation cost observed over the past 75 years. 

The rest of this paper is
structured as follows.
Sec.~\ref{notation} introduces
a compact, event-oriented notation that is simple yet general enough to accommodate both
FNNs and RNNs. 
Sec.~\ref{caps} introduces the concept of {\em Credit Assignment Paths} (CAPs) to measure whether learning in a given  NN application is of the {\em deep} or {\em shallow} type.
Sec.~\ref{themes} lists recurring themes of DL in SL, UL, and RL. 
Sec.~\ref{super}  focuses on SL and UL, 
and on how UL can facilitate SL, although pure SL
has become dominant in recent competitions  
(Sec.~\ref{2009}--\ref{dominant}).
Sec.~\ref{super} is arranged in a
historical timeline format with 
subsections on
important inspirations and technical contributions.
Sec.~\ref{deeprl} on deep RL discusses traditional
{\em Dynamic Programming} (DP)-based RL 
combined with gradient-based search techniques for SL or UL in deep NNs, 
as well as general methods for direct and indirect search in the weight space of deep 
FNNs and RNNs, 
including successful policy gradient and evolutionary methods.

\section{Event-Oriented Notation for Activation Spreading in NNs}
\label{notation}

Throughout this paper, let $i,j,k,t,p,q,r$ denote positive integer variables
assuming ranges implicit in the given contexts. 
Let $n,m,T$ denote positive integer constants.

An NN's topology may change over time (e.g., Sec.~\ref{1965},~\ref{mdlnn}).
At any given moment, 
it can be described as a finite subset of units (or nodes or neurons)  $N=\{u_1,u_2, \ldots, \}$ and a finite set 
$H \subseteq N \times N$ of directed edges or connections between nodes.
FNNs are acyclic graphs, RNNs cyclic. 
The first (input) layer is the set of input units, a subset of $N$.
In FNNs, the $k$-th layer ($k>1$) is the set of all nodes 
$u \in N$ such that there is an edge path of length $k-1$ (but no longer path) between some input unit and $u$.
There may be shortcut connections between distant layers.
In sequence-processing, fully connected 
 RNNs, all units have connections to all non-input units.


The NN's behavior or program is determined by a set of real-valued, possibly modifiable,
parameters or weights 
$w_i$ $(i=1,\ldots,n)$.
We now focus on a single finite {\em episode} or {\em epoch} of information processing 
and activation spreading, without learning through weight changes.
The following slightly unconventional 
notation is designed to compactly describe what is happening
 {\em during the runtime} of the system.

During an episode, 
there is a {\em partially causal sequence}  
$x_t (t=1,\ldots,T)$ of real values that I call events.
Each $x_t$ is either an input set by the environment, 
or the activation of a unit 
that may directly depend on other $x_k (k<t)$ through a current 
NN topology-dependent 
set $in_t$ of indices $k$ representing incoming causal connections or links.
Let the function $v$ encode topology information and map such event index pairs $(k,t)$ to weight indices.
For example, in the non-input case we may have 
$x_t=f_t(net_t)$ 
with real-valued 
$net_t=\sum_{k \in in_t} x_k w_{v(k,t)}$ (additive case)
or $net_t=\prod_{k \in in_t} x_k w_{v(k,t)}$ (multiplicative case),
where $f_t$ is a typically nonlinear real-valued {\em activation function}
such as $tanh$.
In many recent competition-winning NNs 
(Sec.~\ref{2011}, \ref{2012}, \ref{2013})
there also are  events of the type $x_t=max_{k \in in_t}(x_k)$;
some network types may also use complex polynomial activation functions (Sec.~\ref{1965}).
$x_t$ may directly affect certain  $x_k (k>t)$ through outgoing connections or links 
represented through a current 
set $out_t$ of indices $k$ with $t \in in_k$.
Some of the non-input events are called {\em output events}.

Note that many of the $x_t$ may refer to different, time-varying activations of the {\em same} unit
in sequence-processing RNNs~\citep[e.g.,][{\em ``unfolding in time"}]{Williams:89},
or also in FNNs sequentially exposed to time-varying input patterns of a large training set 
encoded as input events.
During an episode, the same weight may get reused over and over again
in topology-dependent ways, e.g., in RNNs, or in convolutional NNs
(Sec.~\ref{1979},~\ref{1989}).
I call this weight sharing {\em across space and/or time}.
Weight sharing may greatly reduce the NN's descriptive complexity, which is the number of bits of information 
required to describe the NN (Sec.~\ref{mdl}).

In {\em Supervised Learning} (SL), 
certain NN output events $x_t$  may be associated with teacher-given, real-valued labels or targets $d_t$ 
yielding errors $e_t$, e.g., $e_t=1/2(x_t-d_t)^2$.
A typical goal of supervised NN training is to find weights that 
yield episodes with small total error $E$, 
the sum of all such $e_t$. 
The hope is that the NN will generalize well in later episodes,
causing only small errors on previously unseen sequences of input events. 
Many alternative error functions for SL and UL are possible.

SL assumes that input events are independent of earlier output events (which may affect
the environment  through actions causing subsequent perceptions).
This assumption does not hold 
in the broader fields of {\em Sequential Decision Making} and  {\em Reinforcement Learning} (RL)~\citep{Kaelbling:96,Sutton:98,Hutter:05book+,wiering2012} (Sec.~\ref{deeprl}).
In RL, some of the input events may encode real-valued reward signals given by the environment, 
and a typical goal is to find weights that yield episodes with a high sum of reward signals,
through sequences of appropriate output actions.

Sec.~\ref{1970} will use the notation above to compactly 
describe a central algorithm of DL, namely,
backpropagation (BP) for supervised weight-sharing FNNs and RNNs.
(FNNs may be viewed as RNNs with certain fixed zero weights.)
Sec.~\ref{deeprl} will address the more general RL case.

\section{Depth of Credit Assignment Paths (CAPs) and of Problems}
\label{caps}

To measure whether credit assignment in a given 
NN application is of the {\em deep} or {\em shallow} type,
I introduce the concept of {\em Credit Assignment Paths} or CAPs,
which are chains of possibly causal links between the events of Sec.~\ref{notation},
e.g., from input through hidden to output layers in FNNs, or through 
transformations over time in RNNs. 

Let us first focus on SL.
Consider two events 
$x_p$ and $x_q$ $(1 \leq p < q \leq T)$.
Depending on the application, they may have a
{\em Potential Direct Causal Connection} (PDCC) expressed by the Boolean
predicate $pdcc(p,q)$, which
is true if and only if $p \in in_q$.
Then the 2-element list $(p,q)$ is defined to be a CAP (a minimal one) from $p$ to $q$.
A learning algorithm may be allowed to change  $w_{v(p,q)}$ to improve performance
in future episodes.

More general, possibly indirect,
{\em Potential Causal Connections} (PCC) are expressed by the 
recursively defined Boolean
predicate $pcc(p,q)$, which in the SL case 
is true only if $pdcc(p,q)$, or if
$pcc(p,k)$ for some $k$ and $pdcc(k,q)$.
In the latter case,
appending $q$ to any CAP from $p$ to $k$ yields a CAP from $p$ to $q$
 (this is a recursive definition, too).
The set of such CAPs may be large but is finite. 
Note that the {\em same} weight may affect {\em many} different PDCCs
between successive events listed by a given CAP, 
e.g., in the case of RNNs, or weight-sharing FNNs.

Suppose a CAP has the form $(\ldots,k,t,\ldots,q)$, where
$k$ and $t$ (possibly $t=q$) are the first successive elements 
with {\em modifiable} $w_{v(k,t)}$. Then the 
length of the suffix list $(t,\ldots, q)$ is called the CAP's {\em depth}
(which is 0 if there are no modifiable links at all).
This depth limits how far backwards credit assignment 
can move down the causal chain 
to find a modifiable 
weight.\footnote{An alternative would be to count only {\em modifiable} links when measuring depth. In many 
typical NN applications this would not make a difference, but in some it would, e.g., Sec.~\ref{worrl}.}

Suppose an episode and its event sequence $x_1,\ldots,x_T$ satisfy a computable criterion 
used to decide whether a given problem has been solved 
(e.g., total error $E$ below some threshold). 
Then the set of used weights is called a {\em solution} to the problem, 
and the depth of the deepest CAP within the sequence is called the {\em solution depth}. 
There may be other solutions (yielding different event sequences) with different depths.
Given some fixed NN topology,
the  smallest depth of any solution is called the {\em problem depth}.

Sometimes we also speak of the {\em depth of an architecture}:
SL FNNs with fixed topology imply a problem-independent maximal problem depth bounded by
 the number of non-input layers.
Certain SL RNNs
with fixed weights for all 
connections except those to output units~\citep{Jaeger2001a,maass2002,Jaeger:04,schrauwen2007} 
have
a maximal problem depth of 1,
because only the final links in the corresponding CAPs are modifiable. 
In general, however, RNNs 
may learn to solve problems of potentially unlimited depth.

Note that the definitions above are 
solely based on the depths of causal chains, and
agnostic to the temporal distance between events.
For example, {\em shallow} FNNs perceiving large ``time windows" of input events may 
correctly classify {\em long} input sequences through appropriate output events, and thus 
solve {\em shallow}
problems involving {\em long} time lags between relevant events.

At which problem depth does {\em Shallow Learning} end, and {\em Deep Learning} begin?
Discussions with DL
experts have not yet yielded a conclusive response to this question. Instead of committing myself to a precise 
answer, let me just define for the purposes of this overview:
problems of depth $>10$ require 
{\em Very Deep Learning}.

The {\em difficulty} of a problem may have little to do with its depth. 
Some NNs can quickly learn to solve certain deep problems,
e.g., through random weight guessing (Sec.~\ref{1991a})
or other types of direct search (Sec.~\ref{evorl}) or indirect search (Sec.~\ref{comrl}) in weight space,
or through training an NN first on shallow problems whose solutions may then generalize to deep problems,
or through collapsing sequences of (non)linear operations into a single (non)linear 
operation~\citep[but see an analysis of non-trivial aspects of deep linear networks,][Section B]{baldihornik95}.
In general, however, finding an NN that precisely models a given training set is an 
NP-complete problem~\citep{judd1990,blum1992},
also in the case of deep NNs~\citep{sima1994,souto1999,windisch2005};
compare a survey of negative results~\citep[Section 1]{sima2002}.

Above we have focused on SL.
In the more general case of RL in unknown environments, 
$pcc(p,q)$ is also true if $x_p$ is an output event and $x_q$ 
any later input event---any action may affect the environment and thus any later perception.
(In the real world, the environment may even influence {\em non-input} events 
computed on a physical hardware entangled with the entire universe, 
but this is ignored here.) 
It is possible to model and replace
such unmodifiable {\em environmental} PCCs
through a part of the NN that has already learned to predict (through some of its units) 
 input events (including reward signals) from
former input events and actions (Sec.~\ref{worrl}). Its weights are frozen,
but can help to assign credit to other, still modifiable weights used to compute actions (Sec.~\ref{worrl}).
This approach may lead to very deep CAPs though.

Some DL research is about automatically rephrasing problems such that their
 depth is reduced (Sec.~\ref{themes}).
In particular,
sometimes UL is used to make SL problems less deep, e.g., Sec.~\ref{1991b}.
Often {\em Dynamic Programming} (Sec.~\ref{dp}) is used to facilitate certain traditional
RL problems, e.g., Sec.~\ref{trarl}.
Sec.~\ref{super} focuses on CAPs for
SL, Sec.~\ref{deeprl} on the more complex case of RL.

\section{Recurring Themes of Deep Learning}
\label{themes}

\subsection{Dynamic Programming for Supervised/Reinforcement Learning (SL/RL)}
\label{dp}

\begin{sloppypar}
One recurring theme of DL is
{\em Dynamic Programming} (DP)~\citep{Bellman:1957},
which  can help to facilitate credit assignment
under certain assumptions. For example, 
in SL NNs, backpropagation itself can be viewed as a DP-derived method (Sec.~\ref{1970}).
In traditional RL based on strong Markovian assumptions,
DP-derived methods can help to greatly reduce problem depth (Sec.~\ref{trarl}). 
DP algorithms are also essential for systems that combine concepts of NNs and
graphical models, such as {\em Hidden Markov
Models} (HMMs)~\citep{stratonovich1960,baum1966}
and {\em Expectation Maximization} (EM)~\citep{dempster77,friedman2001}, e.g., \citep{bottou91,bengio91,bourlard+morgan:1994,baldichauvin96,jordan2001,bishop:2006,hastie2009,domingos2011,dahl2012,speech2012,diwu2014}.

\subsection{Unsupervised Learning (UL) Facilitating SL and RL}
\label{ul}

Another recurring theme is how 
UL 
can facilitate both SL (Sec.~\ref{super}) and RL (Sec.~\ref{deeprl}).
UL (Sec.~\ref{ulnn}) is normally used to 
encode raw incoming data such as video or speech streams
in a form that is more convenient for subsequent goal-directed learning.
In particular, codes that describe the original data in a less redundant or more compact way
can be fed into SL (Sec.~\ref{1991b},~\ref{2006})
or RL machines (Sec.~\ref{unsrl}), whose
search spaces may thus become smaller 
(and whose CAPs shallower)
than those necessary for dealing with the raw data.
UL is closely 
connected to the topics of 
{\em regularization} 
and compression (Sec.~\ref{mdl},~\ref{mdlnn}).

\subsection{Learning Hierarchical Representations Through Deep SL, UL, RL}
\label{hie}

Many methods of 
{\em  Good Old-Fashioned Artificial Intelligence} (GOFAI)~\citep{Nilsson:80} 
as well as more recent approaches to AI~\citep{russell1995} and {\em Machine Learning}~\citep{Mitchell:97} 
learn hierarchies of more and more abstract data representations.
For example, certain methods of syntactic pattern recognition~\citep{Fu:77} such as
{\em grammar induction} discover hierarchies of  
formal rules to model observations.
The partially (un)supervised
{\em Automated Mathematician / EURISKO}~\citep{Lenat:83,Lenat:84} continually learns   
concepts by combining previously learnt concepts.
Such hierarchical representation learning~\citep{Ring:94,bengio2013tpami,lideng2014} is also a recurring theme of
DL NNs for SL (Sec.~\ref{super}), 
UL-aided SL (Sec.~\ref{1987},~\ref{1991b},~\ref{2006}), 
and hierarchical RL (Sec.~\ref{subrl}).
Often, abstract hierarchical representations are natural by-products of
data compression (Sec.~\ref{mdl}), e.g., Sec.~\ref{1991b}.

\subsection{Occam's Razor: Compression and Minimum Description Length (MDL)}
\label{mdl}

Occam's razor favors simple solutions over complex ones.
Given some programming language,
the principle of {\em Minimum Description Length} (MDL) can be used 
to measure the complexity of a solution candidate by
the length of the shortest program that 
computes it~\citep[e.g.,][]{Solomonoff:64,Kolmogorov:65,Chaitin:66,Wallace:68,Levin:73a,Solomonoff:78,Rissanen:86,Blumer:87,LiVitanyi:97,gruenwald2005}.
Some methods explicitly take into account program runtime~\citep{Allender:92,Watanabe:92,Schmidhuber:97nn+,Schmidhuber:02colt}; 
many consider only programs with constant runtime, written
in non-universal programming languages~\citep[e.g.,][]{Rissanen:86,Hinton:93}.
In the NN case, 
the MDL principle suggests that low NN weight complexity
corresponds to high NN probability 
in the  Bayesian view~\citep[e.g.,][]{MacKay:92b,Buntine:91,neal1995,freitas2003},
and to high generalization performance~\citep[e.g.,][]{BaumHaussler:89}, 
without overfitting the training data.
Many methods have been proposed for {\em regularizing} NNs, that is, 
searching for solution-computing but simple, low-complexity SL NNs (Sec.~\ref{mdlnn}) 
and RL NNs (Sec.~\ref{comrl}).
This is closely 
related to certain UL methods (Sec.~\ref{ul},~\ref{ulnn}).

\subsection{Fast Graphics Processing Units (GPUs) for DL in NNs}
\label{gpu}

While the previous millennium saw several attempts at creating fast NN-specific hardware~\citep[e.g.,][]{jackel-90,faggin92,ramacher93,widrow94,heemskerk1995,cbm97,urlbe1999},
and at exploiting standard hardware~\citep[e.g.,][]{anguita1994,muller1995,anguita1996},
the new millennium brought a DL breakthrough in form of cheap, multi-processor
graphics cards or GPUs. GPUs are widely used for video games, a huge and competitive market
that has driven down hardware prices.
GPUs excel at the fast matrix and vector multiplications required not only for 
convincing virtual realities but also for NN training, 
where they can speed up learning by a factor of  50 and more.
Some of the GPU-based FNN implementations (Sec.~\ref{2007}--\ref{2011}) have greatly contributed to recent successes in contests for pattern recognition (Sec.~\ref{2011}--\ref{2013}),
image segmentation (Sec.~\ref{2012}),
and object detection (Sec.~\ref{2012}--\ref{2013}).

\section{Supervised NNs, Some Helped by Unsupervised NNs}
\label{super}

The main focus of current practical applications is on {\em Supervised Learning} (SL),
which has dominated recent pattern recognition contests 
(Sec.~\ref{2009}--\ref{dominant}).
Several methods, however, use additional 
{\em Unsupervised Learning} (UL) to facilitate SL (Sec.~\ref{1987},~\ref{1991b},~\ref{2006}).
It does make sense to treat SL and UL in the same section:
often gradient-based methods, such as BP (Sec.~\ref{bpsec}),
are used to optimize objective functions of both UL and SL, 
and the boundary 
between SL and UL may blur, for example,
when it comes to time series prediction and sequence
classification, e.g., Sec.~\ref{1991b},~\ref{1994}.

A historical timeline format will help to arrange 
subsections on
important inspirations and technical contributions
(although such a subsection may span a time interval of many years).
Sec.~\ref{1940} briefly mentions early, shallow NN models since the 1940s (and 1800s),
Sec.~\ref{1962} additional early neurobiological inspiration relevant for modern Deep Learning (DL).
Sec.~\ref{1965} is about GMDH networks  (since 1965), 
to my knowledge the first (feedforward) DL systems.
Sec.~\ref{1979} is about the relatively deep {\em Neocognitron} NN (1979)
which is very similar to certain modern deep FNN architectures, as it
combines convolutional NNs (CNNs), weight pattern replication, and subsampling mechanisms.
Sec.~\ref{1970} uses the notation of Sec.~\ref{notation} to compactly 
describe a central algorithm of DL, namely,
{\em backpropagation} (BP) for 
supervised weight-sharing FNNs and RNNs. It also summarizes
 the history of BP 1960-1981 and beyond.
Sec.~\ref{1990} describes problems encountered in the late 1980s with BP for deep NNs,
and mentions several ideas from the previous millennium  to overcome them.
Sec.~\ref{1987} discusses a first hierarchical stack (1987) of coupled UL-based Autoencoders (AEs)---this concept
resurfaced in the
new millennium (Sec.~\ref{2006}). 
Sec.~\ref{1989} is about applying BP to CNNs (1989), which is important for today's DL applications.
Sec.~\ref{1991a} explains BP's {\em Fundamental DL Problem} (of vanishing/exploding gradients)
discovered in 1991.
Sec.~\ref{1991b} explains how a deep RNN stack of 1991 (the {\em History Compressor}) pre-trained by UL helped to solve previously unlearnable DL benchmarks 
requiring {\em Credit Assignment Paths} (CAPs, Sec.~\ref{caps}) of depth 1000 and more.
Sec.~\ref{1999} discusses a particular {\em winner-take-all} (WTA) method called {\em Max-Pooling} (MP, 1992)  widely used in today's deep FNNs.
Sec.~\ref{1994} mentions a first important contest won by SL NNs in 1994.
Sec.~\ref{1997} describes a purely supervised DL RNN ({\em Long Short-Term Memory}, LSTM, 1995) for problems of depth 1000 and more.
Sec.~\ref{2003} mentions an early contest of 2003 
won by an ensemble of shallow FNNs, as well as
good pattern recognition results with CNNs and deep FNNs and LSTM RNNs (2003). 
Sec.~\ref{2006} is mostly about {\em Deep Belief Networks} (DBNs, 2006) and related stacks of {\em  Autoencoders} (AEs, Sec.~\ref{1987}), both pre-trained by UL to facilitate subsequent BP-based SL (compare Sec.~\ref{timelags},~\ref{1991b}). 
Sec.~\ref{2007} mentions the first SL-based  GPU-CNNs (2006),  BP-trained MPCNNs (2007),
and LSTM stacks (2007).
Sec.~\ref{2009}--\ref{2013} focus on official competitions with secret test sets 
won by (mostly purely supervised) deep NNs since 2009,
in sequence recognition, image classification, image segmentation, and object detection.
Many RNN results depended on LSTM (Sec.~\ref{1997});
many FNN results depended on GPU-based FNN code developed since 2004 (Sec.~\ref{2007},~\ref{2009},~\ref{2010},~\ref{2011}),
in particular, GPU-MPCNNs  (Sec.~\ref{2011}). 
Sec.~\ref{tricks} mentions
recent tricks for improving DL in NNs, many of them closely
related to earlier tricks from the previous millennium (e.g., Sec.~\ref{betterbp},~\ref{mdlnn}).
Sec.~\ref{bnn} discusses how artificial NNs can  help to understand biological NNs;
Sec.~\ref{spiking} addresses the possibility of DL in NNs with spiking neurons.

\subsection{Early NNs Since the 1940s (and the 1800s)}
\label{1940}

Early NN architectures~\citep{mcculloch:43} did not learn.
The first ideas about UL were published a few years later~\citep{Hebb:49}.
The following decades brought simple NNs trained by SL~\citep[e.g.,][]{rosenblatt1958,Rosenblatt:62,adaline62,Narendra:74}
and UL~\citep[e.g.,][]{Grossberg69a,kohonen1972,malsburg1973,WillshawMalsburg:76},
as well as closely related associative memories~\citep[e.g.,][]{Palm:80,Hopfield:82}.

In a sense NNs have been around even longer, since
early supervised NNs  were essentially variants of 
linear regression methods going back at least to the early 1800s~\citep[e.g.,][]{legendre1805,gauss1809,gauss1821}; Gauss also refers to his work of 1795.
Early NNs had a maximal CAP depth of 1 (Sec.~\ref{caps}).


\subsection{Around 1960: Visual Cortex Provides Inspiration for DL (Sec.~\ref{1979},~\ref{1999})}
\label{1962}

Simple cells and complex cells were found in the cat's 
visual cortex~\citep[e.g.,][]{Hubel:62,wiesel:1959}.
These cells fire in response to certain properties of visual sensory inputs, 
such as the orientation of edges. Complex cells exhibit more spatial invariance than simple cells.
This  inspired later deep NN architectures 
(Sec.~\ref{1979},~\ref{1999}) used in certain modern award-winning Deep Learners (Sec.~\ref{2011}--\ref{2013}).

\subsection{1965: Deep Networks Based on the Group Method of Data Handling}
\label{1965}

Networks trained by the {\em Group Method of Data Handling} (GMDH)~\citep{ivakhnenko1965,ivakhnenko1967,ivakhnenko1968,ivakhnenko1971} 
were perhaps the first DL systems of
the {\em Feedforward Multilayer Perceptron} type,
although there was earlier work on NNs with a single hidden layer~\citep[e.g.,][]{joseph1961,viglione1970}.
The units of GMDH nets may have polynomial activation functions implementing 
{\em Kol\-mo\-go\-rov-Gabor polynomials} (more general than other widely used NN activation functions, Sec.~\ref{notation}).
Given a training set, layers are incrementally grown and trained by regression analysis ~\citep[e.g.,][]{legendre1805,gauss1809,gauss1821} (Sec.~\ref{1940}), 
then pruned with the help of a
separate {\em validation set} (using today's terminology), where
{\em Decision Regularisation} is used to weed out 
superfluous units (compare Sec.~\ref{mdlnn}). The numbers of layers and units per layer can be learned in
problem-dependent fashion.
To my knowledge, this was the first example of open-ended, hierarchical 
representation learning  in NNs (Sec.~\ref{hie}).
A paper of 1971 already described a deep GMDH network with 8 layers
\citep{ivakhnenko1971}.
There have been numerous applications of GMDH-style nets, e.g.~\citep{ikeda1976,farlow1984,madala1994,ivakhnenko1995,kondo1998,kordik2003,witczak2006,kondo2008}.

\subsection{1979:  Convolution $+$ Weight Replication $+$ Subsampling (Neocognitron)}
\label{1979}

Apart from deep GMDH networks (Sec.~\ref{1965}),
the {\em Neocognitron}~\citep{Fukushima:1979neocognitron,fukushima:1980,Fukushima:2013}
was perhaps the first artificial NN that deserved the attribute {\em deep}, and the first 
to incorporate the  neurophysiological insights of Sec.~\ref{1962}. 
It introduced {\em convolutional NNs} (today often called CNNs or convnets), where the
(typically rectangular) receptive field of a {\em convolutional unit} with given weight vector (a {\em filter})
is shifted step by step across a 2-dimensional array of input values, such as the pixels of an image (usually there are several such filters). 
The resulting 2D array of subsequent activation events of this unit can then provide inputs to higher-level units, and so on.
Due to massive {\em weight replication} (Sec.~\ref{notation}),  
relatively few parameters (Sec.~\ref{mdl}) 
may be necessary to describe the behavior of such a {\em convolutional layer}.

{\em Subsampling} or {\em downsampling} layers consist of units whose fixed-weight connections originate from physical neighbours in the convolutional layers below. 
Subsampling units become active if at least one of their inputs is active;
their responses are insensitive to certain small image shifts (compare Sec.~\ref{1962}).


The Neocognitron is
very similar to the architecture of modern, contest-winning, purely {\em  supervised}, 
feedforward, gradient-based Deep Learners with alternating convolutional and downsampling layers
(e.g., Sec.~\ref{2011}--\ref{2013}).
Fukushima, however, did not set the weights by supervised
backpropagation (Sec.~\ref{1970},~\ref{1989}),
but by local, WTA-based
{\em un}supervised learning rules~\citep[e.g.,][]{Fukushima:2013b}, or by pre-wiring. 
In that sense he did not care for the  
DL problem (Sec.~\ref{1991a}), 
although his architecture was comparatively deep indeed. For downsampling purposes
he used {\em  Spatial Averaging}~\citep{fukushima:1980,Fukushima:2011} instead of {\em Max-Pooling} (MP, Sec.~\ref{1999}),
currently a particularly convenient and popular WTA mechanism. 
Today's DL combinations of CNNs and MP and BP also profit a lot from
later work (e.g., Sec.~\ref{1989},~\ref{2007},~\ref{2007},~\ref{2011}).

\subsection{1960-1981 and Beyond: Development of Backpropagation (BP) for NNs}
\label{1970}
The minimisation of  
errors through {\em gradient descent}~\citep{hadamard1908memoire} in
the parameter space of complex, 
nonlinear, differentiable~\citep{leibniz1684}, multi-stage, NN-related systems has been discussed 
at least since the early 1960s~\citep[e.g.,][]{Kelley:1960,bryson:1961,BRYSON-DENHAM-61A,PONTRYAGIN61A,dreyfus:1962,Wilkinson:1965,Amari:1967:TAP,bryson1969applied,director:1969},
initially within the framework of Euler-LaGrange equations in the {\em Calculus of Variations}~\citep[e.g.,][]{Euler:1744}.

{\em Steepest descent} in the weight space of 
such systems can be performed~\citep{bryson:1961,Kelley:1960,bryson1969applied} 
by iterating the chain rule~\citep{leibniz:1676,de1716analyse} 
\`{a} la {\em Dynamic Programming} (DP)~\citep{Bellman:1957}.
A simplified derivation of this backpropagation method uses the chain rule only~\citep{dreyfus:1962}.

The systems of the 1960s were already efficient in the DP sense.
However, they backpropagated derivative information through
standard Jacobian matrix calculations from one ``layer" to the previous one,
without explicitly addressing either direct links across several layers or potential additional efficiency gains due to network sparsity 
(but perhaps such enhancements seemed obvious to the authors).
Given all the prior work on learning in multilayer NN-like systems (see also Sec.~\ref{1965}
on deep nonlinear nets since 1965),
it seems surprising in hindsight that a book~\citep{MinskyPapert:69} 
on the limitations of simple linear 
perceptrons with a single layer (Sec.~\ref{1940})
discouraged some researchers from further studying NNs.

Explicit, efficient error backpropagation (BP) in arbitrary, discrete, possibly sparsely connected, 
NN-like networks apparently was first described 
 in a 1970 master's thesis~\citep{Linnainmaa:1970,Linnainmaa:1976}, albeit without reference to NNs.
BP is also known as the reverse mode of automatic differentiation~\citep{Griewank:2012}, 
where the costs of forward activation spreading essentially equal the costs of backward 
derivative calculation. 
See early FORTRAN code~\citep{Linnainmaa:1970} and closely related work~\citep{ostrovskii:1971}.

Efficient BP was soon explicitly used to minimize cost functions by
adapting control parameters (weights)~\citep{dreyfus:1973}.
Compare some preliminary, NN-specific discussion~\citep[section 5.5.1]{Werbos:74}, 
a method for multilayer threshold NNs~\citep{bobrowski78},
and a computer program for automatically deriving and implementing BP 
for given differentiable systems~\citep{SPEELPENNING80A}.

To my knowledge, the first NN-specific application of
efficient BP as above was described in 1981~\citep{Werbos:81sensitivity,werbos2006backwards}.
Related work was published several  years later~\citep{Parker:85,LeCun:85,lecun-88}.
A paper of 1986 significantly contributed to the popularisation of BP for NNs~\citep{Rumelhart:86}, experimentally demonstrating the emergence of useful 
internal representations in hidden layers. 
See generalisations for sequence-processing 
recurrent NNs~\citep[e.g.,][]{Williams:89,RobinsonFallside:87tr,Werbos:88gasmarket,WilliamsZipser:88,WilliamsZipser:89nc,WilliamsZipser:89cs,Rohwer:89,Pearlmutter:89,Gherrity:89,WilliamsPeng:90,Schmidhuber:92ncn3,Pearlmutter:95,baldi95,kremer2001,atiya2000}, also for equilibrium RNNs~\citep{Almeida:87,Pineda:87} with stationary inputs.

\subsubsection{BP for Weight-Sharing Feedforward NNs (FNNs) and Recurrent NNs (RNNs)}
\label{bpsec}

Using the notation of Sec.~\ref{notation} for weight-sharing FNNs or RNNs, 
after an episode of activation spreading through differentiable $f_t$, 
a single iteration of gradient descent through BP computes
changes of all $w_i$ in proportion to 
$
\frac{\partial E}{\partial w_i}=
\sum_t
\frac{\partial E}{\partial net_t}
\frac{\partial net_t}{\partial w_i}
$ 
as in Algorithm \ref{bp} (for the additive case),
where each weight $w_i$ is associated with a real-valued variable $\triangle_i$ initialized by 0.

\begin{algorithm}{\bf Alg. \ref{bp}: One iteration  of BP for weight-sharing FNNs or RNNs} 
\label{bp}
\begin{algorithmic}
\FOR {$t=T,\ldots,1$}  
\STATE to compute $\frac{\partial E}{\partial net_t}$, inititalize real-valued error signal variable $\delta_t$ by 0;
\STATE if $x_t$ is an input event then continue with next iteration;
\STATE if there is an error $e_t$ then $\delta_t := x_t-d_t$;
\STATE add to $\delta_t$ the value $\sum_{k \in out_t}  w_{v(t,k)} \delta_k$;
{\em (this is the elegant and efficient recursive chain rule
application collecting impacts of $net_t$ on future events)} 
\STATE multiply  $\delta_t$ by $f'_t(net_t)$;
\STATE for all $k \in in_t$ add to  $\triangle_{w_{v(k,t)}}$ the value $x_k \delta_t$
\ENDFOR
\STATE change each $w_i$ in proportion to $\triangle_i$ and a small real-valued learning rate
\end{algorithmic}
\end{algorithm}

The computational costs of the backward (BP) pass are essentially those of the forward pass
(Sec.~\ref{notation}).
Forward and backward passes are re-iterated until sufficient performance is reached.

As of 2014, this simple BP method is still the central learning algorithm for FNNs and RNNs.
Notably,
most contest-winning NNs up to 2014 (Sec.~\ref{1994},~\ref{2003},~\ref{2009},~\ref{2011},~\ref{2012},~\ref{2013})
did {\em not} augment supervised BP by 
some sort of {\em un}supervised learning as discussed in Sec.~\ref{1987},~\ref{1991b},~\ref{2006}.

\subsection{Late 1980s-2000 and Beyond: Numerous Improvements of NNs}
\label{1990}

By the late 1980s it seemed clear that BP by itself (Sec.~\ref{1970}) was no panacea.
Most FNN applications focused on FNNs with few hidden layers.
Additional hidden layers often did not seem to offer empirical benefits.
Many practitioners found solace in a theorem~\citep{Kolmogorov:57,hecht1989,hornik1989}
stating that an NN with a single layer of enough hidden units 
can approximate any multivariate continous function
with arbitrary accuracy.

Likewise, most RNN applications 
did not require backpropagating errors far. 
Many researchers helped their RNNs by first 
training them on shallow problems (Sec.~\ref{caps})
whose solutions then generalized to deeper problems.  
In fact, some popular RNN algorithms restricted credit 
assignment to a single step backwards~\citep{elman1990,Jordan:86,jordan1997},
also in more recent studies~\citep{Jaeger2001a,maass2002,Jaeger:04}. 

Generally speaking, although BP allows for deep problems in principle,
it seemed to work only for {\em shallow} problems.  
The late 1980s and early 1990s saw a few ideas 
with a potential to overcome this problem,
which was fully understood only in 1991 (Sec.~\ref{1991a}).

\subsubsection{Ideas for Dealing with Long Time Lags and Deep CAPs}
\label{timelags}

To deal with long time lags between relevant events, 
several sequence processing methods were proposed, 
including 
{\em Focused BP} based on decay 
factors for activations of units in RNNs~\citep{Mozer:89focus,Mozer:92nips},
{\em  Time-Delay Neural Networks} (TDNNs) \citep{Lang:90} and their 
adaptive extension~\citep{Bodenhausen:91},
{\em Nonlinear AutoRegressive with eXogenous inputs} (NARX) RNNs~\citep{Lin:96},
certain hierarchical RNNs~\citep{hihi:95} (compare Sec.~\ref{1991b}, 1991),
RL economies in RNNs with WTA units and local learning rules~\citep{Schmidhuber:89cs},
and other methods~\citep[e.g.,][]{Ring:93,Ring:94,Plate:93,Vries:91,Sun:93,Bengio:94}.
However, these algorithms either worked for shallow CAPs only,
could not generalize to unseen CAP depths,
had problems with greatly varying time lags between relevant events,
needed external fine tuning of delay constants,
or suffered from other problems.
In fact, it turned out that certain simple but deep benchmark problems 
used to evaluate such methods 
are more quickly solved by {\em randomly guessing} RNN weights until a solution is found~\citep{Hochreiter:96sintra}.

While the RNN methods above were designed for DL of temporal sequences,
the {\em Neural Heat Exchanger} \citep{heat90-96} consists of two parallel {\em deep FNNs} with opposite flow directions. Input patterns enter the first FNN and are propagated ``up''. Desired outputs (targets) enter the ``opposite'' FNN and are propagated ``down''. Using a local learning rule, each layer in each net tries to be similar (in information content) to the preceding layer and to the adjacent  layer of the other net. The input entering the first net slowly ``heats up'' to become the target. The target entering the opposite net slowly ``cools down'' to become the input. The {\em Helmholtz Machine}~\citep{Dayan:95,Dayan:96} may be viewed as an unsupervised (Sec.~\ref{ulnn})
variant thereof (Peter Dayan, personal communication, 1994).

A hybrid approach~\citep{shavlik1989,towell1994} initializes a potentially deep FNN through 
a domain theory in propositional logic,
which may be acquired through explanation-based learning~\citep{mitchell1986,dejong1986,minton1989}.
The NN is then fine-tuned through BP (Sec.~\ref{1970}).
The NN's depth reflects the longest chain of reasoning in the original set of logical rules.
An extension of this approach~\citep{maclin1993,shavlik1994} initializes an RNN by 
domain knowledge expressed as a Finite State Automaton (FSA).
BP-based fine-tuning has become important for later DL systems 
pre-trained by UL, e.g., Sec.~\ref{1991b},~\ref{2006}.


\subsubsection{Better BP Through Advanced Gradient Descent (Compare Sec.~\ref{tricks})}
\label{betterbp}

Numerous improvements of steepest descent through BP (Sec.~\ref{1970}) have been proposed.
Least-squares methods (Gauss-Newton, Levenberg-Marquardt)~\citep{gauss1809,newton1687,levenberg1944,marquardt1963,schaback1992}
and quasi-Newton methods (Broyden-Fletcher-Goldfarb-Shanno, 
BFGS)~\citep{broyden1965,fletcher1963,goldfarb1970,shanno1970}
are computationally too expensive for large NNs.
Partial BFGS~\citep{Battiti:92,Saito:1997} and
conjugate gradient~\citep{HestenesStiefel:1952,Moller:93} 
as well as other methods~\citep{Solla:88,Schmidhuber:89-1,Cauwenberghs:93}
provide sometimes useful fast alternatives.
BP can be treated 
as a linear least-squares problem~\citep{Biegler:93}, where
second-order gradient information is passed back to preceding layers.

To speed up BP, {\em momentum} was introduced~\citep{Rumelhart:86},
ad-hoc constants were added to the slope of the linearized activation
function~\citep{Fahlman:88}, or the
nonlinearity of the slope was exaggerated~\citep{westsaad:96}.

Only the signs of the error derivatives are taken into account by the successful 
and widely used BP variant {\em R-prop}~\citep{rprop93}
and the robust variation {\em iRprop+}~\citep{igel:01},
which was also successfully applied to RNNs.

The local gradient can be normalized based 
on the NN architecture~\citep{Schraudolph:96}, 
through a diagonalized Hessian approach~\citep{Becker:89},
or related efficient methods~\citep{schraudolph02}.

Some algorithms for controlling BP step size
adapt a global learning rate~\citep{Lapedes:86a,Vogl:88,Battiti:89,lecun-simard-pearlmutter-93,Yu:1995},
while others compute individual learning rates for each
weight~\citep{Jacobs:88,SilvaAlmeida:1990}.
In online learning, where BP is applied after each pattern presentation,
the vario-$\eta$ algorithm~\citep{DBLP:conf/nips/NeuneierZ96} sets each weight's learning rate inversely proportional to the empirical standard deviation of its
local gradient, thus normalizing the stochastic weight fluctuations.
Compare a local online step size adaptation method for nonlinear NNs~\citep{Almeida:97}.

Many additional tricks for improving NNs have been described~\citep[e.g.,][]{orr1998neural,tricksofthetrade:2012}.
Compare Sec.~\ref{mdlnn} and recent developments mentioned in Sec.~\ref{tricks}. 

\subsubsection{Searching For Simple, Low-Complexity, Problem-Solving NNs (Sec.~\ref{tricks})}
\label{mdlnn}

Many researchers used BP-like methods to search for
``simple," low-complexity NNs (Sec.~\ref{mdl})
with high generalization capability. Most approaches 
address the {\em bias/variance dilemma}~\citep{Geman:92}
through strong prior 
assumptions. For example, 
{\em weight decay}~\citep{Hanson:89,Weigend:91,Krogh:92} 
encourages near-zero weights, by penalizing large weights. In a Bayesian
framework~\citep{bayes1763}, weight decay can be derived~\citep{Hinton:93} 
from Gaussian or Laplacian weight priors~\citep{gauss1809,laplace1774}; 
see also~\citep{Murray:93}. 
An extension of this approach
postulates that 
a distribution of networks
with many similar weights 
generated by Gaussian mixtures is ``better'' {\em a priori}~\citep{Nowlan:92}.

Often weight priors  are implicit in
additional penalty terms~\citep{MacKay:92b} or
 in methods based on {\em validation sets}
~\citep{Mosteller:68,Stone:74,Eubank:88,Hastie:90,Craven:79,Golub:79},
Akaike's information criterion and
 {\em final prediction error}~\citep{Akaike:70,akaike1973,akaike1974}, or
{\em generalized prediction error}~\citep{Moody:94a,Moody:92}.
See also~\citep{Holden:94,Wang:94,Amari:93,Wang:94,Vapnik:92a,Vapnik:92,Wolpert:94b}.
Similar priors (or biases towards simplicity) are implicit in  constructive and pruning algorithms,
e.g., layer-by-layer 
 {\em sequential network construction}~\citep[e.g.,][]{ivakhnenko1968,ivakhnenko1971,Ash:89,Moody:89,gallant1988,honavar1988,Ring:91,Fahlman:91,weng1992,honavar1993,burgess1994,fritzke94,parekh2000,utgoff2002} (see also Sec.~\ref{1965},~\ref{1999}),
{\em input pruning}~\citep{Moody:92,Refenes:94},
{\em unit pruning}~\citep[e.g.,][]{ivakhnenko1968,ivakhnenko1971,White:89,Mozer:89a,Levin:94},
{\em weight pruning}, e.g., {\em optimal brain damage}~\citep{LeCun:90a},
and {\em optimal brain surgeon}~\citep{Hassibi:93}. 

A very general but not always practical
approach for discovering low-complexity SL NNs or RL NNs searches among weight matrix-computing programs written in a universal programming language, with a bias
towards fast and short programs~\citep{Schmidhuber:97nn+} (Sec.~\ref{comrl}).

{\em Flat Minimum Search} (FMS)~\citep{Hochreiter:97nc1,Hochreiter:99nc} searches 
for a ``flat'' minimum of the error function: 
a large connected region in weight space where error is low and remains 
approximately constant, that is, few bits of information are required to describe 
low-precision weights with high variance. Compare {\em perturbation tolerance conditions}~\citep{Minai:94,Murray:93,hanson1990,Neti:92,Matsuoka:92,Bishop:93,Kerlirzin:93,Carter:90}.
An MDL-based,  Bayesian 
argument suggests that flat minima correspond to 
``simple'' NNs and low 
expected overfitting. 
Compare Sec.~\ref{ulnn} and more recent developments mentioned in Sec.~\ref{tricks}.

\subsubsection{Potential Benefits of UL for SL (Compare Sec.~\ref{1987},~\ref{1991b},~\ref{2006})}
\label{ulnn}

 
The notation of Sec.~\ref{notation} introduced teacher-given labels $d_t$. 
Many papers of the previous millennium, however, were about 
 {\em unsupervised learning (UL) without a teacher} 
\citep[e.g.,][]{Hebb:49,malsburg1973,kohonen1972,Kohonen:82,Kohonen:88,WillshawMalsburg:76,Grossberg:76a,Grossberg:76b,Watanabe:85,PearlmutterHinton:86,Barrow:87,Field:87,Oja:89,Barlow:89,Baldi:89,Sanger:89,ritter1989,RubnerSchulten:90,Foldiak:90,Ritter:90,kosko1990,Mozer:91nips,Palm:92,Atick:92,Miller:94,Saund:94,Foldiak:95,DecoParra:97}; 
see also post-2000 work~\citep[e.g.,][]{carreira2001,WisSej2002,Franzius2007a,koch2008}. 

Many UL methods are designed to 
maximize entropy-related, 
information-theoretic~\citep{boltzmann1909,Shannon:48,kullback1951} objectives
\citep[e.g.,][]{Linsker:88,Barlow:89,MacKay:90,Plumbley:91,chunker91and92,Schmidhuber:92ncfactorial,Schraudolph:93,Redlich:93a,Zemel:93,Zemel:94nips,Field:94,hinton:95,Dayan:95a,Amari:96,DecoParra:97}.

Many do this to uncover and disentangle hidden underlying sources of signals 
\citep[e.g.,][]{Jutten:91,Schuster:92,andrade1993,Molgedey:94,Comon:94,Cardoso:94,Bell:95,karhunen1995,belouchrani1997,hyvarinen2001,szabo2006,shan2007,shan2014}.

Many UL methods automatically and robustly generate distributed, sparse 
representations of input 
patterns~\citep{Foldiak:90,Hinton:97,Lewicki:98b,Hyvarinen:99,Hochreiter:99nc,falconbridge2006}
through well-known feature 
detectors~\citep[e.g.,][]{Olshausen:96,Schmidhuber:96ncedges},
such as {\em off-center-on-surround}-like structures, 
as well as orientation sensitive edge detectors
and Gabor filters~\citep{gabor1946}.
They extract simple features related to those
observed in
early visual pre-processing stages of 
biological systems~\citep[e.g.,][]{valois1982,jones1987}. 

UL can also serve to extract invariant features from different 
data items~\citep[e.g.,][]{Becker:91}
through {\em coupled NNs} observing two different inputs~\citep{siamese92and93},
also called {\em Siamese NNs}~\citep[e.g.,][]{bromley-93,hadsell-chopra-lecun-06,taylor2011,chen2011ieeetnn}.  

UL can help to encode
 input data in a form  advantageous for further processing. 
In the context of DL, 
one important goal of UL is redundancy reduction.
Ideally, given an ensemble of input  patterns, redundancy reduction
through a deep NN 
will create a {\em factorial code}
(a code with statistically independent components) 
of the ensemble~\citep{Barlow:89,Barlow:89review},
to disentangle the unknown factors of variation~\citep[compare][]{bengio2013tpami}.
Such codes may be sparse
and can be advantageous for 
(1) data compression, 
(2) speeding up subsequent BP \citep{Becker:91},
(3) trivialising  the task of subsequent naive yet optimal Bayes classifiers  
\citep{Schmidhuber:96ncedges}.

Most early UL FNNs had a single layer. 
Methods for deeper UL FNNs include 
hierarchical (Sec.~\ref{hie}) 
self-organizing Kohonen maps~\citep[e.g.,][]{koikkalainen1990,lampinen1992,versino1996,dittenbach2000,rauber2002},
hierarchical {\em Gaussian potential function} networks~\citep{lee1991},
layer-wise UL of feature hierarchies fed into SL 
classifiers~\citep{Behnke:IJCNN1999,Behnke:IJCNN2003}, 
the {\em Self-Organising Tree Algorithm} (SOTA)~\citep{herrero2001},
and nonlinear {\em Autoencoders} (AEs) 
with more than 3 (e.g., 5) layers
\citep{Kramer:91,Oja:91,DeMers:93}.
Such AE NNs~\citep{Rumelhart:86} can be trained to map input patterns to themselves,
for example, by  
compactly encoding them through activations of units of a narrow bottleneck hidden layer.
Certain nonlinear AEs suffer from certain limitations~\citep{baldijmlr12}.

{\sc Lococode}~\citep{Hochreiter:99nc} uses 
FMS (Sec.~\ref{mdlnn}) to
find low-complexity AEs with low-precision weights describable by 
few bits of information, often producing sparse or factorial codes.
{\em Predictability Minimization} (PM)
\citep{Schmidhuber:92ncfactorial} searches for factorial codes 
through nonlinear feature detectors that fight nonlinear predictors,
trying to become both as informative and as unpredictable as possible. 
PM-based UL was applied not only to FNNs but also to RNNs~\citep[e.g.,][]{schmidhuber1993,Steffi:93cmss}. 
Compare Sec.~\ref{1991b} on UL-based RNN stacks (1991),
as well as later UL RNNs~\citep[e.g.,][]{Klapper:01,steil2007}.

\subsection{1987: UL Through Autoencoder (AE) Hierarchies (Compare Sec.~\ref{2006})}
\label{1987}

Perhaps the first work to study
 potential benefits of UL-based pre-training was published in 1987.
It proposed unsupervised AE hierarchies~\citep{ballard1987modular}, 
closely related to certain 
post-2000 feedforward Deep Learners based on UL (Sec.~\ref{2006}).
The lowest-level AE NN with a single hidden layer is trained to map input patterns to themselves. Its hidden layer codes are then fed into a higher-level AE of the same type, and so on. The hope is that the codes in the hidden AE layers have properties that facilitate subsequent learning. 
In one experiment, a particular AE-specific learning algorithm (different from traditional  BP of Sec.~\ref{bp}) was used to
learn a mapping in an AE stack pre-trained by this type of UL~\citep{ballard1987modular}. This was faster than 
learning an equivalent mapping by BP through a single deeper AE without pre-training. 
On the other hand, the task did not really require a deep AE, that is, the benefits of UL were not that obvious from this experiment.
Compare an early survey~\citep{hinton1989connectionist} and the somewhat 
related {\em Recursive Auto-Associative Memory} (RAAM)~\citep{pollack1988implications,Pollack:90,Melnik2000},
originally used to encode sequential linguistic structures of arbitrary size 
through a fixed number of hidden units.
More recently, RAAMs were also used as unsupervised pre-processors
to facilitate deep credit assignment for RL \citep{Gisslen2011agi} (Sec.~\ref{unsrl}).

In principle, many UL methods (Sec.~\ref{ulnn}) could be stacked like the  
AEs above,
the  history-compressing RNNs of Sec.~\ref{1991b},
the {\em Restricted Boltzmann Machines} (RBMs) of Sec.~\ref{2006},
or hierarchical Kohonen nets (Sec.~\ref{ulnn}),
to facilitate subsequent SL.
Compare {\em Stacked Generalization}~\citep{wolpert:92stacked,ting1997},
and FNNs that profit from pre-training by {\em competitive} UL~\citep[e.g.,][]{RumelhartZipser:86}
prior to BP-based fine-tuning~\citep{maclin1995}.
See also more recent methods using UL to improve subsequent SL~\citep[e.g.,][]{Behnke:IJCNN1999,Behnke:IJCNN2003,wiskott2013}.

\subsection{1989: BP for Convolutional NNs (CNNs, Sec.~\ref{1979})}
\label{1989}
In 1989, backpropagation (Sec.~\ref{1970}) was applied~\citep{LeCun:89,LeCun:90,LeCun:98} 
to Neocognitron-like, weight-sharing,
convolutional 
neural layers (Sec.~\ref{1979}) with adaptive connections.
This combination, augmented by {\em Max-Pooling} (MP, Sec.~\ref{1999},~\ref{2007}),
and sped up on graphics cards (Sec.~\ref{2011}),  
has become an 
essential ingredient of many modern, competition-winning, 
feedforward, visual Deep Learners (Sec.~\ref{2011}--\ref{dominant}).
This work also introduced
the MNIST data set of handwritten digits~\citep{LeCun:89}, which over time has become 
perhaps the most famous benchmark of Machine Learning. 
CNNs helped to 
achieve good performance on MNIST~\citep{LeCun:90} (CAP depth 5)
and on  fingerprint recognition~\citep{baldi93finger};
similar CNNs were used commercially in the 1990s.

\subsection{1991: Fundamental Deep Learning Problem of Gradient Descent}
\label{1991a}
 
A diploma thesis~\citep{Hochreiter:91} represented a milestone of explicit
DL research. As mentioned in Sec.~\ref{1990}, by the late 1980s, 
experiments had indicated that traditional 
deep feedforward or recurrent networks are hard to
train by backpropagation (BP) (Sec.~\ref{1970}). Hochreiter's work
formally identified a major reason: Typical deep NNs suffer from the now famous problem of vanishing or exploding gradients. With standard activation functions (Sec.~\ref{intro}), cumulative 
 backpropagated error signals (Sec.~\ref{bpsec})  either shrink rapidly, or grow out of bounds. In fact, they decay exponentially in the number of layers or CAP depth (Sec.~\ref{caps}), 
or they explode. 
This is also known as the {\em long time lag problem}.
Much subsequent DL research of the 1990s and 2000s was motivated by this insight. 
Later work~\citep{Bengio:94} also studied
basins of attraction and their stability under noise 
from a dynamical systems point of view: either the 
dynamics are not robust to noise, or the gradients vanish. See also~\citep{Hochreiter:01book,Tino03NC}.
Over the years, several ways of partially overcoming the {\em Fundamental Deep Learning Problem} were explored:

\begin{itemize}
\item[I]
A Very Deep Learner of 1991 (the {\em History Compressor}, Sec.~\ref{1991b}) alleviates the problem through unsupervised pre-training for a hierarchy of RNNs. This greatly facilitates subsequent supervised credit assignment through BP (Sec.~\ref{1970}).
In the FNN case, similar effects can be achieved through conceptually related AE stacks (Sec.~\ref{1987},~\ref{2006}) and {\em Deep Belief Networks} (DBNs, Sec.~\ref{2006}).

\item[II]
LSTM-like networks (Sec.~\ref{1997},~\ref{2007},~\ref{2009},~\ref{2012}--\ref{dominant}) alleviate
 the problem through a special architecture unaffected by it.

\item[III]
Today's
GPU-based computers have a million times the computational power of desktop machines  of the early 1990s. 
This
allows for propagating errors a few layers further down within reasonable time, 
even in traditional NNs (Sec.~\ref{2010}). That is basically what is winning many of the image 
recognition competitions now (Sec.~\ref{2011},~\ref{2012},~\ref{2013}). (Although this does not really overcome the problem in a fundamental way.) 

\item[IV] 
Hessian-free optimization (Sec.~\ref{betterbp})
can alleviate the problem for FNNs~\citep{Moller:93,Pearlmutter:93,schraudolph02,icml2010_094} (Sec.~\ref{betterbp})
and RNNs~\citep{Martens:2011hessfree} (Sec.~\ref{2011rnn}).

\item[V] The space of NN weight matrices can also be searched without relying on error gradients,
thus avoiding the {\em Fundamental Deep Learning Problem} altogether.
Random weight guessing sometimes works better
than more sophisticated methods~\citep{Hochreiter:96sintra}.
Certain more complex problems are better solved by using 
{\em Universal Search}~\citep{Levin:73} for weight matrix-computing programs written in
a universal programming language~\citep{Schmidhuber:97nn+}.
Some are better solved by using linear methods 
to obtain optimal weights for connections to output events (Sec.~\ref{notation}),
and  {\em evolving} weights of connections to other events---this is called {\em Evolino}~\citep{Schmidhuber:07nc}.
Compare also related RNNs pre-trained by certain UL rules~\citep{steil2007},
also in the case of {\em spiking  neurons}~\citep{yin2012,maass2013} (Sec.~\ref{spiking}).
Direct search methods are relevant not only for SL but also for more general RL,
and are discussed in more detail in Sec.~\ref{evorl}.

\end{itemize}

\subsection{1991: UL-Based History Compression Through a Deep Stack of RNNs}
\label{1991b}

A working {\em Very Deep Learner} (Sec.~\ref{caps}) of 1991~\citep{chunker91and92,mydeep2013}  could perform credit assignment across hundreds of nonlinear operators or neural layers, by using {\em unsupervised pre-training} for a hierarchy of RNNs.

The basic idea is still relevant today. Each RNN is trained for a while in unsupervised fashion to predict its next input~\citep[e.g.,][]{connor1994,dorffner1996}. From then on, only unexpected inputs (errors) convey new information and get fed to the next higher RNN which thus ticks on a slower, self-organising time scale. It can easily be shown that no information gets lost. It just gets compressed (much of machine learning is 
essentially about compression, e.g., Sec.~\ref{mdl},~\ref{mdlnn},~\ref{comrl}). For each individual 
input sequence, we get a series of less and less redundant encodings in deeper and deeper levels of this 
{\em History Compressor} or {\em Neural Sequence Chunker}, which can compress data in both space (like feedforward NNs) and time. 
This is another good example of hierarchical representation learning (Sec.~\ref{hie}).
There also is a continuous 
variant of the history compressor~\citep{SchmidhuberMozerPrelinger:93}.

The RNN stack is essentially a deep generative model of the data, 
which can be reconstructed from its compressed form.
Adding another RNN to the stack improves a bound on the data's description length---equivalent to the negative logarithm of its probability~\citep{Huffman:52,Shannon:48}---as long as there is remaining local learnable predictability in the data 
representation on the corresponding level of the hierarchy.
Compare a similar observation for feedforward {\em Deep Belief Networks} (DBNs, 2006, Sec.~\ref{2006}). 

The system was able to learn many previously unlearnable DL tasks.
One ancient illustrative DL experiment~\citep{schmidhuber1993} required 
CAPs (Sec.~\ref{caps}) of depth 1200. 
The top level code of the initially unsupervised RNN stack, however, got so compact that (previously infeasible) sequence classification through additional BP-based SL became possible.
Essentially the system used UL to greatly reduce problem depth. 
Compare earlier BP-based fine-tuning of NNs initialized
by rules of propositional logic~\citep{shavlik1989} (Sec.~\ref{timelags}).

There is a way of compressing higher levels down into lower levels, thus 
fully or partially collapsing the RNN stack. The trick is to retrain a lower-level RNN to continually imitate (predict) the hidden units of an already trained, slower, higher-level RNN (the {\em``conscious" chunker}), through additional predictive output neurons~\citep{chunker91and92}. This helps the lower RNN (the {\em automatizer}) to develop appropriate, rarely changing memories that may bridge very long time lags. Again, this procedure can greatly reduce the required depth of the BP process. 

The 1991 system was a working Deep Learner in the 
modern post-2000 sense, and also a first  {\em Neural Hierarchical Temporal Memory} (HTM).
It is conceptually similar to earlier AE hierarchies (1987, Sec.~\ref{1987}) and
 later {\em Deep Belief Networks}  (2006, Sec.~\ref{2006}), but more general in the sense that it uses sequence-processing RNNs instead of FNNs with unchanging inputs. 
More recently, well-known entrepreneurs~\citep{hawkins2006,kurzweil2012} also got interested in 
HTMs; compare also hierarchical HMMs~\citep[e.g.,][]{tishby1998},
as well as later UL-based recurrent 
systems~\citep{Klapper:01,steil2007,maass2013,young2014}.
Clockwork RNNs~\citep{icml2014} also consist 
of interacting RNN modules with different clock rates,
but do not use UL to set those rates. 
Stacks of RNNs were used in later work on SL with great success, 
e.g., Sec.~\ref{1997},~\ref{2007},~\ref{2009},~\ref{2013}.

\subsection{1992: Max-Pooling (MP): Towards MPCNNs (Compare Sec.~\ref{2007},~\ref{2011})}
\label{1999}

The {\em Neocognitron} (Sec.~\ref{1979}) inspired the
{\em Cresceptron}~\citep{weng1992}, which adapts its topology during training (Sec.~\ref{mdlnn});
compare the incrementally growing and shrinking 
 GMDH networks (1965, Sec.~\ref{1965}).

Instead of using alternative local subsampling or WTA 
methods~\citep[e.g.,][]{fukushima:1980,Schmidhuber:89cs,Maass2000,Fukushima:2013},
the Cresceptron uses {\em Max-Pooling} (MP) layers. Here 
a 2-dimensional layer or array of unit activations is partitioned into 
smaller rectangular arrays. Each is replaced in a downsampling layer by the activation of its maximally active unit.
A later, more complex version of the Cresceptron~\citep{weng1997} also included {\em ``blurring"} layers
to improve object location tolerance.

The neurophysiologically plausible topology of the feedforward HMAX model~\citep{riesenhuber:1999} 
is very similar to the one of the 1992 Cresceptron (and thus to the 1979 Neocognitron).
HMAX does not learn though. Its units have hand-crafted weights; 
biologically plausible learning rules were later proposed for
similar models~\citep[e.g.,][]{serre2002,teichmann2012}.

When CNNs or convnets (Sec.~\ref{1979},~\ref{1989}) 
are combined with MP, they become Cresceptron-like or HMAX-like {\em MPCNNs} with
alternating convolutional and max-pooling layers. 
Unlike Cresceptron and HMAX, however, MPCNNs are trained by 
BP (Sec.~\ref{1970},~\ref{2007}) \citep{ranzato-cvpr-07}.
Advantages of doing this
were pointed out subsequently~\citep{scherer:2010}.
BP-trained MPCNNs 
have become
central to many modern, competition-winning, feedforward, visual Deep Learners (Sec.~\ref{2009},~\ref{2011}--\ref{dominant}).

\subsection{1994: Early Contest-Winning NNs}
\label{1994}

Back in the 1990s, certain NNs already won certain 
controlled pattern recognition contests 
with secret test sets. Notably, 
an NN with internal delay lines 
 won the Santa Fe time-series competition on chaotic intensity 
pulsations of an NH3 laser~\citep{wan1994,weigend1993}.
No very deep CAPs (Sec.~\ref{caps}) were needed though.

\subsection{1995: Supervised Recurrent Very Deep Learner (LSTM RNN)}
\label{1997}

Supervised {\em Long Short-Term Memory} (LSTM) RNN~\citep{lstm97and95,Gers:2000nc,Perez:02} 
could eventually perform similar feats as  the deep RNN hierarchy of 1991
(Sec.~\ref{1991b}),
overcoming the {\em Fundamental Deep Learning Problem} (Sec.~\ref{1991a}) without any unsupervised pre-training.
LSTM could also learn DL tasks {\em without} local sequence predictability (and thus 
{\em un}learnable by the partially 
unsupervised 1991 {\em History Compressor}, Sec.~\ref{1991b}), dealing with  
very deep problems (Sec.~\ref{caps})~\citep[e.g.,][]{Gers:02jmlr}.

The basic LSTM idea is very simple. Some of the units are called {\em Constant Error Carousels} (CECs).
Each CEC uses as an activation function $f$, the identity function, and 
has a connection to itself with fixed weight of 1.0. Due to $f$'s constant derivative of 1.0,  
errors backpropagated through a CEC cannot vanish or explode (Sec.~\ref{1991a})
but stay as they are (unless they ``flow out" 
of the CEC to other, typically {\em adaptive} parts of the NN). 
CECs are connected to several {\em nonlinear adaptive} units (some with multiplicative
activation functions) 
needed for learning nonlinear behavior. Weight changes of these units often 
profit from error signals propagated far back in time
through CECs. 
CECs are the main reason why LSTM nets can learn to discover the importance of (and memorize) events that happened thousands of discrete time steps ago, while previous RNNs already failed in case of minimal time lags of  10 steps.

Many different LSTM variants and topologies are allowed. 
It is possible to evolve good problem-specific topologies~\citep{DBLP:conf/icann/BayerWTS09}.
Some LSTM variants also use {\em modifiable} self-connections of CECs~\citep{Gers:01ieeetnn}.

To a certain extent, LSTM is biologically plausible~\citep{oreilly:2003}.
LSTM learned to solve many previously unlearnable DL tasks involving:
Recognition of the temporal order of widely separated events in noisy input streams; 
Robust storage of high-precision real numbers across extended time intervals;
Arithmetic operations on continuous input streams;
Extraction of information conveyed by the temporal distance between events;
Recognition of temporally extended patterns in noisy input sequences~\citep{lstm97and95,Gers:2000nc};
Stable generation of precisely timed rhythms, as well as 
smooth and non-smooth periodic trajectories~\citep{Gers:2000b}.
LSTM clearly outperformed
previous RNNs on tasks that
require learning the rules of regular languages describable
by deterministic {\em Finite State Automata} (FSAs)~\citep{Watrous:92nips,casey96dynamics,siegelmann93foundations,Blair+Pollack:1997nc,kalinke98computation,zeng94discrete,Manolios:94,Omlin:96,Omlin:04}, both in terms of reliability and speed.

LSTM also worked on tasks involving
context free languages (CFLs) that cannot be represented by HMMs or similar FSAs
discussed in the RNN literature~\citep{Sun93:abRNN,wiles95learning,andrews1995,steijvers96recurrent,tonkes97learning,Rodriguez:1999CS,Rodriguez+Wiles:1998:nips10}.
CFL recognition~\citep{lee-learning:96} requires the functional equivalent of a runtime stack.
Some previous RNNs failed to learn small
CFL training sets~\citep{Rodriguez+Wiles:1998:nips10}.
Those that did not~\citep{Rodriguez:1999CS,boden00context-free}
failed to extract the
general rules, and did not generalize
well on substantially larger test sets.
Similar for context-sensitive languages (CSLs)~\citep[e.g.,][]{ChalupBlairNN2003}.
LSTM generalized  well though,
requiring only the 30 shortest exemplars
($n \leq 10$) of the CSL $a^nb^nc^n$ to
correctly predict the possible continuations of sequence prefixes
for $n$ up to 1000 and more.
A combination of a decoupled extended Kalman filter~\citep{kalman1960,williams1992kalman,Puskorius:94,feldkamp1998kalman,haykin2001,feldkamp2003}
and an LSTM RNN~\citep{Perez:02}
learned to deal correctly with values of $n$ up to 10 million and more.
That is, after training the network was able to
read sequences of 30,000,000 symbols and more,
one symbol at a time, and
finally detect the subtle differences between
{\em legal} strings such as
$a^{10,000,000}b^{10,000,000}c^{10,000,000}$
and
very similar but {\em illegal} strings such as
$a^{10,000,000}b^{9,999,999}c^{10,000,000}$.
Compare also more recent RNN algorithms able to deal with long
time lags~\citep{DBLP:conf/icann/SchaferUZ06,Martens:2011hessfree,DBLP:series/lncs/ZimmermannTG12,icml2014}.

Bi-directional RNNs (BRNNs)~\citep{schuster97bidirectional,schuster99thesis} are designed for input sequences whose
starts and ends are known in advance, such as spoken sentences to be labeled by their phonemes; compare~\citep{fukada99boundary}. 
To take both past and future context of each sequence element into account,
one RNN processes the sequence from start to end, 
the other backwards from end to start. 
At each time step their combined outputs predict the corresponding label (if there is any).
BRNNs were successfully applied to secondary protein structure 
prediction~\citep{baldi99exploiting}.
DAG-RNNs~\citep{baldi2003jmlr,wu2008go} generalize BRNNs to multiple dimensions.
 They
learned to predict properties of small organic molecules~\citep{lusci2013}
as well as
protein contact maps~\citep{tegge2009},
also in conjunction with a growing deep FNN~\citep{baldi2012contact} (Sec.~\ref{2012}).
BRNNs and DAG-RNNs unfold their full potential when
combined with the LSTM concept~\citep{graves05nn,graves:2009nips,Graves:09tpami}.

Particularly successful in recent competitions are stacks (Sec.~\ref{1991b}) of LSTM RNNs~\citep{Santi:07ijcai,graves:2009nips} trained by {\em Connectionist Temporal Classification} (CTC)~\citep{Graves:06icml},
a gradient-based method for finding RNN weights that
maximize the probability of teacher-given label sequences, 
given (typically much longer and more high-dimensional) 
streams of real-valued input vectors.
CTC-LSTM performs simultaneous segmentation (alignment) and recognition
(Sec.~\ref{2013}).

In the early 2000s,  
speech recognition was dominated by
HMMs combined with FNNs \citep[e.g.,][]{bourlard+morgan:1994}.
Nevertheless, when trained from scratch on utterances from the
TIDIGITS speech database, in 2003 
LSTM already obtained results comparable to those of 
HMM-based systems~\citep{graves+eck+beringer+schmidhuber:2003,beringer:05icann,Graves:06icml}.
In 2007, LSTM outperformed HMMs in keyword spotting tasks~\citep{DBLP:conf/icann/FernandezGS07};
compare recent improvements~\citep{indermuhle2011keyword,woellmer2013}.
By 2013, LSTM also achieved best known results  on the famous 
TIMIT phoneme recognition benchmark~\citep{graves:2013icassp} (Sec.~\ref{2013}). 
Recently, LSTM RNN / HMM hybrids obtained best known performance on medium-vocabulary~\citep{geiger2014} and
large-vocabulary speech recognition~\citep{sak2014large}.

LSTM is also applicable to
robot localization~\citep{foerster-esann07},
robot control~\citep{mayer2008}, 
online driver distraction detection~\citep{woellmer2011},
and many other tasks. For example, 
it  helped to  improve the state of the art in diverse applications such as 
protein analysis~\citep{hochreiter:snowbird},
handwriting recognition~\citep{graves:08nips,Graves:09tpami,graves:2009nips,bluche13},
voice activity detection~\citep{eyben2013},
optical character recognition~\citep{breuel2013high},
language identification~\citep{gonzalez2014},
prosody contour prediction~\citep{fernandez2014},
audio onset detection~\citep{marchi2014},
text-to-speech synthesis~\citep{fan2014},
social signal classification~\citep{brueckner2014},
machine translation~\citep{sutskever2014},
 and others.

RNNs can also be used for 
metalearning~\citep{schmidhuber87,scholarpedia2010,prokhorov2002meta},
because they can in principle learn to run their own weight change algorithm~\citep{Schmidhuber:93selfrefann}. 
A successful metalearner~\citep{Hochreiter:01meta}
used an LSTM RNN to quickly {\em learn a  learning algorithm} for quadratic functions
(compare Sec.~\ref{unirl}).

Recently, LSTM RNNs won several international pattern recognition competitions 
and set numerous benchmark records 
on large and complex data sets, e.g., 
Sec.~\ref{2009},~\ref{2012},~\ref{2013}.
Gradient-based LSTM is no panacea though---other methods sometimes outperformed  
it at least on certain tasks~\citep{Jaeger:04,Schmidhuber:07nc,Martens:2011hessfree,pascanu2013,icml2014}; compare Sec.~\ref{2011rnn}.

\subsection{2003: More Contest-Winning/Record-Setting NNs; Successful Deep NNs}
\label{2003}

In the decade around 2000, many practical and commercial
pattern recognition applications were dominated by non-neural
machine learning methods
such as {\em Support Vector Machines} (SVMs)~\citep{Vapnik:95,advkernel}.
Nevertheless, at least in certain domains, NNs 
outperformed other techniques.

A  {\em Bayes NN}~\citep{neal2006b} based on an ensemble~\citep{breiman:1996,Schapire:90,wolpert:92stacked,hashem:1992,Ueda2000,dietterich2000} of NNs won 
the  {\em NIPS 2003 Feature Selection Challenge}
with secret test set~\citep{neal2006}.
The NN was not very deep though---it had two hidden layers
and thus rather shallow CAPs (Sec.~\ref{caps}) of depth 3.

Important for many present competition-winning pattern recognisers  (Sec.~\ref{2011},~\ref{2012},~\ref{2013})
were developments in the CNN department.
A BP-trained~\citep{LeCun:89} CNN (Sec.~\ref{1979}, Sec.~\ref{1989}) 
set a new MNIST record of 0.4\%~\citep{simard:2003},
using training pattern
deformations~\citep{Baird90} but no unsupervised pre-training (Sec.~\ref{1987},~\ref{1991b},~\ref{2006}).
A standard BP net achieved 0.7\%~\citep{simard:2003}.
Again, the corresponding CAP depth was low. 
Compare further improvements in Sec.~\ref{2007},~\ref{2010},~\ref{2011}.

Good image interpretation results~\citep{Behnke:LNCS} were achieved with rather deep NNs trained by the 
BP variant {\em R-prop}~\citep{rprop93} (Sec.~\ref{betterbp});
here feedback through recurrent connections helped to improve image interpretation. 
FNNs with CAP depth up to 6 were used to successfully classify high-dimensional data~\citep{vieira2003}. 

Deep LSTM RNNs started to obtain certain first speech recognition results comparable to those of 
HMM-based systems~\citep{graves+eck+beringer+schmidhuber:2003}; compare Sec.~\ref{1997},~\ref{2007},~\ref{2012},~\ref{2013}.

\subsection{2006/7: UL For Deep Belief Networks  / AE Stacks Fine-Tuned by BP}
\label{2006}

While learning networks with numerous non-linear layers
date back at least to 1965 (Sec.~\ref{1965}),
and explicit DL research results have been published at least since 1991 (Sec.~\ref{1991a},~\ref{1991b}), 
the expression {\em Deep Learning}  was actually 
coined around 2006, 
when   unsupervised pre-training of deep FNNs helped 
to accelerate subsequent SL through BP~\citep{HinSal06,hinton:06afast}.
Compare earlier terminology on {\em loading deep networks}~\citep{sima1994,windisch2005} 
and {\em learning deep memories}~\citep{Gomez:05gecco}. 
Compare also BP-based (Sec.~\ref{1970}) fine-tuning (Sec.~\ref{timelags}) of (not so deep) FNNs 
pre-trained by competitive UL~\citep{maclin1995}.

The {\em Deep Belief Network} (DBN) is a  
stack of {\em Restricted Boltzmann Machines} (RBMs)~\citep{smolensky86}, 
which in turn are {\em Boltzmann Machines} (BMs)~\citep{HintonSejnowski:86} 
with a single layer of feature-detecting units;
compare also {\em Higher-Order BMs}~\citep{memisevic2010}.
Each RBM perceives pattern representations from the level below and learns to encode
them in unsupervised fashion.
At least in theory under certain assumptions, 
adding more layers improves a bound on the data's negative log probability~\citep{hinton:06afast} 
(equivalent to the data's 
description length---compare the corresponding observation for RNN stacks, Sec.~\ref{1991b}).  
There are extensions for {\em Temporal RBMs}~\citep{sutskever2008}.


Without any training pattern deformations (Sec.~\ref{2003}),
a DBN fine-tuned by BP
achieved 1.2\% error rate~\citep{HinSal06} on the MNIST handwritten digits
(Sec.~\ref{1989},~\ref{2003}).
This result 
helped to arouse interest in DBNs. 
DBNs  also achieved good results on phoneme recognition,
with an error rate of 26.7\% on the
TIMIT core test set~\citep{mohamed2010};
compare further improvements through FNNs~\citep{speech2012,lideng2014}
and LSTM RNNs (Sec.~\ref{2013}).  

A DBN-based technique called
{\em Semantic Hashing}~\citep{salakhutdinov2009}
maps semantically similar documents (of variable size) to nearby addresses in 
a space of document representations. It
outperformed previous searchers for similar documents, 
such as {\em Locality Sensitive Hashing}~\citep{buhler2001,datar2004}.
See the RBM/DBN tutorial~\citep{fischer:13}.

Autoencoder (AE) stacks~\citep{ballard1987modular} (Sec.~\ref{1987}) 
became a popular alternative way of pre-training deep FNNs in 
unsupervised fashion, before fine-tuning (Sec.~\ref{timelags}) them through BP (Sec.~\ref{1970})~\citep{bengio2006,vincent:2008,erhan:10whydoes}. 
Sparse coding (Sec.~\ref{ulnn}) was formulated as 
a combination of convex optimization 
problems~\citep{sparse2007ng}.
Recent surveys of stacked RBM and AE methods focus
on post-2006 developments~\citep{bengio09,itamar2010}.
Unsupervised DBNs and AE stacks are conceptually similar to, but in a certain sense less general than, the
unsupervised RNN stack-based {\em History Compressor}
 of 1991 (Sec.~\ref{1991b}), which can process and re-encode
not only stationary input patterns, but 
entire pattern sequences.

\subsection{2006/7: Improved CNNs / GPU-CNNs / BP for MPCNNs / LSTM Stacks}
\label{2007}

Also in 2006, a BP-trained~\citep{LeCun:89} CNN (Sec.~\ref{1979}, Sec.~\ref{1989}) 
set a new MNIST record of 0.39\%~\citep{ranzato-06},
using training pattern
deformations (Sec.~\ref{2003}) but no unsupervised pre-training.
Compare further improvements in Sec.~\ref{2010},~\ref{2011}.
Similar CNNs were used for off-road obstacle avoidance~\citep{LeCun:06}.
A combination of CNNs and TDNNs later learned to map fixed-size representations of
variable-size sentences to features
relevant for language processing, 
using a combination of SL and UL~\citep{weston2008}.

2006 also saw an early GPU-based CNN implementation~\citep{chellapilla:2006b} up to 4 times faster
than CPU-CNNs;
compare also earlier GPU implementations of standard FNNs with a reported speed-up factor of 20~\citep{gpu2004}.
GPUs or graphics cards  have become more and more important for DL in 
subsequent years (Sec.~\ref{2010}--\ref{2013}).

In 2007, BP (Sec.~\ref{1970}) was applied for the first time~\citep{ranzato-cvpr-07}  to 
Neocognitron-inspired (Sec.~\ref{1979}),
Cresceptron-like (or HMAX-like) MPCNNs (Sec.~\ref{1999})
with alternating convolutional and max-pooling layers.
BP-trained MPCNNs  have become an 
essential ingredient of many modern, competition-winning, feedforward, visual Deep Learners (Sec.~\ref{2009},~\ref{2011}--\ref{dominant}). 

Also in 2007, 
hierarchical stacks of LSTM RNNs were introduced~\citep{Santi:07ijcai}. They can be 
trained by hierarchical {\em Connectionist Temporal Classification} (CTC)~\citep{Graves:06icml}. 
For tasks of sequence labelling, 
every LSTM RNN level (Sec.~\ref{1997}) predicts a sequence of labels fed to the next level. 
Error signals at every level are back-propagated through all the
lower levels. On spoken digit recognition, LSTM stacks
outperformed HMMs, despite making fewer assumptions about the domain.
LSTM stacks do not necessarily require unsupervised pre-training
like the earlier UL-based RNN stacks~\citep{chunker91and92} of Sec.~\ref{1991b}.

\subsection{2009: First Official Competitions Won by RNNs, and with MPCNNs}
\label{2009}

Stacks  of LSTM RNNs trained by CTC (Sec.~\ref{1997},~\ref{2007})
 became
the first   RNNs  to win
official international pattern recognition contests (with secret test sets known
only to the organisers). More precisely,
three connected handwriting competitions at ICDAR 2009 in three different languages 
(French, Arab, Farsi) were won by deep
LSTM RNNs without any {\em a priori} linguistic knowledge,
performing simultaneous segmentation and recognition.
Compare
\citep{graves05nn,Graves:09tpami,schmidhuber2011agi,graves:2013icassp,graves2014} (Sec.~\ref{2013}).

To detect human actions in surveillance videos,
a 3-dimensional CNN~\citep[e.g.,][]{seung2009,prokhorov2010}, combined with SVMs, was part of a larger 
system~\citep{trecvid2009}
using a {\em bag of features} approach \citep{nowak2006}
to extract regions of interest.
The system won three 2009 TRECVID competitions.
These were possibly the first official international contests won with the help of (MP)CNNs (Sec.~\ref{2007}).
An improved version of the method was published later~\citep{ji2013}.

2009 also saw a GPU-DBN 
implementation~\citep{raina2009large} orders of magnitudes faster than previous CPU-DBNs
(see Sec.~\ref{2006}); see also~\citep{coates:2013icml}.
The {\em Convolutional DBN}~\citep{lee:2009} (with a probabilistic 
variant of MP, Sec.~\ref{1999})
combines ideas from CNNs and DBNs,
and was successfully applied to audio classification~\citep{lee2009audio}.

\subsection{2010: Plain Backprop ($+$ Distortions) on GPU Breaks MNIST Record}
\label{2010}

 In 2010, a new MNIST (Sec.~\ref{1989}) record of 0.35\% error rate was set 
by good old BP (Sec.~\ref{1970}) in deep but otherwise
standard NNs~\citep{ciresan:2010}, 
using neither unsupervised pre-training 
(e.g., Sec.~\ref{1987},~\ref{1991b},~\ref{2006}) nor convolution 
(e.g., Sec.~\ref{1979},~\ref{1989},~\ref{2003},~\ref{2007}).
However,  training pattern
deformations (e.g., Sec.~\ref{2003}) were important to generate a big training set
and avoid overfitting.
This success was made possible mainly through a  GPU implementation of BP that was up to 50 times faster than standard
CPU versions. 
A good value of 0.95\% was obtained
without distortions except for small saccadic eye movement-like translations---compare
Sec.~\ref{2006}.

Since BP was 3-5 decades old by then (Sec.~\ref{1970}),
and pattern deformations 2 decades~\citep{Baird90} (Sec.~\ref{2003}),
these results seemed to suggest that
advances in exploiting modern
computing hardware were more important than advances in algorithms. 


\subsection{2011: MPCNNs on GPU Achieve Superhuman Vision Performance}
\label{2011}

In 2011, a flexible {\em GPU-implementation}~\citep{ciresan:2011ijcai}
 of {\em Max-Pooling (MP) CNNs or Convnets}  was  described (a GPU-MPCNN),
building on earlier MP work~\citep{weng1992} (Sec.~\ref{1999}) CNNs ~\citep{Fukushima:1979neocognitron,LeCun:89} (Sec.~\ref{1979},~\ref{1989},~\ref{2007}), 
and on early GPU-based CNNs {\em without} MP~\citep{chellapilla:2006b} (Sec.~\ref{2007});
compare  early GPU-NNs~\citep{gpu2004} and GPU-DBNs~\citep{raina2009large} (Sec.~\ref{2009}).
MPCNNs have alternating convolutional layers (Sec.~\ref{1979}) and max-pooling layers (MP, Sec.~\ref{1999}) topped by 
standard fully connected layers. All weights are trained by BP (Sec.~\ref{1970},~\ref{1989},~\ref{2007})~\citep{ranzato-cvpr-07,scherer:2010}.
GPU-MPCNNs have become essential for many contest-winning
FNNs (Sec.~\ref{2012}, Sec.~\ref{2013}).

{\em Multi-Column} GPU-MPCNNs ~\citep{ciresan:2011ijcnn} 
are committees~\citep{breiman:1996,Schapire:90,wolpert:92stacked,hashem:1992,Ueda2000,dietterich2000} of GPU-MPCNNs  with simple democratic output averaging. 
Several MPCNNs see the same input; 
their output vectors are used to assign probabilities to the various possible classes. 
The class with the on average highest probability is chosen as the system's classification of the present input. 
Compare earlier, more sophisticated ensemble methods~\citep{Schapire:90},
the contest-winning ensemble Bayes-NN~\citep{neal2006b}
of Sec.~\ref{2003},
and recent related work~\citep{shao2014}.

An ensemble of GPU-MPCNNs was the first system to achieve
superhuman visual pattern recognition~\citep{ciresan:2011ijcnn,ciresan:2012NN} in a controlled competition, namely,
the IJCNN 2011 traffic sign recognition contest in San Jose (CA)~\citep{stallkamp:11,stallkamp:12}.
This is of interest for fully autonomous, self-driving cars in traffic~\citep[e.g.,][]{Dickmanns:94}.
The GPU-MPCNN ensemble obtained 0.56\% error rate and was twice better
 than human test subjects, 
three times better than the closest artificial NN competitor~\citep{sermanet-ijcnn-11}, and
six times better than the best non-neural method.

A few months earlier, the qualifying round was won in a 1st stage online competition, albeit by a much smaller margin: 1.02\%~\citep{ciresan:2011ijcnn} vs 1.03\% for second place~\citep{sermanet-ijcnn-11}. After the deadline, the organisers revealed that human performance on the test set was 1.19\%. That is, the best methods already seemed human-competitive. However, during the qualifying it was possible to incrementally gain information about the test set by probing it through repeated submissions. This is illustrated by better and better results obtained by various teams over time~\citep{stallkamp:12}
(the organisers eventually imposed a limit of 10 resubmissions). 
In the final competition this was not possible.

This illustrates a general problem with benchmarks whose test sets are public, or at least can be probed
to some extent: competing teams tend to overfit on the test set even when it 
cannot be directly used for training, only for evaluation.

In 1997 many thought it a big deal that human chess world champion Kasparov was beaten by an IBM computer. But back then computers could not at all compete with little kids in visual pattern recognition, which seems much harder than chess from a computational perspective. 
Of course, the traffic sign domain is highly restricted,
and kids are still much better general pattern recognisers.
Nevertheless, by 2011, deep NNs could
 already learn to rival them in important limited visual domains.

An ensemble of GPU-MPCNNs was also the first method to achieve
human-competitive performance (around 0.2\%) on MNIST~\citep{ciresan2012cvpr}. 
This represented a dramatic improvement, since
by then the MNIST record had hovered around 0.4\% for almost a decade 
(Sec.~\ref{2003},~\ref{2007},~\ref{2010}). 

Given all the prior work on (MP)CNNs (Sec.~\ref{1979},~\ref{1989},~\ref{1999},~\ref{2007}) and GPU-CNNs (Sec.~\ref{2007}), 
GPU-MPCNNs are not a breakthrough in the scientific sense. 
But they are a commercially relevant breakthrough in efficient coding that has made a
difference in several contests since 2011. 
Today, most {\em feedforward} competition-winning deep NNs are (ensembles of) 
GPU-MPCNNs (Sec.~\ref{2012}--\ref{dominant}).

\subsection{2011: Hessian-Free Optimization for RNNs}
\label{2011rnn}

Also in 2011 it was shown~\citep{Martens:2011hessfree} that 
Hessian-free optimization~\citep[e.g.,][]{Moller:93,Pearlmutter:93,schraudolph02} (Sec.~\ref{betterbp})
can alleviate the 
{\em Fundamental Deep Learning Problem} (Sec.~\ref{1991a})
in RNNs, outperforming standard gradient-based 
LSTM RNNs (Sec.~\ref{1997}) on several tasks.
Compare other  RNN algorithms~\citep{Jaeger:04,Schmidhuber:07nc,pascanu2013,icml2014}
that also at least sometimes yield better results than steepest descent for LSTM RNNs.

\subsection{2012: First Contests Won on ImageNet, Object Detection, Segmentation}
\label{2012}

In 2012, an ensemble of GPU-MPCNNs (Sec.~\ref{2011}) 
achieved best results on the {\em ImageNet} classification benchmark~\citep{Krizhevsky:2012},
which is popular in the computer vision community.
Here relatively large image sizes of  256x256 pixels were necessary, 
as opposed to only 48x48 pixels for the 2011 traffic sign competition (Sec.~\ref{2011}).
See further improvements in Sec.~\ref{2013}.

Also in 2012, the biggest NN so far ($10^9$ free parameters) was trained
in unsupervised mode (Sec.~\ref{1987}, \ref{2006}) on unlabeled data~\citep{ng2012}, then applied to ImageNet. The codes across its top layer 
were used to train a simple supervised classifier,
which achieved best results so far on 20,000 classes.
Instead of relying on efficient GPU programming, this was done by brute force on 
1,000 standard machines with 16,000 cores.

So by 2011/2012, excellent results had been achieved by Deep Learners 
in image {\em recognition and classification} (Sec.~\ref{2011},~\ref{2012}).
The computer vision community, however, is especially interested in
 {\em object  detection} in large images,
for applications such as image-based search engines,
or for biomedical diagnosis where the goal may be to 
automatically detect tumors etc in images of human tissue.
Object detection presents additional challenges. 
One natural approach is to train a deep NN classifier on patches of big images,
then use it as a feature detector to be shifted 
across unknown visual scenes, using various rotations and zoom factors.
Image parts that yield highly active output units are likely
to contain objects similar to those the NN was trained on.

2012 finally saw the first DL system 
(an ensemble of GPU-MPCNNs, Sec.~\ref{2011}) to win a
contest on visual {\em object detection}~\citep{miccai2013} in large images of
several million pixels
\citep{icpr12,icpr12report}. 
Such biomedical applications may turn out to be among 
the most important applications of DL.
 The world spends over 10\% of GDP on healthcare ($>6$ trillion USD per year), much of it on medical diagnosis through expensive experts. Partial automation of this could not only save lots of money, but also make expert diagnostics accessible to many who currently cannot afford it.
It is gratifying to observe that today deep NNs may actually help to improve healthcare and 
perhaps save human lives.

2012 also saw the 
first pure {\em image segmentation} contest won by DL~\citep{ciresan2012nips}, 
again through an GPU-MPCNN ensemble 
\citep{isbi12}.\footnote{It should be mentioned, however, that LSTM RNNs already performed simultaneous segmentation and recognition when they became the first recurrent Deep Learners to win official international pattern recognition contests---see Sec.~\ref{2009}.}
EM stacks are relevant for the recently approved huge brain projects in Europe and the US~\citep[e.g.,][]{markram2012}. Given electron microscopy images of stacks of thin slices of animal brains, the goal is to build a detailed 3D model of the brain's neurons and dendrites. But human experts need many hours and days and weeks to annotate the images: Which parts depict neuronal membranes? Which parts are irrelevant background? This needs to be automated~\citep[e.g.,][]{turaga2010}. Deep Multi-Column GPU-MPCNNs learned to solve this task through experience with many training images, and won the contest on all three evaluation metrics by a large margin, with superhuman performance in terms of pixel error. 

Both object detection~\citep{miccai2013}  and image segmentation~\citep{ciresan2012nips}  
profit from fast MPCNN-based image scans that avoid redundant computations.
Recent MPCNN scanners  
speed up naive implementations  by up to 
three orders of magnitude~\citep{masci:2013icip,Giusti:2013a};
compare earlier efficient methods for CNNs {\em without} MP~\citep{vaillant-monrocq-lecun-94}.

Also in 2012,
a system consisting of  growing deep FNNs and 
2D-BRNNs~\citep{baldi2012contact} won the CASP 2012 contest
on protein contact map prediction.
On the IAM-OnDoDB benchmark,
LSTM RNNs (Sec.~\ref{1997}) outperformed all other methods (HMMs, SVMs) on 
online mode detection~\citep{otte2012local,indermuhle2012mode}
and keyword spotting~\citep{indermuhle2011keyword}.
On the long time lag problem of language modelling, LSTM RNNs 
outperformed all statistical approaches on the IAM-DB benchmark~\citep{frinken2012long};
improved results were later obtained through a combination of NNs and HMMs~\citep{zamora2014}.
Compare earlier RNNs for object 
recognition through iterative image interpretation~\citep{Behnke:IJCNN1998,Behnke:ICANN2002,Behnke:LNCS};
see also more recent publications~\citep{wyatte2012b,oreilly2013} extending
work on biologically plausible learning rules for RNNs~\citep{oreilly1996}.

\subsection{2013-: More Contests and Benchmark Records}
\label{2013}

A stack~\citep{Santi:07ijcai,graves:2009nips} (Sec.~\ref{1991b}) of bi-directional LSTM 
RNNs~\citep{graves05nn}  
trained by CTC (Sec.~\ref{1997},~\ref{2009}) 
broke a famous TIMIT speech (phoneme) recognition record, achieving 17.7\% test set error rate~\citep{graves:2013icassp}, despite thousands of man years previously spent on {\em Hidden Markov Model} (HMMs)-based speech recognition research. 
Compare earlier DBN results (Sec.~\ref{2006}).

CTC-LSTM also 
helped to score first at NIST's OpenHaRT2013 evaluation~\citep{bluche13}.
For {\em optical character recognition} (OCR), LSTM RNNs outperformed commercial 
recognizers of historical data~\citep{breuel2013high}.
LSTM-based systems  also set benchmark records in
{\em language identification}~\citep{gonzalez2014},
medium-vocabulary {\em speech recognition}~\citep{geiger2014},
{\em prosody contour prediction}~\citep{fernandez2014},
{\em audio onset detection}~\citep{marchi2014},
{\em text-to-speech synthesis}~\citep{fan2014}, and
{\em social signal classification}~\citep{brueckner2014}.

An LSTM RNN was used to estimate the state posteriors of an HMM; this system
 beat the previous state of the art in {\em large vocabulary speech recognition}~\citep{sak2014,sak2014large}. 
Another LSTM RNN with hundreds of millions of connections was used to rerank hypotheses of a statistical machine translation system; this system
beat the previous state of the art in {\em English to French translation}~\citep{sutskever2014}.

A new record on the {\em ICDAR Chinese handwriting recognition 
benchmark} (over 3700 classes) was set on a desktop machine
by an ensemble of GPU-MPCNNs (Sec.~\ref{2011}) with almost human performance~\citep{chinese2013};
compare~\citep{icdar2013}.

The {\em MICCAI 2013 Grand Challenge on Mitosis Detection}~\citep{miccai13}
also was won by an object-detecting  GPU-MPCNN ensemble~\citep{miccai2013}.
Its data set was even larger and more challenging than the one of ICPR 2012 (Sec.~\ref{2012}): a real-world dataset including many ambiguous cases and frequently encountered problems such as imperfect slide staining. 

Three 2D-CNNs (with {\em mean-pooling} instead of MP, Sec.~\ref{1999}) observing three orthogonal projections of 3D images
outperformed traditional full 3D methods on the task of segmenting tibial
cartilage in low field knee MRI scans~\citep{prasoon:13}.

Deep GPU-MPCNNs (Sec.~\ref{2011}) also helped to achieve new best results
on important benchmarks of the computer vision community:
ImageNet classification~\citep{zeiler2013,szegedy2014} 
and---in 
conjunction with traditional 
approaches---PASCAL object detection~\citep{malik2013}.
They also learned to predict bounding box coordinates  of
 objects in the Imagenet 2013 database,
and obtained state-of-the-art results on tasks of localization and 
detection~\citep{sermanet2013}.
GPU-MPCNNs also helped to recognise
multi-digit numbers  in Google Street View images~\citep{goodfellow2014multi},
where part of the NN was trained to {\em count} visible digits;
compare earlier work on detecting ``numerosity" through DBNs~\citep{stoianov2012}.
This system also excelled at recognising distorted synthetic text in {\em reCAPTCHA} puzzles.
Other successful CNN applications include 
scene parsing~\citep{Farabet2013}, object detection~\citep{Szegedy2013},
shadow detection~\citep{khan2014},
video classification~\citep{karpathy2014},
and Alzheimer’s disease neuroimaging~\citep{shuiwang2014}.

Additional contests are mentioned in the web pages of
the Swiss AI Lab IDSIA,
the University of Toronto,
NY University,
and the University of Montreal.


\subsection{Currently Successful Techniques: LSTM RNNs and GPU-MPCNNs}
\label{dominant}

Most competition-winning or benchmark record-setting Deep Learners actually use one of two {\em supervised} techniques: (a) recurrent LSTM (1997)  trained by CTC (2006) (Sec.~\ref{1997},~\ref{2009},~\ref{2012},~\ref{2013}), or (b) 
 feedforward GPU-MPCNNs (2011, Sec.~\ref{2011},~\ref{2012},~\ref{2013})
based on CNNs (1979, Sec.~\ref{1979}) with MP (1992, Sec.~\ref{1999}) 
trained through BP (1989--2007, Sec.~\ref{1989},~\ref{2007}).

Exceptions include two 2011 contests~\citep{goodfellow2011,transfer2011,goodfellow:2012icml} specialised on {\em Transfer 
Learning} from one dataset to another~\citep[e.g.,][]{caruana1997,Schmidhuber:04oops,transfer2010}.
However, deep GPU-MPCNNs do allow for {\em pure SL-based transfer}~\citep{Ciresan:2012a}, 
where pre-training
on one training set greatly improves performance on quite different sets,
also in more recent studies~\citep{oquab2013,donahue2013}. 
In fact, deep MPCNNs pre-trained by SL  can extract useful 
features from quite diverse off-training-set images, yielding better results than traditional, 
widely used features 
such as SIFT~\citep{Lowe:1999,Lowe:04} on many vision tasks~\citep{razavian2014}.
To deal with changing datasets, 
slowly learning deep NNs were also combined with  
rapidly adapting {\em ``surface"} NNs~\citep{kak2010}.

Remarkably, 
in the 1990s a trend went from partially unsupervised RNN stacks (Sec.~\ref{1991b}) to {\em purely} supervised LSTM RNNs (Sec.~\ref{1997}), just like in the 2000s a trend went from partially unsupervised FNN stacks (Sec.~\ref{2006}) to {\em purely} supervised MPCNNs 
(Sec.~\ref{2007}--\ref{2013}).
Nevertheless, in many applications it can still be advantageous to combine the best of both worlds---{\em supervised} learning and {\em unsupervised} pre-training (Sec.~\ref{1991b},~\ref{2006}).

\subsection{Recent Tricks for Improving SL Deep NNs (Compare Sec.~\ref{betterbp},~\ref{mdlnn})}
\label{tricks}

DBN training (Sec.~\ref{2006}) can be improved through gradient enhancements and
automatic  learning rate adjustments during stochastic gradient descent~\citep{cho2013,cho2014}, and through 
Tikhonov-type~\citep{tikhonov1977} regularization of RBMs~\citep{cho2012}.
Contractive AEs~\citep{vincent2011} 
discourage hidden unit perturbations in response to input perturbations, 
similar to how FMS (Sec.~\ref{mdlnn}) for {\sc Lococode} AEs (Sec.~\ref{ulnn}) 
discourages output perturbations in response to weight perturbations.

Hierarchical CNNs in a {\em Neural Abstraction Pyramid}~\citep[e.g.,][]{Behnke:LNCS,Behnke:NCA}
were trained to 
 reconstruct images corrupted by structured noise~\citep{Behnke:IJCIA2001},
thus enforcing increasingly abstract image representations in 
deeper and deeper layers.
{\em Denoising AEs} later used a similar procedure~\citep{vincent:2008}.

{\em Dropout}~\citep{Hinton2012,frey2013}  removes units from NNs during training to improve generalisation. Some view it as an ensemble method that trains multiple data models simultaneously~\citep{baldidropout2014}. 
Under certain circumstances, 
it could also be viewed as a form of training set augmentation: 
effectively, more and more informative complex features are removed from the training data. 
Compare dropout for RNNs~\citep{Pham2013,pachitariu2013regularization,pascanu2013construct}. 
A deterministic approximation coined {\em fast dropout}~\citep{wang2013fast} can lead to faster learning and evaluation and was adapted for RNNs~\citep{bayer2013fast}.
Dropout is closely related to older, biologically plausible techniques 
for adding noise to neurons 
or synapses during training~\citep[e.g.,][]{hanson1990,Murray:93,Schuster:92,Nadal:94,giles95,an96},
which in turn are closely related to finding perturbation-resistant low-complexity NNs,
e.g., through FMS (Sec.~\ref{mdlnn}).
MDL-based stochastic variational methods~\citep{Graves2011} 
are also related to FMS.
They are useful for RNNs, where classic regularizers such as  weight decay
(Sec.~\ref{mdlnn}) represent a 
bias towards limited memory capacity~\citep[e.g.,][]{pascanu2013}. 
Compare recent work on variational recurrent AEs~\citep{bayer2014variational}.

The activation function $f$ of {\em Rectified Linear Units} (ReLUs)  is
$f(x) =  x$ for $x > 0, f(x)=0$ otherwise---compare the old concept of half-wave rectified units~\citep{malik1990}.
ReLU NNs are useful for RBMs~\citep{Nair2010,Maas2013}, outperformed sigmoidal activation
functions in deep NNs~\citep{Glorot2011a}, and helped to obtain  best results on several benchmark 
problems across multiple domains~\citep[e.g.,][]{Krizhevsky:2012,Dahl2013}.

NNs with competing linear units tend to outperform those with non-competing nonlinear
units, and avoid catastrophic forgetting through BP 
when training sets change over time~\citep{srivastava2013compete}. In this context,
choosing a learning algorithm may be more important than choosing activation functions~\citep{Goodfellow2014}.
{\em Maxout} NNs~\citep{goodfellow2013maxout} combine
competitive interactions and 
dropout (see above) to achieve excellent results on certain
benchmarks.
Compare early RNNs with competing units for SL and RL~\citep{Schmidhuber:89cs}.
To address overfitting, instead of depending on  pre-wired regularizers and hyper-parameters~\citep{Hertz:91,bishop:2006},
self-delimiting RNNs (SLIM NNs) with competing units~\citep{Schmidhuber:12slimnn}
can in principle learn to select their own runtime and their own numbers of effective free parameters,
thus learning their own computable regularisers (Sec.~\ref{mdl},~\ref{mdlnn}), 
 becoming fast and  {\em slim} when necessary.
One may  penalize the task-specific  total length of connections~\citep[e.g.,][]{maass2002wire,Schmidhuber:12slimnn,Schmidhuber:13powerplay,clune2013modular} 
and communication costs of 
SLIM NNs implemented on the 3-dimensional brain-like multi-processor 
hardware to be expected in the future.

{\em RmsProp}~\citep{Tieleman2012,Schaul2012} can speed up first order gradient descent methods (Sec.~\ref{1970},~\ref{betterbp}); 
compare vario-$\eta$~\citep{DBLP:conf/nips/NeuneierZ96},
{\em Adagrad}~\citep{Duchi2011} and {\em Adadelta}~\citep{zeiler2012}.
DL in NNs can also be improved
by transforming hidden unit activations such that they have zero output and slope on average~\citep{raiko2012}.
Many additional, older tricks (Sec.~\ref{betterbp},~\ref{mdlnn}) should also be applicable to today's deep NNs;
compare~\citep{orr1998neural,tricksofthetrade:2012}.

\subsection{Consequences for Neuroscience}
\label{bnn}
It is ironic that artificial NNs (ANNs) can help to better understand biological NNs (BNNs)---see the ISBI 2012 results 
mentioned in Sec.~\ref{2012}~\citep{isbi12,ciresan2012nips}.

The feature detectors learned by single-layer visual ANNs 
are similar to those found in early visual processing stages of BNNs
(e.g., Sec.~\ref{ulnn}).
Likewise, the feature detectors learned in {\em deep} layers of visual ANNs should be highly predictive of what neuroscientists will find in {\em deep} layers of BNNs. While the visual cortex of BNNs may use quite different learning algorithms, its objective function to be minimised may be quite similar to the one of visual ANNs. 
In fact, results obtained 
with relatively deep artificial DBNs~\citep{lee2007sparse} and CNNs~\citep{Yamins2013}
seem compatible with insights about
the visual pathway in the primate cerebral cortex,
which has been studied for many decades~\citep[e.g.,][]{hubel1968,perrett1982,desimone1984,felleman1991,perrett1992,Tanaka:94,logothetis1995,Bichot:05,poggio2005,lennie2005,connor2007,kriegeskorte2008,dicarlo2012}; compare a 
computer vision-oriented survey~\citep{kruger2013}.

\subsection{DL with Spiking Neurons?}
\label{spiking}
Many recent DL results profit from GPU-based traditional deep NNs, e.g., Sec.~\ref{2007}--\ref{2011}. Current GPUs, however, are little ovens, much hungrier for energy than biological brains, whose neurons efficiently communicate by brief spikes~\citep{hodgkin1952,fitzhugh1961,nagumo1962}, 
and often remain quiet. Many computational models of such 
{\em spiking neurons} have been proposed and analyzed~\citep[e.g.,][]{gerstner1992,zipser1993,stemmler1996,tsodyks1996,maex1996,maass1996,maass1997,kistler1997,amit1997,tsodyks1998,kempter1999,song2000,stoop2000,brunel2000,bohte2002,gerstnerbook,izhikevich2003,seung2003,deco2005,brette2007,brea2013,nessler2013,kasabov2014,maass2014,rezende2014}.

Future energy-efficient hardware for DL in NNs may implement aspects of such models~\citep[e.g.,][]{liu2001,roggen2003,glackin2005,schemmel2006,fieres2008,khan2008,serrano2009,jin2010,indiveri2011,neil2014,merolla2014}. 
A simulated, event-driven, spiking variant~\citep{nefti2014} of an RBM (Sec.~\ref{2006})
was trained by a variant of the
{\em Contrastive Divergence} algorithm~\citep{hinton:2002}.
Spiking nets were evolved to achieve reasonable performance on small face recognition data sets~\citep{wysoski2010} and to control simple robots~\citep{floreano2001,hagras2004}.
A spiking DBN with about 250,000 neurons~\citep[as part of a larger NN;][]{eliasmith2012,eliasmith2013} 
achieved 6\% error rate on MNIST;
compare similar results with a spiking DBN variant of depth 3 using a neuromorphic event-based sensor~\citep{oconnor2013}. 
In practical applications, however,
current artificial networks of
spiking neurons cannot yet compete with the best traditional deep NNs
(e.g., compare MNIST results of Sec.~\ref{2011}). 

\newpage
\section{DL in FNNs and RNNs for Reinforcement Learning (RL)}
\label{deeprl}

So far we have focused on 
Deep Learning (DL) in supervised or unsupervised NNs.
Such NNs learn to perceive / encode / predict / classify 
patterns or pattern sequences, 
but they do not learn to act in the more general
sense of {\em Reinforcement Learning} (RL) in unknown environments \citep[see surveys, e.g.,][]{Kaelbling:96,Sutton:98,wiering2012}.
Here we add a discussion of DL FNNs and RNNs for RL. 
It will be shorter than the discussion of FNNs and RNNs for SL and UL (Sec.~\ref{super}),
reflecting the current size of the various fields.

Without a teacher, solely from occasional  real-valued
pain and pleasure signals,  RL agents must discover how to interact with a
dynamic, initially unknown environment to maximize their expected cumulative reward
signals (Sec.~\ref{notation}).
There may be arbitrary, a priori unknown delays between actions and perceivable consequences.
The problem is as hard as any problem of computer science,
since any task with a computable description can be formulated in the RL framework \citep[e.g.,][]{Hutter:05book+}.
For example, an answer to the famous question of 
whether $P=NP$~\citep{Levin:73,Cook:71}
would also set limits for what is achievable by general RL.
Compare more specific limitations, e.g.,~\citep{blondel2000,madani2003,vlassis2012}.
The following subsections mostly focus on certain obvious intersections 
between DL and RL---they cannot serve as a general RL survey.  

\subsection{RL Through NN World Models Yields RNNs With Deep CAPs}
\label{worrl}

In the special case of an RL FNN controller 
$C$ interacting with a 
{\em deterministic, predictable} environment, 
a separate FNN called $M$ 
can learn to become $C$'s {\em world model} through {\em system identification},
predicting $C$'s inputs from previous actions and inputs~\citep[e.g.,][]{Werbos:81sensitivity,Werbos:87,Munro:87,Jordan:88,Werbos:89identification,Werbos:89neurocontrol,RobinsonFallside:89,JordanRumelhart:90,Schmidhuber:90sandiego,narendra1990,Werbos:92sticky,kawato1993,cochocki1993,levin1995,miller1995,ljung1998,prokhorov2001,ge2010}.
Assume $M$ has learned to produce accurate predictions.
We can use $M$ to substitute the environment.
Then $M$ and $C$ form an RNN where $M$'s outputs become inputs of $C$, 
whose outputs (actions) in turn become inputs of $M$.
Now BP  for RNNs (Sec.~\ref{bp}) can be used 
to achieve {\em desired input events}  such as high real-valued reward signals:
While $M$'s weights remain fixed,
gradient information for $C$'s weights is propagated
back through $M$ down into $C$ and
back through $M$ etc. 
To a certain extent, the approach is also applicable in probabilistic or uncertain environments, as long as the inner products of $M$'s $C$-based gradient estimates and $M$'s ``true" gradients tend to be positive. 

In general, this approach
 implies deep CAPs for $C$, unlike in DP-based traditional  RL (Sec.~\ref{trarl}).
Decades ago, the method was used to
 learn to back up a model truck~\citep{NguyenWidrow:89}.
An RL active vision system used it to learn sequential shifts (saccades) of a fovea, to
detect targets in visual scenes~\citep{SchmidhuberHuber:91},
thus learning to control selective attention.
Compare RL-based attention learning without NNs~\citep{Whitehead:92}.

To allow for memories of previous events in
{\em partially observable worlds}
 (Sec.~\ref{pomrl}),
the most general variant of this technique uses 
RNNs instead of FNNs to implement both $M$ and $C$ ~\citep{Schmidhuber:90sandiego,Schmidhuber:91nips,feldkamp1998}.
This may cause deep CAPs not only for $C$ but also for $M$.

$M$ can also be used to optimize expected reward by {\em planning} future action sequences~\citep{Schmidhuber:90sandiego}.
In fact, the winners of the 2004 RoboCup World Championship in the fast league~\citep{egorova03}
trained NNs to predict the effects of steering signals on fast
robots with 4 motors for 4 different wheels. During play, such NN models were used 
to achieve desirable subgoals, 
by optimizing action sequences through quickly planning ahead. The approach also
was used to create {\em self-healing} robots able to compensate for faulty motors whose effects do not longer 
match the predictions of the NN models~\citep{gloye05,schmidhuber2007pro}. 

Typically $M$ is not given in advance. Then 
an essential question is: which experiments should $C$ conduct to quickly improve $M$?
The {\em Formal Theory of Fun and Creativity}~\citep[e.g.,][]{Schmidhuber:06cs,Schmidhuber:13powerplay}
 formalizes driving forces and value functions behind such curious and exploratory behavior: 
A measure of the {\em learning progress} of $M$ becomes the intrinsic reward of $C$~\citep{Schmidhuber:91singaporecur}; compare~\citep{Singh:05nips,Oudeyer:12intrinsic}.
This motivates $C$ to create action sequences (experiments) such that $M$ makes quick progress.

\subsection{Deep FNNs for Traditional RL and Markov Decision Processes (MDPs)}
\label{trarl}

The classical
approach to RL~\citep{Samuel:59,Bertsekas:96} makes the simplifying 
assumption of {\em Markov Decision Processes} (MDPs): 
the current input of the RL agent
conveys all information necessary to compute an optimal next 
output event or decision. 
This allows for greatly reducing CAP depth in RL NNs (Sec.~\ref{caps},~\ref{worrl}) 
by using the {\em Dynamic Programming} (DP) trick~\citep{Bellman:1957}.
The latter is often explained in a probabilistic framework~\citep[e.g.,][]{Sutton:98}, 
but its basic idea can already be conveyed in a deterministic setting.
For simplicity, 
using the notation of Sec.~\ref{notation},
let input events $x_t$ encode the entire current state of
the environment, including a real-valued reward $r_t$
(no need to introduce additional vector-valued notation, 
since real values can encode arbitrary vectors of real values).
The original RL goal (find weights that maximize the sum of all rewards of an episode) 
is replaced by an equivalent set of alternative goals set by a
real-valued value function $V$ defined on input events.
Consider any two subsequent input events $x_t,x_k$.
Recursively define $V(x_t)=r_t+V(x_k)$, where $V(x_k)=r_k$ if $x_k$ is the {\em last} input event.
Now search for weights that maximize the $V$ 
of all input events, 
by causing appropriate output events or actions. 

Due to the Markov assumption,
an FNN suffices to implement the policy that maps input to output events.
Relevant CAPs are not deeper than this FNN. 
$V$ itself is often modeled by a {\em separate FNN} (also yielding typically short CAPs) 
learning to approximate $V(x_t)$ 
only from {\em local} information $r_t, V(x_k)$.

Many variants of traditional RL exist~\citep[e.g.,][]{BartoSuttonAnderson:83,Watkins:89,WatkinsDayan:92,Moore:93,Schwartz:93,Rummery:94,Singh:94R,Baird:95,Kaelbling:95,Peng:96,Mahadevan:96,Tsitsiklis:96,96-BradtkeLstd,Santamaria:97,prwu97,Sutton:98,Wiering:98,baird:nips12,meuleau:icuai99,Doya:00,Bertsekas:01,brafman02,Abounadi:02,03-LspiLagoudakis,09-Gtd,10-GqLambda,hasselt2012}.
Most are formulated in a probabilistic framework,
and evaluate pairs of input and output (action) events (instead of input events only).
To facilitate certain mathematical derivations,
some discount delayed rewards,
but such distortions of the original RL problem are problematic.

Perhaps the most well-known RL NN is the  world-class RL backgammon player~\citep{Tesauro:94},
which achieved the level of human world champions by playing against itself.
Its nonlinear, rather shallow FNN maps a large but finite
number of discrete board states to values.   
More recently, a rather deep GPU-CNN  was used in
a traditional RL framework  to play several Atari 2600 computer games directly from 
84x84 pixel 60 Hz video input~\citep{atari2013},
using experience replay~\citep{Lin:93},
extending previous work on {\em Neural Fitted Q-Learning} (NFQ)~\citep{nfq}.
Even better results are achieved by using (slow) Monte Carlo tree planning to
train comparatively fast deep NNs~\citep{atarimcts2014}.
Compare RBM-based RL~\citep{sallans2004} with high-dimensional inputs~\citep{elfwing2010},
earlier RL Atari players~\citep{gruettner2010multi},
and an earlier, raw video-based RL NN for computer games~\citep{koutnik:gecco13} trained by {\em Indirect Policy Search}
(Sec.~\ref{comrl}).

\subsection{Deep RL RNNs for Partially Observable MDPs (POMDPs)}
\label{pomrl}

The {\em Markov assumption} 
(Sec.~\ref{trarl}) is often unrealistic. We cannot directly perceive 
what is behind our back, let alone the current state of the entire universe. 
However, memories of previous events
can help to deal with  
{\em partially observable Markov decision problems} (POMDPs)~\citep[e.g.,][]{Schmidhuber:90sandiego,Schmidhuber:91nips,Ring:91,Ring:93,Ring:94,Williams:92,Lin:93,Teller:94,Kaelbling:95,Littman:95,Boutilier:96,Jaakkola:95,McCallum:96,kimura1997,Wiering:96levin,Wiering:97ab,otsuka2010}.
A naive way of implementing memories without leaving the MDP framework 
(Sec.~\ref{trarl}) would be to simply consider a 
possibly huge state space, namely, 
the set of all possible observation histories and their prefixes. 
A more realistic way is to use function approximators such as RNNs that produce 
compact state features as a function of the entire history seen so far.
Generally speaking, POMDP RL often uses DL RNNs to learn 
which events to memorize and which to ignore.
Three basic alternatives are: 
\begin{enumerate}
\item
Use an RNN as a value function mapping arbitrary event histories to values~\citep[e.g.,][]{Schmidhuber:90cmss,Schmidhuber:91nips,Lin:93,Bakker:01nips}. 
For example, deep LSTM RNNs were used in this way for RL robots~\citep{Bakker:03robot}.
\item
Use an RNN controller in conjunction with a second RNN as predictive world model,
to obtain a combined RNN with deep CAPs---see Sec.~\ref{worrl}.
\item
Use an RNN for RL by {\em Direct Search} (Sec.~\ref{evorl}) or {\em Indirect Search} (Sec.~\ref{comrl}) in weight space.
\end{enumerate}
In general, however, POMDPs may imply greatly increased CAP depth.

\subsection{RL Facilitated by Deep UL in FNNs and RNNs}
\label{unsrl}

RL machines 
may profit from UL for input preprocessing~\citep[e.g.,][]{Jodogne07}.
In particular, an UL NN  can 
learn to compactly encode environmental inputs such as images or videos,
e.g., Sec.~\ref{1987},~\ref{1991b},~\ref{2006}.
The compact codes (instead of the high-dimensional 
raw data) can be fed into an RL machine,
whose job thus may become much easier~\citep{Legenstein2010,cuccu2011},
just like SL may profit from UL, e.g., Sec.~\ref{1987},~\ref{1991b},~\ref{2006}.
For example, NFQ~\citep{nfq} was applied to real-world control tasks~\citep{lange,rieijcnn12} 
where purely visual inputs were compactly encoded by deep
autoencoders (Sec.~\ref{1987},~\ref{2006}). 
RL combined with 
UL based on {\em Slow Feature Analysis}~\citep{WisSej2002,DBLP:journals/neco/KompellaLS12} enabled
a real humanoid robot to learn skills from raw high-dimensional video streams~\citep{luciwkomp13}.
To deal with POMDPs (Sec.~\ref{pomrl}) involving high-dimensional inputs, 
RBM-based RL was used~\citep{otsuka2010phd},
and a RAAM~\citep{pollack1988implications} (Sec.~\ref{1987})
was employed as a deep unsupervised sequence encoder  
 for RL~\citep{Gisslen2011agi}.
Certain types of RL and UL also were combined in biologically plausible RNNs with spiking 
neurons (Sec.~\ref{spiking})~\citep[e.g.,][]{yin2012,maass2013,rezende2014}.

\subsection{Deep Hierarchical RL (HRL) and Subgoal Learning with FNNs and RNNs}
\label{subrl}
 
Multiple learnable 
levels of abstraction~\citep{Fu:77,Lenat:84,Ring:94,bengio2013tpami,lideng2014} seem as important for RL as for SL. 
Work on NN-based {\em Hierarchical RL} (HRL) has been
published since the early 1990s. 
In particular,
gradient-based {\em subgoal discovery} with FNNs or RNNs decomposes RL tasks into subtasks
for RL submodules~\citep{Schmidhuber:91icannsubgoals,SchmidhuberWahnsiedler:92sab}.
Numerous alternative HRL techniques have been proposed~\citep[e.g.,][]{Ring:91,Ring:94,Jameson:91,TenenbergKarlssonWhitehead,Weiss:94a,partigame,Precup:MTimeNIPS98,Dietterich:MAXQ,menache2002,DoyaSamejimaKatagiriKawato,ghavamzadehICML03,barto2003hrl,SamejimaDoyaKawato,Bakker:04ias,stoneML05,simsek2008skill}.
While HRL frameworks such as {\em Feudal RL}~\citep{Dayan:93} 
and {\em options}~\citep{sutton1999between,Barto:04,Singh:05nips} 
do not directly address the problem of automatic subgoal discovery,
{\em HQ-Learning}~\citep{Wiering:97ab}  automatically decomposes POMDPs (Sec.~\ref{pomrl})
into sequences of simpler subtasks that can be solved by memoryless policies
learnable by reactive sub-agents.    
Recent HRL organizes potentially deep NN-based RL sub-modules into
self-organizing, 2-dimensional motor control maps~\citep{ring:icdl2011}
inspired by neurophysiological findings~\citep{Graziano:book}.

\subsection{Deep RL by Direct NN Search / Policy Gradients / Evolution}
\label{evorl}

Not quite as universal as the methods of Sec.~\ref{unirl},
yet both practical and more general than most traditional RL algorithms (Sec.~\ref{trarl}), are 
methods for {\em Direct Policy Search} (DS).
Without a need for value functions or Markovian assumptions (Sec.~\ref{trarl},~\ref{pomrl}), 
the weights of an FNN or RNN are directly evaluated on the given RL problem.
The results of successive trials inform further search for better weights.
Unlike with RL supported by BP (Sec.~\ref{1970},~\ref{pomrl},~\ref{worrl}),
CAP depth (Sec.~\ref{caps},~\ref{1991a}) is not a crucial issue.
DS may solve the credit assignment problem without 
backtracking through deep causal chains of 
modifiable parameters---it neither cares for their existence,
nor tries to exploit them.

An important class of DS methods for NNs are 
{\em Policy Gradient}
methods~\citep{Williams:86,Williams:88,Williams:92,Sutton:99,baxter2001,aberdeenthesis,ghavamzadehICML03,stoneICRA04,wierstraCEC08,rueckstiess2008b,peters2008neuralnetworks,peters2008neurocomputing,sehnke2009parameter,gruettner2010multi,wierstra2010,peters2010,grondman2012,heess2012}.
Gradients of the total reward with respect to policies (NN weights) are 
estimated (and then exploited) through repeated NN evaluations.

RL NNs can also be evolved through
{\em Evolutionary Algorithms} (EAs)~\citep{Rechenberg:71,Schwefel:74,Holland:75,Fogel:66,goldberg:gabook89}
 in a series of trials.
Here several policies are represented by a population
of NNs improved through mutations and/or
repeated recombinations of the population's fittest individuals~\citep[e.g.,][]{montana1989,fogel1990,maniezzo1994,happel1994,nolfi1994}.
Compare {\em Genetic Programming} (GP)~\citep{Cramer:85}~\citep[see also][]{smith80} which
can be used to evolve computer programs of variable size~\citep{gp87,Koza:92},
and {\em Cartesian GP}~\citep{miller2000,miller2009} 
for evolving graph-like programs,
including NNs~\citep{khan2010} and their topology~\citep{turner2013}.
Related methods include 
{\em probability distribution-based EAs}~\citep{Baluja:94,saravanan:ieeeexpert95,Salustowicz:97ecj,Larraanaga2001}, 
{\em Covariance Matrix Estimation Evolution Strategies} (CMA-ES)~\citep{hansenCMA,hansen2003,igel:cec03,heidrich-meisner:09},
and {\em NeuroEvolution of Augmenting Topologies} (NEAT)~\citep{stanley:ec02}.
Hybrid methods combine traditional NN-based RL (Sec.~\ref{trarl}) and EAs~\citep[e.g.,][]{whiteson2006}.

Since RNNs are general computers, 
RNN evolution is like GP in the sense that it can evolve general programs.
Unlike sequential programs learned by traditional GP, however, RNNs can mix sequential and parallel information processing in a natural and efficient way, as already 
mentioned in Sec.~\ref{intro}. Many RNN evolvers have been proposed \citep[e.g.,][]{miller:icga89,wieland1991,cliff1993,yao:review93,nolfi:alife4,Sims:1994:EVC,yamauchi94sequential,miglino95evolving,moriarty:phd,pasemann99,juang2004,whiteson2012}.
One particularly effective  family of methods {\em coevolves} neurons, combining them into networks, and
selecting those neurons for reproduction that participated in the best-performing
networks~\citep{moriarty:ml96,gomez:phd,Gomez:03}. This
can help to solve deep POMDPs~\citep{Gomez:05gecco}.
{\em Co-Synaptic Neuro-Evolution} (CoSyNE) does something similar on the level of synapses or weights~\citep{Gomez:08jmlr};
benefits of this were shown on difficult nonlinear POMDP benchmarks.

{\em Natural Evolution Strategies} (NES)~\citep{wierstraCEC08,glasmachers:2010b,Sun2009a,sun:gecco13} link policy
gradient methods and evolutionary approaches through the concept of {\em Natural Gradients}~\citep{amari1998natural}.
RNN evolution may also help to improve SL for deep RNNs 
through {\em Evolino}~\citep{Schmidhuber:07nc} (Sec.~\ref{1991a}).

\subsection{Deep RL by Indirect Policy Search / Compressed NN Search}
\label{comrl}

Some DS methods (Sec.~\ref{evorl}) can evolve NNs  
with hundreds or thousands of weights, but not millions. 
How to search for large and deep NNs? 
Most SL and RL methods mentioned so far somehow search the space of weights $w_i$. 
Some profit from a reduction of the search space through shared $w_i$ 
that get reused over and over again, e.g., in CNNs (Sec.~\ref{1979},~\ref{1989},~\ref{2007},~\ref{2012}),
or in RNNs for SL (Sec.~\ref{1970},~\ref{1997},~\ref{2009}) and RL (Sec.~\ref{worrl},~\ref{pomrl},~\ref{evorl}).

It may be possible, however, to exploit {\em additional} regularities/compressibilities 
in the space of solutions, through {\em indirect search in weight space}.
Instead of evolving large NNs directly (Sec.~\ref{evorl}), one can sometimes greatly reduce
the search space by evolving 
{\em compact encodings} of NNs, e.g.,  through  {\em Lindenmeyer Systems}~\citep{lindenmayer68,lindenmayer94}, {\em graph rewriting}~\citep{kitano90}, {\em Cellular Encoding}~\citep{gruau:tr96-048}, {\em HyperNEAT}~\citep{stanley07,stanley09,clune2011,vandenberg2013} (extending
NEAT; Sec.~\ref{evorl}), and extensions thereof~\citep[e.g.,][]{risi2012}. 
This helps to avoid overfitting (compare Sec.~\ref{mdlnn},~\ref{tricks}) and is 
closely related to the topics of regularisation 
and MDL (Sec.~\ref{mdl}).

A general approach~\citep{Schmidhuber:97nn+} for both SL and RL seeks to compactly encode weights of large NNs~\citep{Schmidhuber:97nn+} through programs written in a {\em universal programming language}~\citep{Goedel:31,Church:36,Turing:36,Post:36}. Often it is much more efficient to systematically search the space of such programs with a bias towards short and fast 
programs~\citep{Levin:73,Schmidhuber:97nn+,Schmidhuber:04oops}, 
instead of directly 
searching the huge space of possible NN weight matrices.
A previous 
universal language for encoding NNs  was assembler-like~\citep{Schmidhuber:97nn+}. More recent work uses more practical languages based on coefficients of popular transforms (Fourier, wavelet, etc). 
In particular, 
 RNN weight matrices may be compressed like images, by encoding them through the coefficients of a 
{\em discrete cosine transform} (DCT)~\citep{koutnik:gecco10,koutnik:gecco13}.
Compact DCT-based descriptions can be evolved through NES or CoSyNE
(Sec.~\ref{evorl}).
An RNN with over a million weights learned (without a teacher) to drive a simulated car 
in the TORCS driving game~\citep{wcci:torcs:09,torcs-manual:2011},
based on a high-dimensional video-like visual input stream~\citep{koutnik:gecco13}.
The RNN learned both control and visual processing from scratch, without being
aided by UL. (Of course, UL might help to generate more compact image codes
(Sec.~\ref{unsrl},~\ref{ul})
 to be fed into a smaller RNN, 
to reduce the overall computational effort.) 

\subsection{Universal RL}
\label{unirl}


{\em General purpose learning algorithms}
may improve themselves in open-ended fashion
and environment-specific 
ways in a lifelong learning 
context~\citep{schmidhuber87,Schmidhuber:97bias,Schmidhuber:97ssa,scholarpedia2010}. 
The most general type of RL is constrained only by the
fundamental limitations of computability identified by 
the founders of theoretical computer science 
\citep{Goedel:31,Church:36,Turing:36,Post:36}.
Remarkably, there exist blueprints of
 {\em universal problem solvers} or {\em universal RL machines}
for unlimited problem depth 
that are  
time-optimal in various theoretical senses~\citep{Hutter:05book+,Hutter:01fast+,Schmidhuber:02colt,Schmidhuber:05gmai}. 
In particular, the {\em G\"{o}del Machine} can be implemented 
on general computers such as RNNs and may improve 
any part of its software (including the learning algorithm itself)  
in a way that is provably time-optimal in a certain sense~\citep{Schmidhuber:05gmai}. It can be initialized by an 
{\em asymptotically optimal}
meta-method~\citep{Hutter:01fast+} (also applicable to RNNs)
which will solve any well-defined problem as quickly as the unknown fastest way of solving it, save for an additive constant overhead that becomes negligible as problem size grows. Note that most problems are large; only few are small. AI and DL researchers are still in business because many are interested in problems so small that it is worth trying to reduce the overhead through less general methods, including heuristics. Here I won't further discuss universal RL methods, which go beyond what is usually called DL. 

\section{Conclusion and Outlook}
\label{outlook}

{\em Deep Learning} (DL) in 
{\em Neural Networks} (NNs)
is relevant for
{\em Supervised Learning} (SL) (Sec.~\ref{super}),
{\em Unsupervised Learning} (UL) (Sec.~\ref{super}), and
{\em Reinforcement Learning} (RL) (Sec.~\ref{deeprl}).
By alleviating problems with deep {\em Credit Assignment Paths} (CAPs, Sec.~\ref{caps},~\ref{1991a}),
UL (Sec.~\ref{ulnn}) can not only facilitate 
SL of sequences  
(Sec.~\ref{1991b}) and 
stationary patterns
(Sec.~\ref{1987},~\ref{2006}), but also 
RL (Sec.~\ref{unsrl},~\ref{ul}). 
{\em Dynamic Programming} (DP, Sec.~\ref{dp}) is important for both 
deep SL (Sec.~\ref{1970})
and traditional RL with deep NNs (Sec.~\ref{trarl}).
A search for solution-computing, 
perturbation-resistant (Sec.~\ref{mdlnn},~\ref{2006},~\ref{tricks}),
low-complexity  NNs 
describable by few bits of information (Sec.~\ref{mdl}) can 
reduce overfitting and 
improve
deep SL \& UL (Sec.~\ref{mdlnn},~\ref{ulnn}) as well as RL (Sec.~\ref{comrl}),
also in the case of partially observable environments (Sec.~\ref{pomrl}). 
Deep SL, UL, RL often create hierarchies of more and more abstract  
representations of stationary data (Sec.~\ref{1965},~\ref{1987},~\ref{2006}),
sequential data (Sec.~\ref{1991b}), or RL policies (Sec.~\ref{subrl}). 
While UL can facilitate SL, pure SL for feedforward NNs (FNNs) (Sec.~\ref{1970},~\ref{1989},~\ref{2007},~\ref{2010})
and recurrent NNs (RNNs) (Sec.~\ref{1970},~\ref{1997})
did not only win early contests (Sec.~\ref{1994},~\ref{2003}) but also
most of the recent ones   
(Sec.~\ref{2009}--\ref{2013}).
Especially DL in FNNs profited from 
GPU implementations (Sec.~\ref{2007}--\ref{2011}).
In particular, 
GPU-based (Sec.~\ref{2011}) Max-Pooling (Sec.~\ref{1999}) Convolutional NNs (Sec.~\ref{1979},~\ref{1989},~\ref{2007})
won competitions not only in pattern recognition (Sec.~\ref{2011}--\ref{2013}) 
but also
image segmentation (Sec.~\ref{2012})
and object detection (Sec.~\ref{2012}, \ref{2013}).

Unlike these systems, humans {\em learn to actively perceive} patterns by sequentially directing attention to relevant parts of the available data. Near future deep NNs will do so, too, extending previous work since 1990 on NNs that learn selective attention through RL of (a) {\em motor actions} such as saccade control (Sec.~\ref{worrl}) and (b)  {\em internal actions} controlling spotlights of attention within RNNs, thus closing the general sensorimotor loop through both external and internal feedback (e.g., Sec.~\ref{notation},~\ref{2012},~\ref{evorl},~\ref{comrl}).

Many future deep NNs will also take into account that it costs energy
to activate neurons, and to send signals between them. Brains seem to
minimize such computational costs during problem solving in at least
two ways: (1) At a given time, only a small fraction of all neurons is
active because local competition through winner-take-all mechanisms
shuts down many neighbouring neurons, and only winners can activate
other neurons through outgoing connections (compare SLIM NNs;
Sec.~\ref{tricks}). (2) Numerous neurons are sparsely connected in a
compact 3D volume by many short-range and few long-range connections
(much like microchips in traditional supercomputers). Often
neighbouring neurons are allocated to solve a single task, thus
reducing communication costs.  Physics seems to dictate that any
efficient computational hardware will in the future also have to be
brain-like in keeping with these two constraints. The most successful
current deep RNNs, however, are not. Unlike certain spiking NNs
(Sec.~\ref{spiking}), they usually activate all units at least slightly,
and tend to be strongly connected, ignoring natural constraints of 3D
hardware. It should be possible to improve them by adopting (1) and
(2), and by minimizing non-differentiable energy and communication
costs through direct search in program (weight) space (e.g.,
Sec.~\ref{evorl},~\ref{comrl}).  These more brain-like RNNs will
allocate neighboring RNN parts to related behaviors, and distant RNN
parts to less related ones, thus self-modularizing in a way more
general than that of traditional self-organizing maps in FNNs
(Sec.~\ref{ulnn}). They will also implement Occam's razor
(Sec.~\ref{mdl},~\ref{mdlnn}) as a by-product of energy minimization,
by finding simple (highly generalizing) problem solutions that require
few active neurons and few, mostly short connections.

The more distant future may belong to general purpose learning
algorithms that improve themselves in provably optimal ways
(Sec.~\ref{unirl}), but these are not yet practical or commercially
relevant.

\section{Acknowledgments}
\label{ack}

Since 16 April 2014, drafts of this paper have undergone massive open online peer review through public mailing lists including 
{\em connectionists\-@cs.cmu.edu, ml-news\-@googlegroups.com, comp-neuro\-@neuro\-inf.org, genetic\_pro\-gramming\-@yahoo\-groups.com, rl-list\-@googlegroups.com, image\-world\-@diku.dk, Google+ machine learning forum.} 
Thanks to numerous NN / DL experts for valuable comments. Thanks to SNF, DFG, and the European Commission for partially funding my DL research group in the past quarter-century.
The contents of this paper may be used for educational and non-commercial purposes, including articles for Wikipedia and similar sites.

\end{sloppypar}

\bibliography{deep}

\begin{thebibliography}{}

\bibitem[Aberdeen, 2003]{aberdeenthesis}
Aberdeen, D. (2003).
\newblock {\em Policy-Gradient Algorithms for Partially Observable Markov
  Decision Processes}.
\newblock PhD thesis, Australian National University.

\bibitem[Abounadi et~al., 2002]{Abounadi:02}
Abounadi, J., Bertsekas, D., and Borkar, V.~S. (2002).
\newblock Learning algorithms for {Markov} decision processes with average
  cost.
\newblock {\em SIAM Journal on Control and Optimization}, 40(3):681--698.

\bibitem[Akaike, 1970]{Akaike:70}
Akaike, H. (1970).
\newblock Statistical predictor identification.
\newblock {\em Ann. Inst. Statist. Math.}, 22:203--217.

\bibitem[Akaike, 1973]{akaike1973}
Akaike, H. (1973).
\newblock Information theory and an extension of the maximum likelihood
  principle.
\newblock In {\em Second Intl. Symposium on Information Theory}, pages
  267--281. Akademinai Kiado.

\bibitem[Akaike, 1974]{akaike1974}
Akaike, H. (1974).
\newblock A new look at the statistical model identification.
\newblock {\em IEEE Transactions on Automatic Control}, 19(6):716--723.

\bibitem[Allender, 1992]{Allender:92}
Allender, A. (1992).
\newblock Application of time-bounded {Kolmogorov} complexity in complexity
  theory.
\newblock In Watanabe, O., editor, {\em {Kolmogorov} complexity and
  computational complexity}, pages 6--22. EATCS Monographs on Theoretical
  Computer Science, Springer.

\bibitem[Almeida, 1987]{Almeida:87}
Almeida, L.~B. (1987).
\newblock A learning rule for asynchronous perceptrons with feedback in a
  combinatorial environment.
\newblock In {\em IEEE 1st International Conference on Neural Networks, San
  Diego}, volume~2, pages 609--618.

\bibitem[Almeida et~al., 1997]{Almeida:97}
Almeida, L.~B., Almeida, L.~B., Langlois, T., Amaral, J.~D., and Redol, R.~A.
  (1997).
\newblock On-line step size adaptation.
\newblock Technical report, INESC, 9 Rua Alves Redol, 1000.

\bibitem[Amari, 1967]{Amari:1967:TAP}
Amari, S. (1967).
\newblock A theory of adaptive pattern classifiers.
\newblock {\em IEEE Trans. EC}, 16(3):299--307.

\bibitem[Amari et~al., 1996]{Amari:96}
Amari, S., Cichocki, A., and Yang, H. (1996).
\newblock A new learning algorithm for blind signal separation.
\newblock In Touretzky, D.~S., Mozer, M.~C., and Hasselmo, M.~E., editors, {\em
  Advances in Neural Information Processing Systems (NIPS)}, volume~8. The
  {MIT} Press.

\bibitem[Amari and Murata, 1993]{Amari:93}
Amari, S. and Murata, N. (1993).
\newblock Statistical theory of learning curves under entropic loss criterion.
\newblock {\em Neural Computation}, 5(1):140--153.

\bibitem[Amari, 1998]{amari1998natural}
Amari, S.-I. (1998).
\newblock Natural gradient works efficiently in learning.
\newblock {\em Neural Computation}, 10(2):251--276.

\bibitem[Amit and Brunel, 1997]{amit1997}
Amit, D.~J. and Brunel, N. (1997).
\newblock Dynamics of a recurrent network of spiking neurons before and
  following learning.
\newblock {\em Network: Computation in Neural Systems}, 8(4):373--404.

\bibitem[An, 1996]{an96}
An, G. (1996).
\newblock The effects of adding noise during backpropagation training on a
  generalization performance.
\newblock {\em Neural Computation}, 8(3):643--674.

\bibitem[Andrade et~al., 1993]{andrade1993}
Andrade, M.~A., Chacon, P., Merelo, J.~J., and Moran, F. (1993).
\newblock Evaluation of secondary structure of proteins from {UV} circular
  dichroism spectra using an unsupervised learning neural network.
\newblock {\em Protein Engineering}, 6(4):383--390.

\bibitem[Andrews et~al., 1995]{andrews1995}
Andrews, R., Diederich, J., and Tickle, A.~B. (1995).
\newblock Survey and critique of techniques for extracting rules from trained
  artificial neural networks.
\newblock {\em Knowledge-Based Systems}, 8(6):373--389.

\bibitem[Anguita and Gomes, 1996]{anguita1996}
Anguita, D. and Gomes, B.~A. (1996).
\newblock Mixing floating- and fixed-point formats for neural network learning
  on neuroprocessors.
\newblock {\em Microprocessing and Microprogramming}, 41(10):757 -- 769.

\bibitem[Anguita et~al., 1994]{anguita1994}
Anguita, D., Parodi, G., and Zunino, R. (1994).
\newblock An efficient implementation of {BP} on {RISC}-based workstations.
\newblock {\em Neurocomputing}, 6(1):57 -- 65.

\bibitem[Arel et~al., 2010]{itamar2010}
Arel, I., Rose, D.~C., and Karnowski, T.~P. (2010).
\newblock Deep machine learning -- a new frontier in artificial intelligence
  research.
\newblock {\em Computational Intelligence Magazine, IEEE}, 5(4):13--18.

\bibitem[Ash, 1989]{Ash:89}
Ash, T. (1989).
\newblock Dynamic node creation in backpropagation neural networks.
\newblock {\em Connection Science}, 1(4):365--375.

\bibitem[Atick et~al., 1992]{Atick:92}
Atick, J.~J., Li, Z., and Redlich, A.~N. (1992).
\newblock Understanding retinal color coding from first principles.
\newblock {\em Neural Computation}, 4:559--572.

\bibitem[Atiya and Parlos, 2000]{atiya2000}
Atiya, A.~F. and Parlos, A.~G. (2000).
\newblock New results on recurrent network training: unifying the algorithms
  and accelerating convergence.
\newblock {\em IEEE Transactions on Neural Networks}, 11(3):697--709.

\bibitem[Ba and Frey, 2013]{frey2013}
Ba, J. and Frey, B. (2013).
\newblock Adaptive dropout for training deep neural networks.
\newblock In {\em Advances in Neural Information Processing Systems (NIPS)},
  pages 3084--3092.

\bibitem[Baird, 1990]{Baird90}
Baird, H. (1990).
\newblock Document image defect models.
\newblock In {\em Proceddings, IAPR Workshop on Syntactic and Structural
  Pattern Recognition}, Murray Hill, NJ.

\bibitem[Baird and Moore, 1999]{baird:nips12}
Baird, L. and Moore, A.~W. (1999).
\newblock Gradient descent for general reinforcement learning.
\newblock In {\em Advances in neural information processing systems 12
  {(NIPS)}}, pages 968--974. MIT Press.

\bibitem[Baird, 1995]{Baird:95}
Baird, L.~C. (1995).
\newblock Residual algorithms: Reinforcement learning with function
  approximation.
\newblock In {\em International Conference on Machine Learning}, pages 30--37.

\bibitem[Bakker, 2002]{Bakker:01nips}
Bakker, B. (2002).
\newblock Reinforcement learning with {L}ong {S}hort-{T}erm {M}emory.
\newblock In Dietterich, T.~G., Becker, S., and Ghahramani, Z., editors, {\em
  Advances in {N}eural {I}nformation {P}rocessing {S}ystems 14}, pages
  1475--1482. MIT Press, Cambridge, MA.

\bibitem[Bakker and Schmidhuber, 2004]{Bakker:04ias}
Bakker, B. and Schmidhuber, J. (2004).
\newblock Hierarchical reinforcement learning based on subgoal discovery and
  subpolicy specialization.
\newblock In et~al., F.~G., editor, {\em Proc. 8th Conference on Intelligent
  Autonomous Systems IAS-8}, pages 438--445, Amsterdam, NL. IOS Press.

\bibitem[Bakker et~al., 2003]{Bakker:03robot}
Bakker, B., Zhumatiy, V., Gruener, G., and Schmidhuber, J. (2003).
\newblock A robot that reinforcement-learns to identify and memorize important
  previous observations.
\newblock In {\em Proceedings of the 2003 IEEE/RSJ International Conference on
  Intelligent Robots and Systems, IROS 2003}, pages 430--435.

\bibitem[Baldi, 1995]{baldi95}
Baldi, P. (1995).
\newblock Gradient descent learning algorithms overview: A general dynamical
  systems perspective.
\newblock {\em IEEE Transactions on Neural Networks}, 6(1):182--195.

\bibitem[Baldi, 2012]{baldijmlr12}
Baldi, P. (2012).
\newblock Autoencoders, unsupervised learning, and deep architectures.
\newblock {\em Journal of Machine Learning Research (Proc. 2011 ICML Workshop
  on Unsupervised and Transfer Learning)}, 27:37--50.

\bibitem[Baldi et~al., 1999]{baldi99exploiting}
Baldi, P., Brunak, S., Frasconi, P., Pollastri, G., and Soda, G. (1999).
\newblock Exploiting the past and the future in protein secondary structure
  prediction.
\newblock {\em Bioinformatics}, 15:937--946.

\bibitem[Baldi and Chauvin, 1993]{baldi93finger}
Baldi, P. and Chauvin, Y. (1993).
\newblock Neural networks for fingerprint recognition.
\newblock {\em Neural Computation}, 5(3):402--418.

\bibitem[Baldi and Chauvin, 1996]{baldichauvin96}
Baldi, P. and Chauvin, Y. (1996).
\newblock Hybrid modeling, {HMM/NN} architectures, and protein applications.
\newblock {\em Neural Computation}, 8(7):1541--1565.

\bibitem[Baldi and Hornik, 1989]{Baldi:89}
Baldi, P. and Hornik, K. (1989).
\newblock Neural networks and principal component analysis: Learning from
  examples without local minima.
\newblock {\em Neural Networks}, 2:53--58.

\bibitem[Baldi and Hornik, 1994]{baldihornik95}
Baldi, P. and Hornik, K. (1994).
\newblock Learning in linear networks: a survey.
\newblock {\em IEEE Transactions on Neural Networks}, 6(4):837--858.
\newblock 1995.

\bibitem[Baldi and Pollastri, 2003]{baldi2003jmlr}
Baldi, P. and Pollastri, G. (2003).
\newblock The principled design of large-scale recursive neural network
  architectures -- {DAG-RNN}s and the protein structure prediction problem.
\newblock {\em J. Mach. Learn. Res.}, 4:575--602.

\bibitem[Baldi and Sadowski, 2014]{baldidropout2014}
Baldi, P. and Sadowski, P. (2014).
\newblock The dropout learning algorithm.
\newblock {\em Artificial Intelligence}, 210C:78--122.

\bibitem[Ballard, 1987]{ballard1987modular}
Ballard, D.~H. (1987).
\newblock Modular learning in neural networks.
\newblock In {\em Proc. AAAI}, pages 279--284.

\bibitem[Baluja, 1994]{Baluja:94}
Baluja, S. (1994).
\newblock Population-based incremental learning: A method for integrating
  genetic search based function optimization and competitive learning.
\newblock Technical Report CMU-CS-94-163, Carnegie Mellon University.

\bibitem[Balzer, 1985]{balzer1985}
Balzer, R. (1985).
\newblock A 15 year perspective on automatic programming.
\newblock {\em IEEE Transactions on Software Engineering}, 11(11):1257--1268.

\bibitem[Barlow, 1989]{Barlow:89review}
Barlow, H.~B. (1989).
\newblock Unsupervised learning.
\newblock {\em Neural Computation}, 1(3):295--311.

\bibitem[Barlow et~al., 1989]{Barlow:89}
Barlow, H.~B., Kaushal, T.~P., and Mitchison, G.~J. (1989).
\newblock Finding minimum entropy codes.
\newblock {\em Neural Computation}, 1(3):412--423.

\bibitem[Barrow, 1987]{Barrow:87}
Barrow, H.~G. (1987).
\newblock Learning receptive fields.
\newblock In {\em Proceedings of the IEEE 1st Annual Conference on Neural
  Networks}, volume~IV, pages 115--121. IEEE.

\bibitem[Barto and Mahadevan, 2003]{barto2003hrl}
Barto, A.~G. and Mahadevan, S. (2003).
\newblock Recent advances in hierarchical reinforcement learning.
\newblock {\em Discrete Event Dynamic Systems}, 13(4):341--379.

\bibitem[Barto et~al., 2004]{Barto:04}
Barto, A.~G., Singh, S., and Chentanez, N. (2004).
\newblock Intrinsically motivated learning of hierarchical collections of
  skills.
\newblock In {\em Proceedings of International Conference on Developmental
  Learning (ICDL)}, pages 112--119. MIT Press, Cambridge, MA.

\bibitem[Barto et~al., 1983]{BartoSuttonAnderson:83}
Barto, A.~G., Sutton, R.~S., and Anderson, C.~W. (1983).
\newblock Neuronlike adaptive elements that can solve difficult learning
  control problems.
\newblock {\em IEEE Transactions on Systems, Man, and Cybernetics},
  SMC-13:834--846.

\bibitem[Battiti, 1989]{Battiti:89}
Battiti, R. (1989).
\newblock Accelerated backpropagation learning: two optimization methods.
\newblock {\em Complex Systems}, 3(4):331--342.

\bibitem[Battiti, 1992]{Battiti:92}
Battiti, T. (1992).
\newblock First- and second-order methods for learning: Between steepest
  descent and {N}ewton's method.
\newblock {\em Neural Computation}, 4(2):141--166.

\bibitem[Baum and Haussler, 1989]{BaumHaussler:89}
Baum, E.~B. and Haussler, D. (1989).
\newblock What size net gives valid generalization?
\newblock {\em Neural Computation}, 1(1):151--160.

\bibitem[Baum and Petrie, 1966]{baum1966}
Baum, L.~E. and Petrie, T. (1966).
\newblock Statistical inference for probabilistic functions of finite state
  {Markov} chains.
\newblock {\em The Annals of Mathematical Statistics}, pages 1554--1563.

\bibitem[Baxter and Bartlett, 2001]{baxter2001}
Baxter, J. and Bartlett, P.~L. (2001).
\newblock Infinite-horizon policy-gradient estimation.
\newblock {\em J. Artif. Int. Res.}, 15(1):319--350.

\bibitem[Bayer and Osendorfer, 2014]{bayer2014variational}
Bayer, J. and Osendorfer, C. (2014).
\newblock Variational inference of latent state sequences using recurrent
  networks.
\newblock {\em arXiv preprint arXiv:1406.1655}.

\bibitem[Bayer et~al., 2013]{bayer2013fast}
Bayer, J., Osendorfer, C., Chen, N., Urban, S., and van~der Smagt, P. (2013).
\newblock On fast dropout and its applicability to recurrent networks.
\newblock {\em arXiv preprint arXiv:1311.0701}.

\bibitem[Bayer et~al., 2009]{DBLP:conf/icann/BayerWTS09}
Bayer, J., Wierstra, D., Togelius, J., and Schmidhuber, J. (2009).
\newblock Evolving memory cell structures for sequence learning.
\newblock In {\em Proc. ICANN (2)}, pages 755--764.

\bibitem[Bayes, 1763]{bayes1763}
Bayes, T. (1763).
\newblock An essay toward solving a problem in the doctrine of chances.
\newblock {\em Philosophical Transactions of the Royal Society of London},
  53:370--418.
\newblock Communicated by R. Price, in a letter to J. Canton.

\bibitem[Becker, 1991]{Becker:91}
Becker, S. (1991).
\newblock Unsupervised learning procedures for neural networks.
\newblock {\em International Journal of Neural Systems}, 2(1 \& 2):17--33.

\bibitem[Becker and Le~Cun, 1989]{Becker:89}
Becker, S. and Le~Cun, Y. (1989).
\newblock Improving the convergence of back-propagation learning with second
  order methods.
\newblock In Touretzky, D., Hinton, G., and Sejnowski, T., editors, {\em Proc.
  1988 Connectionist Models Summer School}, pages 29--37, Pittsburg 1988.
  Morgan Kaufmann, San Mateo.

\bibitem[Behnke, 1999]{Behnke:IJCNN1999}
Behnke, S. (1999).
\newblock Hebbian learning and competition in the neural abstraction pyramid.
\newblock In {\em Proceedings of the International Joint Conference on Neural
  Networks (IJCNN)}, volume~2, pages 1356--1361.

\bibitem[Behnke, 2001]{Behnke:IJCIA2001}
Behnke, S. (2001).
\newblock Learning iterative image reconstruction in the neural abstraction
  pyramid.
\newblock {\em International Journal of Computational Intelligence and
  Applications}, 1(4):427--438.

\bibitem[Behnke, 2002]{Behnke:ICANN2002}
Behnke, S. (2002).
\newblock Learning face localization using hierarchical recurrent networks.
\newblock In {\em Proceedings of the 12th International Conference on
  Artificial Neural Networks (ICANN), Madrid, Spain}, pages 1319--1324.

\bibitem[Behnke, 2003a]{Behnke:IJCNN2003}
Behnke, S. (2003a).
\newblock Discovering hierarchical speech features using convolutional
  non-negative matrix factorization.
\newblock In {\em Proceedings of the International Joint Conference on Neural
  Networks (IJCNN)}, volume~4, pages 2758--2763.

\bibitem[Behnke, 2003b]{Behnke:LNCS}
Behnke, S. (2003b).
\newblock {\em Hierarchical Neural Networks for Image Interpretation}, volume
  LNCS 2766 of {\em Lecture Notes in Computer Science}.
\newblock Springer.

\bibitem[Behnke, 2005]{Behnke:NCA}
Behnke, S. (2005).
\newblock Face localization and tracking in the {N}eural {A}bstraction
  {P}yramid.
\newblock {\em Neural Computing and Applications}, 14(2):97--103.

\bibitem[Behnke and Rojas, 1998]{Behnke:IJCNN1998}
Behnke, S. and Rojas, R. (1998).
\newblock Neural abstraction pyramid: A hierarchical image understanding
  architecture.
\newblock In {\em Proceedings of International Joint Conference on Neural
  Networks (IJCNN)}, volume~2, pages 820--825.

\bibitem[Bell and Sejnowski, 1995]{Bell:95}
Bell, A.~J. and Sejnowski, T.~J. (1995).
\newblock An information-maximization approach to blind separation and blind
  deconvolution.
\newblock {\em Neural Computation}, 7(6):1129--1159.

\bibitem[Bellman, 1957]{Bellman:1957}
Bellman, R. (1957).
\newblock {\em Dynamic Programming}.
\newblock Princeton University Press, Princeton, NJ, USA, 1st edition.

\bibitem[Belouchrani et~al., 1997]{belouchrani1997}
Belouchrani, A., Abed-Meraim, K., Cardoso, J.-F., and Moulines, E. (1997).
\newblock A blind source separation technique using second-order statistics.
\newblock {\em IEEE Transactions on Signal Processing}, 45(2):434--444.

\bibitem[Bengio, 1991]{bengio91}
Bengio, Y. (1991).
\newblock {\em Artificial Neural Networks and their Application to Sequence
  Recognition}.
\newblock PhD thesis, McGill University, (Computer Science), Montreal, Qc.,
  Canada.

\bibitem[Bengio, 2009]{bengio09}
Bengio, Y. (2009).
\newblock {\em Learning Deep Architectures for AI. Foundations and Trends in
  Machine Learning, V2(1)}.
\newblock Now Publishers.

\bibitem[Bengio et~al., 2013]{bengio2013tpami}
Bengio, Y., Courville, A., and Vincent, P. (2013).
\newblock Representation learning: A review and new perspectives.
\newblock {\em Pattern Analysis and Machine Intelligence, IEEE Transactions
  on}, 35(8):1798--1828.

\bibitem[Bengio et~al., 2007]{bengio2006}
Bengio, Y., Lamblin, P., Popovici, D., and Larochelle, H. (2007).
\newblock Greedy layer-wise training of deep networks.
\newblock In Cowan, J.~D., Tesauro, G., and Alspector, J., editors, {\em
  Advances in Neural Information Processing Systems 19 (NIPS)}, pages 153--160.
  MIT Press.

\bibitem[Bengio et~al., 1994]{Bengio:94}
Bengio, Y., Simard, P., and Frasconi, P. (1994).
\newblock Learning long-term dependencies with gradient descent is difficult.
\newblock {\em IEEE Transactions on Neural Networks}, 5(2):157--166.

\bibitem[Beringer et~al., 2005]{beringer:05icann}
Beringer, N., Graves, A., Schiel, F., and Schmidhuber, J. (2005).
\newblock Classifying unprompted speech by retraining {LSTM} nets.
\newblock In Duch, W., Kacprzyk, J., Oja, E., and Zadrozny, S., editors, {\em
  Artificial Neural Networks: Biological Inspirations - ICANN 2005, LNCS 3696},
  pages 575--581. Springer-Verlag Berlin Heidelberg.

\bibitem[Bertsekas, 2001]{Bertsekas:01}
Bertsekas, D.~P. (2001).
\newblock {\em Dynamic Programming and Optimal Control}.
\newblock Athena Scientific.

\bibitem[Bertsekas and Tsitsiklis, 1996]{Bertsekas:96}
Bertsekas, D.~P. and Tsitsiklis, J.~N. (1996).
\newblock {\em Neuro-dynamic Programming}.
\newblock Athena Scientific, Belmont, MA.

\bibitem[Bichot et~al., 2005]{Bichot:05}
Bichot, N.~P., Rossi, A.~F., and Desimone, R. (2005).
\newblock Parallel and serial neural mechanisms for visual search in macaque
  area {V4}.
\newblock {\em Science}, 308:529--534.

\bibitem[Biegler-K{\"o}nig and B{\"a}rmann, 1993]{Biegler:93}
Biegler-K{\"o}nig, F. and B{\"a}rmann, F. (1993).
\newblock A learning algorithm for multilayered neural networks based on linear
  least squares problems.
\newblock {\em Neural Networks}, 6(1):127--131.

\bibitem[Bishop, 1993]{Bishop:93}
Bishop, C.~M. (1993).
\newblock Curvature-driven smoothing: A learning algorithm for feed-forward
  networks.
\newblock {\em IEEE Transactions on Neural Networks}, 4(5):882--884.

\bibitem[Bishop, 2006]{bishop:2006}
Bishop, C.~M. (2006).
\newblock {\em Pattern Recognition and Machine Learning}.
\newblock Springer.

\bibitem[Blair and Pollack, 1997]{Blair+Pollack:1997nc}
Blair, A.~D. and Pollack, J.~B. (1997).
\newblock Analysis of dynamical recognizers.
\newblock {\em Neural Computation}, 9(5):1127--1142.

\bibitem[Blondel and Tsitsiklis, 2000]{blondel2000}
Blondel, V.~D. and Tsitsiklis, J.~N. (2000).
\newblock A survey of computational complexity results in systems and control.
\newblock {\em Automatica}, 36(9):1249--1274.

\bibitem[Bluche et~al., 2014]{bluche13}
Bluche, T., Louradour, J., Knibbe, M., Moysset, B., Benzeghiba, F., and
  Kermorvant, C. (2014).
\newblock {The A2iA Arabic Handwritten Text Recognition System at the
  OpenHaRT2013 Evaluation}.
\newblock In {\em International Workshop on Document Analysis Systems}.

\bibitem[Blum and Rivest, 1992]{blum1992}
Blum, A.~L. and Rivest, R.~L. (1992).
\newblock Training a 3-node neural network is np-complete.
\newblock {\em Neural Networks}, 5(1):117--127.

\bibitem[Blumer et~al., 1987]{Blumer:87}
Blumer, A., Ehrenfeucht, A., Haussler, D., and Warmuth, M.~K. (1987).
\newblock Occam's razor.
\newblock {\em Information Processing Letters}, 24:377--380.

\bibitem[Bobrowski, 1978]{bobrowski78}
Bobrowski, L. (1978).
\newblock Learning processes in multilayer threshold nets.
\newblock {\em Biological Cybernetics}, 31:1--6.

\bibitem[Bod{\'e}n and Wiles, 2000]{boden00context-free}
Bod{\'e}n, M. and Wiles, J. (2000).
\newblock Context-free and context-sensitive dynamics in recurrent neural
  networks.
\newblock {\em Connection Science}, 12(3-4):197--210.

\bibitem[Bodenhausen and Waibel, 1991]{Bodenhausen:91}
Bodenhausen, U. and Waibel, A. (1991).
\newblock The {Tempo 2} algorithm: Adjusting time-delays by supervised
  learning.
\newblock In Lippman, D.~S., Moody, J.~E., and Touretzky, D.~S., editors, {\em
  Advances in Neural Information Processing Systems 3}, pages 155--161. Morgan
  Kaufmann.

\bibitem[Bohte et~al., 2002]{bohte2002}
Bohte, S.~M., Kok, J.~N., and La~Poutre, H. (2002).
\newblock Error-backpropagation in temporally encoded networks of spiking
  neurons.
\newblock {\em Neurocomputing}, 48(1):17--37.

\bibitem[Boltzmann, 1909]{boltzmann1909}
Boltzmann, L. (1909).
\newblock In Hasen\"{o}hrl, F., editor, {\em {Wissenschaftliche Abhandlungen}
  (collection of {Boltzmann's} articles in scientific journals)}. Barth,
  Leipzig.

\bibitem[Bottou, 1991]{bottou91}
Bottou, L. (1991).
\newblock {\em Une approche th\'eorique de l'apprentissage connexioniste;
  applications \`a la reconnaissance de la parole}.
\newblock PhD thesis, Universit\'e de Paris XI.

\bibitem[Bourlard and Morgan, 1994]{bourlard+morgan:1994}
Bourlard, H. and Morgan, N. (1994).
\newblock {\em Connnectionist Speech Recognition: A Hybrid Approach}.
\newblock Kluwer Academic Publishers.

\bibitem[Boutilier and Poole, 1996]{Boutilier:96}
Boutilier, C. and Poole, D. (1996).
\newblock Computing optimal policies for partially observable {Markov} decision
  processes using compact representations.
\newblock In {\em Proceedings of the AAAI, Portland, OR}.

\bibitem[Bradtke et~al., 1996]{96-BradtkeLstd}
Bradtke, S.~J., Barto, A.~G., and Kaelbling, L.~P. (1996).
\newblock Linear least-squares algorithms for temporal difference learning.
\newblock In {\em Machine Learning}, pages 22--33.

\bibitem[Brafman and Tennenholtz, 2002]{brafman02}
Brafman, R.~I. and Tennenholtz, M. (2002).
\newblock {R-MAX}---a general polynomial time algorithm for near-optimal
  reinforcement learning.
\newblock {\em Journal of Machine Learning Research}, 3:213--231.

\bibitem[Brea et~al., 2013]{brea2013}
Brea, J., Senn, W., and Pfister, J.-P. (2013).
\newblock Matching recall and storage in sequence learning with spiking neural
  networks.
\newblock {\em The Journal of Neuroscience}, 33(23):9565--9575.

\bibitem[Breiman, 1996]{breiman:1996}
Breiman, L. (1996).
\newblock Bagging predictors.
\newblock {\em Machine Learning}, 24:123--140.

\bibitem[Brette et~al., 2007]{brette2007}
Brette, R., Rudolph, M., Carnevale, T., Hines, M., Beeman, D., Bower, J.~M.,
  Diesmann, M., Morrison, A., Goodman, P.~H., Harris~Jr, F.~C., et~al. (2007).
\newblock Simulation of networks of spiking neurons: a review of tools and
  strategies.
\newblock {\em Journal of Computational Neuroscience}, 23(3):349--398.

\bibitem[Breuel et~al., 2013]{breuel2013high}
Breuel, T.~M., Ul-Hasan, A., Al-Azawi, M.~A., and Shafait, F. (2013).
\newblock High-performance {OCR} for printed {English} and {Fraktur} using
  {LSTM} networks.
\newblock In {\em 12th International Conference on Document Analysis and
  Recognition (ICDAR)}, pages 683--687. IEEE.

\bibitem[Bromley et~al., 1993]{bromley-93}
Bromley, J., Bentz, J.~W., Bottou, L., Guyon, I., LeCun, Y., Moore, C.,
  Sackinger, E., and Shah, R. (1993).
\newblock Signature verification using a {Siamese} time delay neural network.
\newblock {\em International Journal of Pattern Recognition and Artificial
  Intelligence}, 7(4):669--688.

\bibitem[Broyden et~al., 1965]{broyden1965}
Broyden, C.~G. et~al. (1965).
\newblock A class of methods for solving nonlinear simultaneous equations.
\newblock {\em Math. Comp}, 19(92):577--593.

\bibitem[Brueckner and Schulter, 2014]{brueckner2014}
Brueckner, R. and Schulter, B. (2014).
\newblock Social signal classification using deep {BLSTM} recurrent neural
  networks.
\newblock In {\em Proceedings 39th IEEE International Conference on Acoustics,
  Speech, and Signal Processing, ICASSP 2014, Florence, Italy}, pages
  4856--4860.

\bibitem[Brunel, 2000]{brunel2000}
Brunel, N. (2000).
\newblock Dynamics of sparsely connected networks of excitatory and inhibitory
  spiking neurons.
\newblock {\em Journal of Computational Neuroscience}, 8(3):183--208.

\bibitem[Bryson and Ho, 1969]{bryson1969applied}
Bryson, A. and Ho, Y. (1969).
\newblock {\em Applied optimal control: optimization, estimation, and control}.
\newblock Blaisdell Pub. Co.

\bibitem[Bryson, 1961]{bryson:1961}
Bryson, A.~E. (1961).
\newblock A gradient method for optimizing multi-stage allocation processes.
\newblock In {\em Proc. Harvard Univ. Symposium on digital computers and their
  applications}.

\bibitem[Bryson and Denham, 1961]{BRYSON-DENHAM-61A}
Bryson, Jr., A.~E. and Denham, W.~F. (1961).
\newblock A steepest-ascent method for solving optimum programming problems.
\newblock Technical Report BR-1303, Raytheon Company, Missle and Space
  Division.

\bibitem[Buhler, 2001]{buhler2001}
Buhler, J. (2001).
\newblock Efficient large-scale sequence comparison by locality-sensitive
  hashing.
\newblock {\em Bioinformatics}, 17(5):419--428.

\bibitem[Buntine and Weigend, 1991]{Buntine:91}
Buntine, W.~L. and Weigend, A.~S. (1991).
\newblock {Bayesian} back-propagation.
\newblock {\em Complex Systems}, 5:603--643.

\bibitem[Burgess, 1994]{burgess1994}
Burgess, N. (1994).
\newblock A constructive algorithm that converges for real-valued input
  patterns.
\newblock {\em International Journal of Neural Systems}, 5(1):59--66.

\bibitem[Cardoso, 1994]{Cardoso:94}
Cardoso, J.-F. (1994).
\newblock On the performance of orthogonal source separation algorithms.
\newblock In {\em Proc. {EUSIPCO}}, pages 776--779.

\bibitem[Carreira-Perpinan, 2001]{carreira2001}
Carreira-Perpinan, M.~A. (2001).
\newblock {\em Continuous latent variable models for dimensionality reduction
  and sequential data reconstruction}.
\newblock PhD thesis, University of Sheffield UK.

\bibitem[Carter et~al., 1990]{Carter:90}
Carter, M.~J., Rudolph, F.~J., and Nucci, A.~J. (1990).
\newblock Operational fault tolerance of {CMAC} networks.
\newblock In Touretzky, D.~S., editor, {\em Advances in Neural Information
  Processing Systems (NIPS) 2}, pages 340--347. San Mateo, CA: Morgan Kaufmann.

\bibitem[Caruana, 1997]{caruana1997}
Caruana, R. (1997).
\newblock Multitask learning.
\newblock {\em Machine Learning}, 28(1):41--75.

\bibitem[Casey, 1996]{casey96dynamics}
Casey, M.~P. (1996).
\newblock The dynamics of discrete-time computation, with application to
  recurrent neural networks and finite state machine extraction.
\newblock {\em Neural Computation}, 8(6):1135--1178.

\bibitem[Cauwenberghs, 1993]{Cauwenberghs:93}
Cauwenberghs, G. (1993).
\newblock A fast stochastic error-descent algorithm for supervised learning and
  optimization.
\newblock In Lippman, D.~S., Moody, J.~E., and Touretzky, D.~S., editors, {\em
  Advances in Neural Information Processing Systems 5}, pages 244--244. Morgan
  Kaufmann.

\bibitem[Chaitin, 1966]{Chaitin:66}
Chaitin, G.~J. (1966).
\newblock On the length of programs for computing finite binary sequences.
\newblock {\em Journal of the ACM}, 13:547--569.

\bibitem[Chalup and Blair, 2003]{ChalupBlairNN2003}
Chalup, S.~K. and Blair, A.~D. (2003).
\newblock Incremental training of first order recurrent neural networks to
  predict a context-sensitive language.
\newblock {\em Neural Networks}, 16(7):955--972.

\bibitem[Chellapilla et~al., 2006]{chellapilla:2006b}
Chellapilla, K., Puri, S., and Simard, P. (2006).
\newblock High performance convolutional neural networks for document
  processing.
\newblock In {\em International Workshop on Frontiers in Handwriting
  Recognition}.

\bibitem[Chen and Salman, 2011]{chen2011ieeetnn}
Chen, K. and Salman, A. (2011).
\newblock Learning speaker-specific characteristics with a deep neural
  architecture.
\newblock {\em IEEE Transactions on Neural Networks}, 22(11):1744--1756.

\bibitem[Cho, 2014]{cho2014}
Cho, K. (2014).
\newblock {\em Foundations and Advances in Deep Learning}.
\newblock PhD thesis, Aalto University School of Science.

\bibitem[Cho et~al., 2012]{cho2012}
Cho, K., Ilin, A., and Raiko, T. (2012).
\newblock Tikhonov-type regularization for restricted {Boltzmann} machines.
\newblock In {\em Intl. Conf. on Artificial Neural Networks (ICANN) 2012},
  pages 81--88. Springer.

\bibitem[Cho et~al., 2013]{cho2013}
Cho, K., Raiko, T., and Ilin, A. (2013).
\newblock Enhanced gradient for training restricted {Boltzmann} machines.
\newblock {\em Neural Computation}, 25(3):805--831.

\bibitem[Church, 1936]{Church:36}
Church, A. (1936).
\newblock An unsolvable problem of elementary number theory.
\newblock {\em American Journal of Mathematics}, 58:345--363.

\bibitem[Ciresan et~al., 2012a]{ciresan2012nips}
Ciresan, D.~C., Giusti, A., Gambardella, L.~M., and Schmidhuber, J. (2012a).
\newblock Deep neural networks segment neuronal membranes in electron
  microscopy images.
\newblock In {\em Advances in Neural Information Processing Systems {(NIPS)}},
  pages 2852--2860.

\bibitem[Ciresan et~al., 2013]{miccai2013}
Ciresan, D.~C., Giusti, A., Gambardella, L.~M., and Schmidhuber, J. (2013).
\newblock Mitosis detection in breast cancer histology images with deep neural
  networks.
\newblock In {\em Proc. MICCAI}, volume~2, pages 411--418.

\bibitem[Ciresan et~al., 2010]{ciresan:2010}
Ciresan, D.~C., Meier, U., Gambardella, L.~M., and Schmidhuber, J. (2010).
\newblock Deep big simple neural nets for handwritten digit recogntion.
\newblock {\em Neural Computation}, 22(12):3207--3220.

\bibitem[Ciresan et~al., 2011a]{ciresan:2011ijcai}
Ciresan, D.~C., Meier, U., Masci, J., Gambardella, L.~M., and Schmidhuber, J.
  (2011a).
\newblock Flexible, high performance convolutional neural networks for image
  classification.
\newblock In {\em Intl. Joint Conference on Artificial Intelligence IJCAI},
  pages 1237--1242.

\bibitem[Ciresan et~al., 2011b]{ciresan:2011ijcnn}
Ciresan, D.~C., Meier, U., Masci, J., and Schmidhuber, J. (2011b).
\newblock A committee of neural networks for traffic sign classification.
\newblock In {\em International Joint Conference on Neural Networks (IJCNN)},
  pages 1918--1921.

\bibitem[Ciresan et~al., 2012b]{ciresan:2012NN}
Ciresan, D.~C., Meier, U., Masci, J., and Schmidhuber, J. (2012b).
\newblock Multi-column deep neural network for traffic sign classification.
\newblock {\em Neural Networks}, 32:333--338.

\bibitem[Ciresan et~al., 2012c]{ciresan2012cvpr}
Ciresan, D.~C., Meier, U., and Schmidhuber, J. (2012c).
\newblock Multi-column deep neural networks for image classification.
\newblock In {\em IEEE Conference on Computer Vision and Pattern Recognition
  CVPR 2012}.
\newblock Long preprint arXiv:1202.2745v1 [cs.CV].

\bibitem[Ciresan et~al., 2012d]{Ciresan:2012a}
Ciresan, D.~C., Meier, U., and Schmidhuber, J. (2012d).
\newblock Transfer learning for {Latin} and {Chinese} characters with deep
  neural networks.
\newblock In {\em International Joint Conference on Neural Networks (IJCNN)},
  pages 1301--1306.

\bibitem[Ciresan and Schmidhuber, 2013]{chinese2013}
Ciresan, D.~C. and Schmidhuber, J. (2013).
\newblock Multi-column deep neural networks for offline handwritten {Chinese}
  character classification.
\newblock Technical report, IDSIA.
\newblock arXiv:1309.0261.

\bibitem[Cliff et~al., 1993]{cliff1993}
Cliff, D.~T., Husbands, P., and Harvey, I. (1993).
\newblock Evolving recurrent dynamical networks for robot control.
\newblock In {\em Artificial Neural Nets and Genetic Algorithms}, pages
  428--435. Springer.

\bibitem[Clune et~al., 2013]{clune2013modular}
Clune, J., Mouret, J.-B., and Lipson, H. (2013).
\newblock The evolutionary origins of modularity.
\newblock {\em Proceedings of the Royal Society B: Biological Sciences},
  280(1755):20122863.

\bibitem[Clune et~al., 2011]{clune2011}
Clune, J., Stanley, K.~O., Pennock, R.~T., and Ofria, C. (2011).
\newblock On the performance of indirect encoding across the continuum of
  regularity.
\newblock {\em Trans. Evol. Comp}, 15(3):346--367.

\bibitem[Coates et~al., 2013]{coates:2013icml}
Coates, A., Huval, B., Wang, T., Wu, D.~J., Ng, A.~Y., and Catanzaro, B.
  (2013).
\newblock Deep learning with {COTS HPC} systems.
\newblock In {\em Proc. International Conference on Machine learning
  (ICML'13)}.

\bibitem[Cochocki and Unbehauen, 1993]{cochocki1993}
Cochocki, A. and Unbehauen, R. (1993).
\newblock {\em Neural networks for optimization and signal processing}.
\newblock John Wiley \& Sons, Inc.

\bibitem[Collobert and Weston, 2008]{weston2008}
Collobert, R. and Weston, J. (2008).
\newblock A unified architecture for natural language processing: Deep neural
  networks with multitask learning.
\newblock In {\em Proceedings of the 25th International Conference on Machine
  Learning (ICML)}, pages 160--167. ACM.

\bibitem[Comon, 1994]{Comon:94}
Comon, P. (1994).
\newblock Independent component analysis -- a new concept?
\newblock {\em Signal Processing}, 36(3):287--314.

\bibitem[Connor et~al., 2007]{connor2007}
Connor, C.~E., Brincat, S.~L., and Pasupathy, A. (2007).
\newblock Transformation of shape information in the ventral pathway.
\newblock {\em Current Opinion in Neurobiology}, 17(2):140--147.

\bibitem[Connor et~al., 1994]{connor1994}
Connor, J., Martin, D.~R., and Atlas, L.~E. (1994).
\newblock Recurrent neural networks and robust time series prediction.
\newblock {\em IEEE Transactions on Neural Networks}, 5(2):240--254.

\bibitem[Cook, 1971]{Cook:71}
Cook, S.~A. (1971).
\newblock The complexity of theorem-proving procedures.
\newblock In {\em Proceedings of the 3rd Annual ACM Symposium on the Theory of
  Computing (STOC'71)}, pages 151--158. ACM, New York.

\bibitem[Cramer, 1985]{Cramer:85}
Cramer, N.~L. (1985).
\newblock A representation for the adaptive generation of simple sequential
  programs.
\newblock In Grefenstette, J., editor, {\em Proceedings of an International
  Conference on Genetic Algorithms and Their Applications, Carnegie-Mellon
  University, July 24-26, 1985}, Hillsdale NJ. Lawrence Erlbaum Associates.

\bibitem[Craven and Wahba, 1979]{Craven:79}
Craven, P. and Wahba, G. (1979).
\newblock Smoothing noisy data with spline functions: Estimating the correct
  degree of smoothing by the method of generalized cross-validation.
\newblock {\em Numer. Math.}, 31:377--403.

\bibitem[Cuccu et~al., 2011]{cuccu2011}
Cuccu, G., Luciw, M., Schmidhuber, J., and Gomez, F. (2011).
\newblock Intrinsically motivated evolutionary search for vision-based
  reinforcement learning.
\newblock In {\em Proceedings of the 2011 IEEE Conference on Development and
  Learning and Epigenetic Robotics IEEE-ICDL-EPIROB}, volume~2, pages 1--7.
  IEEE.

\bibitem[Dahl et~al., 2012]{dahl2012}
Dahl, G., Yu, D., Deng, L., and Acero, A. (2012).
\newblock Context-dependent pre-trained deep neural networks for
  large-vocabulary speech recognition.
\newblock {\em Audio, Speech, and Language Processing, IEEE Transactions on},
  20(1):30--42.

\bibitem[Dahl et~al., 2013]{Dahl2013}
Dahl, G.~E., Sainath, T.~N., and Hinton, G.~E. (2013).
\newblock Improving deep neural networks for {LVCSR} using rectified linear
  units and dropout.
\newblock In {\em IEEE International Conference on Acoustics, Speech and Signal
  Processing (ICASSP)}, pages 8609--8613. IEEE.

\bibitem[D'Ambrosio and Stanley, 2007]{stanley07}
D'Ambrosio, D.~B. and Stanley, K.~O. (2007).
\newblock A novel generative encoding for exploiting neural network sensor and
  output geometry.
\newblock In {\em Proceedings of the Conference on Genetic and Evolutionary
  Computation (GECCO)}, pages 974--981.

\bibitem[Datar et~al., 2004]{datar2004}
Datar, M., Immorlica, N., Indyk, P., and Mirrokni, V.~S. (2004).
\newblock Locality-sensitive hashing scheme based on p-stable distributions.
\newblock In {\em Proceedings of the 20th Annual Symposium on Computational
  Geometry}, pages 253--262. ACM.

\bibitem[Dayan and Hinton, 1993]{Dayan:93}
Dayan, P. and Hinton, G. (1993).
\newblock Feudal reinforcement learning.
\newblock In Lippman, D.~S., Moody, J.~E., and Touretzky, D.~S., editors, {\em
  Advances in Neural Information Processing Systems (NIPS) 5}, pages 271--278.
  Morgan Kaufmann.

\bibitem[Dayan and Hinton, 1996]{Dayan:96}
Dayan, P. and Hinton, G.~E. (1996).
\newblock Varieties of {Helmholtz} machine.
\newblock {\em Neural Networks}, 9(8):1385--1403.

\bibitem[Dayan et~al., 1995]{Dayan:95}
Dayan, P., Hinton, G.~E., Neal, R.~M., and Zemel, R.~S. (1995).
\newblock The {Helmholtz} machine.
\newblock {\em Neural Computation}, 7:889--904.

\bibitem[Dayan and Zemel, 1995]{Dayan:95a}
Dayan, P. and Zemel, R. (1995).
\newblock Competition and multiple cause models.
\newblock {\em Neural Computation}, 7:565--579.

\bibitem[De~Freitas, 2003]{freitas2003}
De~Freitas, J. F.~G. (2003).
\newblock {\em Bayesian methods for neural networks}.
\newblock PhD thesis, University of Cambridge.

\bibitem[de~Souto et~al., 1999]{souto1999}
de~Souto, M.~C., Souto, M. C. P.~D., and Oliveira, W. R.~D. (1999).
\newblock The loading problem for pyramidal neural networks.
\newblock In {\em Electronic Journal on Mathematics of Computation}.

\bibitem[De~Valois et~al., 1982]{valois1982}
De~Valois, R.~L., Albrecht, D.~G., and Thorell, L.~G. (1982).
\newblock Spatial frequency selectivity of cells in macaque visual cortex.
\newblock {\em Vision Research}, 22(5):545--559.

\bibitem[de~Vries and Principe, 1991]{Vries:91}
de~Vries, B. and Principe, J.~C. (1991).
\newblock A theory for neural networks with time delays.
\newblock In Lippmann, R.~P., Moody, J.~E., and Touretzky, D.~S., editors, {\em
  Advances in Neural Information Processing Systems (NIPS) 3}, pages 162--168.
  Morgan Kaufmann.

\bibitem[Deco and Parra, 1997]{DecoParra:97}
Deco, G. and Parra, L. (1997).
\newblock Non-linear feature extraction by redundancy reduction in an
  unsupervised stochastic neural network.
\newblock {\em Neural Networks}, 10(4):683--691.

\bibitem[Deco and Rolls, 2005]{deco2005}
Deco, G. and Rolls, E.~T. (2005).
\newblock Neurodynamics of biased competition and cooperation for attention: a
  model with spiking neurons.
\newblock {\em Journal of Neurophysiology}, 94(1):295--313.

\bibitem[DeJong and Mooney, 1986]{dejong1986}
DeJong, G. and Mooney, R. (1986).
\newblock Explanation-based learning: An alternative view.
\newblock {\em Machine Learning}, 1(2):145--176.

\bibitem[DeMers and Cottrell, 1993]{DeMers:93}
DeMers, D. and Cottrell, G. (1993).
\newblock Non-linear dimensionality reduction.
\newblock In Hanson, S.~J., Cowan, J.~D., and Giles, C.~L., editors, {\em
  Advances in Neural Information Processing Systems (NIPS) 5}, pages 580--587.
  Morgan Kaufmann.

\bibitem[Dempster et~al., 1977]{dempster77}
Dempster, A.~P., Laird, N.~M., and Rubin, D.~B. (1977).
\newblock Maximum likelihood from incomplete data via the {EM} algorithm.
\newblock {\em Journal of the Royal Statistical Society, B}, 39.

\bibitem[Deng and Yu, 2014]{lideng2014}
Deng, L. and Yu, D. (2014).
\newblock {\em Deep Learning: Methods and Applications}.
\newblock NOW Publishers.

\bibitem[Desimone et~al., 1984]{desimone1984}
Desimone, R., Albright, T.~D., Gross, C.~G., and Bruce, C. (1984).
\newblock Stimulus-selective properties of inferior temporal neurons in the
  macaque.
\newblock {\em The Journal of Neuroscience}, 4(8):2051--2062.

\bibitem[Deville and Lau, 1994]{Deville:94}
Deville, Y. and Lau, K.~K. (1994).
\newblock Logic program synthesis.
\newblock {\em Journal of Logic Programming}, 19(20):321--350.

\bibitem[Di~Lena et~al., 2012]{baldi2012contact}
Di~Lena, P., Nagata, K., and Baldi, P. (2012).
\newblock Deep architectures for protein contact map prediction.
\newblock {\em Bioinformatics}, 28:2449--2457.

\bibitem[DiCarlo et~al., 2012]{dicarlo2012}
DiCarlo, J.~J., Zoccolan, D., and Rust, N.~C. (2012).
\newblock How does the brain solve visual object recognition?
\newblock {\em Neuron}, 73(3):415--434.

\bibitem[Dickmanns et~al., 1987]{gp87}
Dickmanns, D., Schmidhuber, J., and Winklhofer, A. (1987).
\newblock {Der genetische Algorithmus: Eine Implementierung in Prolog.
  Technical Report, Inst. of Informatics, Tech. Univ. Munich}.
\newblock http://www.idsia.ch/\~{ }juergen/geneticprogramming.html.

\bibitem[Dickmanns et~al., 1994]{Dickmanns:94}
Dickmanns, E.~D., Behringer, R., Dickmanns, D., Hildebrandt, T., Maurer, M.,
  Thomanek, F., and Schiehlen, J. (1994).
\newblock The seeing passenger car {'VaMoRs-P'}.
\newblock In {\em Proc. Int. Symp. on Intelligent Vehicles '94, Paris}, pages
  68--73.

\bibitem[Dietterich, 2000a]{dietterich2000}
Dietterich, T.~G. (2000a).
\newblock Ensemble methods in machine learning.
\newblock In {\em Multiple classifier systems}, pages 1--15. Springer.

\bibitem[Dietterich, 2000b]{Dietterich:MAXQ}
Dietterich, T.~G. (2000b).
\newblock Hierarchical reinforcement learning with the {MAXQ} value function
  decomposition.
\newblock {\em J. Artif. Intell. Res. (JAIR)}, 13:227--303.

\bibitem[Director and Rohrer, 1969]{director:1969}
Director, S.~W. and Rohrer, R.~A. (1969).
\newblock Automated network design - the frequency-domain case.
\newblock {\em IEEE Trans. Circuit Theory}, CT-16:330--337.

\bibitem[Dittenbach et~al., 2000]{dittenbach2000}
Dittenbach, M., Merkl, D., and Rauber, A. (2000).
\newblock The growing hierarchical self-organizing map.
\newblock In {\em IEEE-INNS-ENNS International Joint Conference on Neural
  Networks}, volume~6, pages 6015--6015. IEEE Computer Society.

\bibitem[Donahue et~al., 2013]{donahue2013}
Donahue, J., Jia, Y., Vinyals, O., Hoffman, J., Zhang, N., Tzeng, E., and
  Darrell, T. (2013).
\newblock {DeCAF}: A deep convolutional activation feature for generic visual
  recognition.
\newblock {\em arXiv preprint arXiv:1310.1531}.

\bibitem[Dorffner, 1996]{dorffner1996}
Dorffner, G. (1996).
\newblock Neural networks for time series processing.
\newblock In {\em Neural Network World}.

\bibitem[Doya et~al., 2002]{DoyaSamejimaKatagiriKawato}
Doya, K., Samejima, K., ichi Katagiri, K., and Kawato, M. (2002).
\newblock Multiple model-based reinforcement learning.
\newblock {\em Neural Computation}, 14(6):1347--1369.

\bibitem[Dreyfus, 1962]{dreyfus:1962}
Dreyfus, S.~E. (1962).
\newblock The numerical solution of variational problems.
\newblock {\em Journal of Mathematical Analysis and Applications}, 5(1):30--45.

\bibitem[Dreyfus, 1973]{dreyfus:1973}
Dreyfus, S.~E. (1973).
\newblock The computational solution of optimal control problems with time lag.
\newblock {\em IEEE Transactions on Automatic Control}, 18(4):383--385.

\bibitem[Duchi et~al., 2011]{Duchi2011}
Duchi, J., Hazan, E., and Singer, Y. (2011).
\newblock {Adaptive subgradient methods for online learning and stochastic
  optimization}.
\newblock {\em The Journal of Machine Learning}, 12:2121--2159.

\bibitem[Egorova et~al., 2004]{egorova03}
Egorova, A., Gloye, A., G{\"o}ktekin, C., Liers, A., Luft, M., Rojas, R.,
  Simon, M., Tenchio, O., and Wiesel, F. (2004).
\newblock {FU-Fighters Small Size 2004, Team Description}.
\newblock RoboCup 2004 Symposium: Papers and Team Description Papers.
\newblock CD edition.

\bibitem[Elfwing et~al., 2010]{elfwing2010}
Elfwing, S., Otsuka, M., Uchibe, E., and Doya, K. (2010).
\newblock Free-energy based reinforcement learning for vision-based navigation
  with high-dimensional sensory inputs.
\newblock In {\em Neural Information Processing. Theory and Algorithms
  (ICONIP)}, volume~1, pages 215--222. Springer.

\bibitem[Eliasmith, 2013]{eliasmith2013}
Eliasmith, C. (2013).
\newblock {\em How to build a brain: A neural architecture for biological
  cognition}.
\newblock Oxford University Press, New York, NY.

\bibitem[Eliasmith et~al., 2012]{eliasmith2012}
Eliasmith, C., Stewart, T.~C., Choo, X., Bekolay, T., DeWolf, T., Tang, Y., and
  Rasmussen, D. (2012).
\newblock A large-scale model of the functioning brain.
\newblock {\em Science}, 338(6111):1202--1205.

\bibitem[Elman, 1990]{elman1990}
Elman, J.~L. (1990).
\newblock Finding structure in time.
\newblock {\em Cognitive Science}, 14(2):179--211.

\bibitem[Erhan et~al., 2010]{erhan:10whydoes}
Erhan, D., Bengio, Y., Courville, A., Manzagol, P.-A., Vincent, P., and Bengio,
  S. (2010).
\newblock Why does unsupervised pre-training help deep learning?
\newblock {\em J. Mach. Learn. Res.}, 11:625{\textendash}660.

\bibitem[Escalante-B. and Wiskott, 2013]{wiskott2013}
Escalante-B., A.~N. and Wiskott, L. (2013).
\newblock How to solve classification and regression problems on
  high-dimensional data with a supervised extension of slow feature analysis.
\newblock {\em Journal of Machine Learning Research}, 14:3683--3719.

\bibitem[Eubank, 1988]{Eubank:88}
Eubank, R.~L. (1988).
\newblock Spline smoothing and nonparametric regression.
\newblock In Farlow, S., editor, {\em Self-Organizing Methods in Modeling}.
  Marcel Dekker, New York.

\bibitem[Euler, 1744]{Euler:1744}
Euler, L. (1744).
\newblock {\em Methodus inveniendi}.

\bibitem[Eyben et~al., 2013]{eyben2013}
Eyben, F., Weninger, F., Squartini, S., and Schuller, B. (2013).
\newblock Real-life voice activity detection with {LSTM} recurrent neural
  networks and an application to {Hollywood} movies.
\newblock In {\em Proc. 38th IEEE International Conference on Acoustics,
  Speech, and Signal Processing, ICASSP 2013, Vancouver, Canada}, pages
  483--487.

\bibitem[Faggin, 1992]{faggin92}
Faggin, F. (1992).
\newblock Neural network hardware.
\newblock In {\em International Joint Conference on Neural Networks (IJCNN)},
  volume~1, page 153.

\bibitem[Fahlman, 1988]{Fahlman:88}
Fahlman, S.~E. (1988).
\newblock An empirical study of learning speed in back-propagation networks.
\newblock Technical Report CMU-CS-88-162, Carnegie-Mellon Univ.

\bibitem[Fahlman, 1991]{Fahlman:91}
Fahlman, S.~E. (1991).
\newblock The recurrent cascade-correlation learning algorithm.
\newblock In Lippmann, R.~P., Moody, J.~E., and Touretzky, D.~S., editors, {\em
  Advances in Neural Information Processing Systems (NIPS) 3}, pages 190--196.
  Morgan Kaufmann.

\bibitem[Falconbridge et~al., 2006]{falconbridge2006}
Falconbridge, M.~S., Stamps, R.~L., and Badcock, D.~R. (2006).
\newblock A simple {Hebbian/anti-Hebbian} network learns the sparse,
  independent components of natural images.
\newblock {\em Neural Computation}, 18(2):415--429.

\bibitem[Fan et~al., 2014]{fan2014}
Fan, Y., Qian, Y., Xie, F., and Soong, F.~K. (2014).
\newblock {TTS} synthesis with bidirectional {LSTM} based recurrent neural
  networks.
\newblock In {\em Proc. Interspeech}.

\bibitem[Farabet et~al., 2013]{Farabet2013}
Farabet, C., Couprie, C., Najman, L., and LeCun, Y. (2013).
\newblock Learning hierarchical features for scene labeling.
\newblock {\em IEEE Transactions on Pattern Analysis and Machine Intelligence},
  35(8):1915--1929.

\bibitem[Farlow, 1984]{farlow1984}
Farlow, S.~J. (1984).
\newblock {\em Self-organizing methods in modeling: {GMDH} type algorithms},
  volume~54.
\newblock CRC Press.

\bibitem[Feldkamp et~al., 1998]{feldkamp1998kalman}
Feldkamp, L.~A., Prokhorov, D.~V., Eagen, C.~F., and Yuan, F. (1998).
\newblock Enhanced multi-stream {Kalman} filter training for recurrent
  networks.
\newblock In {\em Nonlinear Modeling}, pages 29--53. Springer.

\bibitem[Feldkamp et~al., 2003]{feldkamp2003}
Feldkamp, L.~A., Prokhorov, D.~V., and Feldkamp, T.~M. (2003).
\newblock Simple and conditioned adaptive behavior from {Kalman} filter trained
  recurrent networks.
\newblock {\em Neural Networks}, 16(5):683--689.

\bibitem[Feldkamp and Puskorius, 1998]{feldkamp1998}
Feldkamp, L.~A. and Puskorius, G.~V. (1998).
\newblock A signal processing framework based on dynamic neural networks with
  application to problems in adaptation, filtering, and classification.
\newblock {\em Proceedings of the IEEE}, 86(11):2259--2277.

\bibitem[Felleman and Van~Essen, 1991]{felleman1991}
Felleman, D.~J. and Van~Essen, D.~C. (1991).
\newblock Distributed hierarchical processing in the primate cerebral cortex.
\newblock {\em Cerebral Cortex}, 1(1):1--47.

\bibitem[Fernandez et~al., 2014]{fernandez2014}
Fernandez, R., Rendel, A., Ramabhadran, B., and Hoory, R. (2014).
\newblock Prosody contour prediction with {Long Short-Term Memory},
  bi-directional, deep recurrent neural networks.
\newblock In {\em Proc. Interspeech}.

\bibitem[Fern{\'a}ndez et~al., 2007]{DBLP:conf/icann/FernandezGS07}
Fern{\'a}ndez, S., Graves, A., and Schmidhuber, J. (2007).
\newblock An application of recurrent neural networks to discriminative keyword
  spotting.
\newblock In {\em Proc. ICANN (2)}, pages 220--229.

\bibitem[Fernandez et~al., 2007]{Santi:07ijcai}
Fernandez, S., Graves, A., and Schmidhuber, J. (2007).
\newblock Sequence labelling in structured domains with hierarchical recurrent
  neural networks.
\newblock In {\em Proceedings of the 20th International Joint Conference on
  Artificial Intelligence (IJCAI)}.

\bibitem[Field, 1987]{Field:87}
Field, D.~J. (1987).
\newblock Relations between the statistics of natural images and the response
  properties of cortical cells.
\newblock {\em Journal of the Optical Society of America}, 4:2379--2394.

\bibitem[Field, 1994]{Field:94}
Field, D.~J. (1994).
\newblock What is the goal of sensory coding?
\newblock {\em Neural Computation}, 6:559--601.

\bibitem[Fieres et~al., 2008]{fieres2008}
Fieres, J., Schemmel, J., and Meier, K. (2008).
\newblock Realizing biological spiking network models in a configurable
  wafer-scale hardware system.
\newblock In {\em IEEE International Joint Conference on Neural Networks},
  pages 969--976.

\bibitem[Fine et~al., 1998]{tishby1998}
Fine, S., Singer, Y., and Tishby, N. (1998).
\newblock The hierarchical hidden {Markov} model: Analysis and applications.
\newblock {\em Machine Learning}, 32(1):41--62.

\bibitem[Fischer and Igel, 2014]{fischer:13}
Fischer, A. and Igel, C. (2014).
\newblock Training restricted {B}oltzmann machines: {An} introduction.
\newblock {\em Pattern Recognition}, 47:25--39.

\bibitem[FitzHugh, 1961]{fitzhugh1961}
FitzHugh, R. (1961).
\newblock Impulses and physiological states in theoretical models of nerve
  membrane.
\newblock {\em Biophysical Journal}, 1(6):445--466.

\bibitem[Fletcher and Powell, 1963]{fletcher1963}
Fletcher, R. and Powell, M.~J. (1963).
\newblock A rapidly convergent descent method for minimization.
\newblock {\em The Computer Journal}, 6(2):163--168.

\bibitem[Floreano and Mattiussi, 2001]{floreano2001}
Floreano, D. and Mattiussi, C. (2001).
\newblock Evolution of spiking neural controllers for autonomous vision-based
  robots.
\newblock In {\em Evolutionary Robotics. From Intelligent Robotics to
  Artificial Life}, pages 38--61. Springer.

\bibitem[Fogel et~al., 1990]{fogel1990}
Fogel, D.~B., Fogel, L.~J., and Porto, V. (1990).
\newblock Evolving neural networks.
\newblock {\em Biological Cybernetics}, 63(6):487--493.

\bibitem[Fogel et~al., 1966]{Fogel:66}
Fogel, L., Owens, A., and Walsh, M. (1966).
\newblock {\em Artificial Intelligence through Simulated Evolution}.
\newblock Wiley, New York.

\bibitem[F\"{o}ldi\'{a}k, 1990]{Foldiak:90}
F\"{o}ldi\'{a}k, P. (1990).
\newblock Forming sparse representations by local anti-{Hebbian} learning.
\newblock {\em Biological Cybernetics}, 64:165--170.

\bibitem[F\"{o}ldi\'{a}k and Young, 1995]{Foldiak:95}
F\"{o}ldi\'{a}k, P. and Young, M.~P. (1995).
\newblock Sparse coding in the primate cortex.
\newblock In Arbib, M.~A., editor, {\em The Handbook of Brain Theory and Neural
  Networks}, pages 895--898. The MIT Press.

\bibitem[F{\"o}rster et~al., 2007]{foerster-esann07}
F{\"o}rster, A., Graves, A., and Schmidhuber, J. (2007).
\newblock {RNN-based Learning of Compact Maps for Efficient Robot
  Localization}.
\newblock In {\em {15th European Symposium on Artificial Neural Networks,
  ESANN}}, pages 537--542, Bruges, Belgium.

\bibitem[Franzius et~al., 2007]{Franzius2007a}
Franzius, M., Sprekeler, H., and Wiskott, L. (2007).
\newblock {Slowness and sparseness lead to place, head-direction, and
  spatial-view cells}.
\newblock {\em PLoS Computational Biology}, 3(8):166.

\bibitem[Friedman et~al., 2001]{friedman2001}
Friedman, J., Hastie, T., and Tibshirani, R. (2001).
\newblock {\em The elements of statistical learning}, volume~1.
\newblock Springer Series in Statistics, New York.

\bibitem[Frinken et~al., 2012]{frinken2012long}
Frinken, V., Zamora-Martinez, F., Espana-Boquera, S., Castro-Bleda, M.~J.,
  Fischer, A., and Bunke, H. (2012).
\newblock Long-short term memory neural networks language modeling for
  handwriting recognition.
\newblock In {\em Pattern Recognition (ICPR), 2012 21st International
  Conference on}, pages 701--704. IEEE.

\bibitem[Fritzke, 1994]{fritzke94}
Fritzke, B. (1994).
\newblock A growing neural gas network learns topologies.
\newblock In Tesauro, G., Touretzky, D.~S., and Leen, T.~K., editors, {\em
  NIPS}, pages 625--632. MIT Press.

\bibitem[Fu, 1977]{Fu:77}
Fu, K.~S. (1977).
\newblock {\em Syntactic Pattern Recognition and Applications}.
\newblock Berlin, Springer.

\bibitem[Fukada et~al., 1999]{fukada99boundary}
Fukada, T., Schuster, M., and Sagisaka, Y. (1999).
\newblock Phoneme boundary estimation using bidirectional recurrent neural
  networks and its applications.
\newblock {\em Systems and Computers in Japan}, 30(4):20--30.

\bibitem[Fukushima, 1979]{Fukushima:1979neocognitron}
Fukushima, K. (1979).
\newblock Neural network model for a mechanism of pattern recognition
  unaffected by shift in position - {Neocognitron}.
\newblock {\em Trans. IECE}, J62-A(10):658--665.

\bibitem[Fukushima, 1980]{fukushima:1980}
Fukushima, K. (1980).
\newblock Neocognitron: A self-organizing neural network for a mechanism of
  pattern recognition unaffected by shift in position.
\newblock {\em Biological Cybernetics}, 36(4):193--202.

\bibitem[Fukushima, 2011]{Fukushima:2011}
Fukushima, K. (2011).
\newblock {Increasing robustness against background noise: visual pattern
  recognition by a Neocognitron}.
\newblock {\em Neural Networks}, 24(7):767--778.

\bibitem[Fukushima, 2013a]{Fukushima:2013}
Fukushima, K. (2013a).
\newblock {Artificial vision by multi-layered neural networks: Neocognitron and
  its advances}.
\newblock {\em Neural Networks}, 37:103--119.

\bibitem[Fukushima, 2013b]{Fukushima:2013b}
Fukushima, K. (2013b).
\newblock {Training multi-layered neural network Neocognitron}.
\newblock {\em Neural Networks}, 40:18--31.

\bibitem[Gabor, 1946]{gabor1946}
Gabor, D. (1946).
\newblock Theory of communication. {Part} 1: The analysis of information.
\newblock {\em Electrical Engineers-Part III: Journal of the Institution of
  Radio and Communication Engineering}, 93(26):429--441.

\bibitem[Gallant, 1988]{gallant1988}
Gallant, S.~I. (1988).
\newblock Connectionist expert systems.
\newblock {\em Communications of the ACM}, 31(2):152--169.

\bibitem[Gauss, 1809]{gauss1809}
Gauss, C.~F. (1809).
\newblock {\em Theoria motus corporum coelestium in sectionibus conicis solem
  ambientium}.

\bibitem[Gauss, 1821]{gauss1821}
Gauss, C.~F. (1821).
\newblock {\em Theoria combinationis observationum erroribus minimis obnoxiae
  (Theory of the combination of observations least subject to error)}.

\bibitem[Ge et~al., 2010]{ge2010}
Ge, S., Hang, C.~C., Lee, T.~H., and Zhang, T. (2010).
\newblock {\em Stable adaptive neural network control}.
\newblock Springer.

\bibitem[Geiger et~al., 2014]{geiger2014}
Geiger, J.~T., Zhang, Z., Weninger, F., Schuller, B., and Rigoll, G. (2014).
\newblock Robust speech recognition using long short-term memory recurrent
  neural networks for hybrid acoustic modelling.
\newblock In {\em Proc. Interspeech}.

\bibitem[Geman et~al., 1992]{Geman:92}
Geman, S., Bienenstock, E., and Doursat, R. (1992).
\newblock Neural networks and the bias/variance dilemma.
\newblock {\em Neural Computation}, 4:1--58.

\bibitem[Gers and Schmidhuber, 2000]{Gers:2000b}
Gers, F.~A. and Schmidhuber, J. (2000).
\newblock Recurrent nets that time and count.
\newblock In {\em Neural Networks, 2000. IJCNN 2000, Proceedings of the
  IEEE-INNS-ENNS International Joint Conference on}, volume~3, pages 189--194.
  IEEE.

\bibitem[Gers and Schmidhuber, 2001]{Gers:01ieeetnn}
Gers, F.~A. and Schmidhuber, J. (2001).
\newblock {LSTM} recurrent networks learn simple context free and context
  sensitive languages.
\newblock {\em IEEE Transactions on Neural Networks}, 12(6):1333--1340.

\bibitem[Gers et~al., 2000]{Gers:2000nc}
Gers, F.~A., Schmidhuber, J., and Cummins, F. (2000).
\newblock Learning to forget: Continual prediction with {LSTM}.
\newblock {\em Neural Computation}, 12(10):2451--2471.

\bibitem[Gers et~al., 2002]{Gers:02jmlr}
Gers, F.~A., Schraudolph, N., and Schmidhuber, J. (2002).
\newblock Learning precise timing with {LSTM} recurrent networks.
\newblock {\em Journal of Machine Learning Research}, 3:115--143.

\bibitem[Gerstner and Kistler, 2002]{gerstnerbook}
Gerstner, W. and Kistler, W.~K. (2002).
\newblock {\em Spiking {N}euron {M}odels}.
\newblock Cambridge University Press.

\bibitem[Gerstner and van Hemmen, 1992]{gerstner1992}
Gerstner, W. and van Hemmen, J.~L. (1992).
\newblock Associative memory in a network of spiking neurons.
\newblock {\em Network: Computation in Neural Systems}, 3(2):139--164.

\bibitem[Ghavamzadeh and Mahadevan, 2003]{ghavamzadehICML03}
Ghavamzadeh, M. and Mahadevan, S. (2003).
\newblock Hierarchical policy gradient algorithms.
\newblock In {\em Proceedings of the Twentieth Conference on Machine Learning
  (ICML-2003)}, pages 226--233.

\bibitem[Gherrity, 1989]{Gherrity:89}
Gherrity, M. (1989).
\newblock A learning algorithm for analog fully recurrent neural networks.
\newblock In {\em IEEE/INNS International Joint Conference on Neural Networks,
  San Diego}, volume~1, pages 643--644.

\bibitem[Girshick et~al., 2013]{malik2013}
Girshick, R., Donahue, J., Darrell, T., and Malik, J. (2013).
\newblock Rich feature hierarchies for accurate object detection and semantic
  segmentation.
\newblock Technical Report arxiv.org/abs/1311.2524, UC Berkeley and ICSI.

\bibitem[Gisslen et~al., 2011]{Gisslen2011agi}
Gisslen, L., Luciw, M., Graziano, V., and Schmidhuber, J. (2011).
\newblock Sequential constant size compressor for reinforcement learning.
\newblock In {\em {Proc. Fourth Conference on Artificial General Intelligence
  (AGI), Google, Mountain View, CA}}, pages 31--40. Springer.

\bibitem[Giusti et~al., 2013]{Giusti:2013a}
Giusti, A., Ciresan, D.~C., Masci, J., Gambardella, L.~M., and Schmidhuber, J.
  (2013).
\newblock Fast image scanning with deep max-pooling convolutional neural
  networks.
\newblock In {\em Proc. ICIP}.

\bibitem[Glackin et~al., 2005]{glackin2005}
Glackin, B., McGinnity, T.~M., Maguire, L.~P., Wu, Q., and Belatreche, A.
  (2005).
\newblock A novel approach for the implementation of large scale spiking neural
  networks on {FPGA} hardware.
\newblock In {\em Computational Intelligence and Bioinspired Systems}, pages
  552--563. Springer.

\bibitem[Glasmachers et~al., 2010]{glasmachers:2010b}
Glasmachers, T., Schaul, T., Sun, Y., Wierstra, D., and Schmidhuber, J. (2010).
\newblock Exponential natural evolution strategies.
\newblock In {\em Proceedings of the Genetic and Evolutionary Computation
  Conference (GECCO)}, pages 393--400. ACM.

\bibitem[Glorot et~al., 2011]{Glorot2011a}
Glorot, X., Bordes, A., and Bengio, Y. (2011).
\newblock Deep sparse rectifier networks.
\newblock In {\em AISTATS}, volume~15, pages 315--323.

\bibitem[Gloye et~al., 2005]{gloye05}
Gloye, A., Wiesel, F., Tenchio, O., and Simon, M. (2005).
\newblock Reinforcing the driving quality of soccer playing robots by
  anticipation.
\newblock {\em IT - Information Technology}, 47(5).

\bibitem[G\"{o}del, 1931]{Goedel:31}
G\"{o}del, K. (1931).
\newblock \"{U}ber formal unentscheidbare {S\"{a}tze der Principia Mathematica
  und verwandter Systeme I}.
\newblock {\em Monatshefte f\"{u}r Mathematik und Physik}, 38:173--198.

\bibitem[Goldberg, 1989]{goldberg:gabook89}
Goldberg, D.~E. (1989).
\newblock {\em Genetic Algorithms in Search, Optimization and Machine
  Learning}.
\newblock Addison-Wesley, Reading, MA.

\bibitem[Goldfarb, 1970]{goldfarb1970}
Goldfarb, D. (1970).
\newblock A family of variable-metric methods derived by variational means.
\newblock {\em Mathematics of computation}, 24(109):23--26.

\bibitem[Golub et~al., 1979]{Golub:79}
Golub, G., Heath, H., and Wahba, G. (1979).
\newblock Generalized cross-validation as a method for choosing a good ridge
  parameter.
\newblock {\em Technometrics}, 21:215--224.

\bibitem[Gomez, 2003]{gomez:phd}
Gomez, F.~J. (2003).
\newblock {\em Robust Nonlinear Control through Neuroevolution}.
\newblock PhD thesis, Department of Computer Sciences, University of Texas at
  Austin.

\bibitem[Gomez and Miikkulainen, 2003]{Gomez:03}
Gomez, F.~J. and Miikkulainen, R. (2003).
\newblock Active guidance for a finless rocket using neuroevolution.
\newblock In {\em Proc. GECCO 2003, Chicago}.

\bibitem[Gomez and Schmidhuber, 2005]{Gomez:05gecco}
Gomez, F.~J. and Schmidhuber, J. (2005).
\newblock Co-evolving recurrent neurons learn deep memory {POMDPs}.
\newblock In {\em Proc. of the 2005 conference on genetic and evolutionary
  computation (GECCO), Washington, D. C.} ACM Press, New York, NY, USA.

\bibitem[Gomez et~al., 2008]{Gomez:08jmlr}
Gomez, F.~J., Schmidhuber, J., and Miikkulainen, R. (2008).
\newblock Accelerated neural evolution through cooperatively coevolved
  synapses.
\newblock {\em Journal of Machine Learning Research}, 9(May):937--965.

\bibitem[Gomi and Kawato, 1993]{kawato1993}
Gomi, H. and Kawato, M. (1993).
\newblock Neural network control for a closed-loop system using
  feedback-error-learning.
\newblock {\em Neural Networks}, 6(7):933--946.

\bibitem[Gonzalez-Dominguez et~al., 2014]{gonzalez2014}
Gonzalez-Dominguez, J., Lopez-Moreno, I., Sak, H., Gonzalez-Rodriguez, J., and
  Moreno, P.~J. (2014).
\newblock Automatic language identification using {Long Short-Term Memory}
  recurrent neural networks.
\newblock In {\em Proc. Interspeech}.

\bibitem[Goodfellow et~al., 2014a]{Goodfellow2014}
Goodfellow, I., Mirza, M., Da, X., Courville, A., and Bengio, Y. (2014a).
\newblock {An Empirical Investigation of Catastrophic Forgetting in
  Gradient-Based Neural Networks}.
\newblock {\em TR arXiv:1312.6211v2}.

\bibitem[Goodfellow et~al., 2014b]{goodfellow2014multi}
Goodfellow, I.~J., Bulatov, Y., Ibarz, J., Arnoud, S., and Shet, V. (2014b).
\newblock Multi-digit number recognition from street view imagery using deep
  convolutional neural networks.
\newblock {\em arXiv preprint arXiv:1312.6082 v4}.

\bibitem[Goodfellow et~al., 2011]{goodfellow2011}
Goodfellow, I.~J., Courville, A., and Bengio, Y. (2011).
\newblock Spike-and-slab sparse coding for unsupervised feature discovery.
\newblock In {\em NIPS Workshop on Challenges in Learning Hierarchical Models}.

\bibitem[Goodfellow et~al., 2012]{goodfellow:2012icml}
Goodfellow, I.~J., Courville, A.~C., and Bengio, Y. (2012).
\newblock Large-scale feature learning with spike-and-slab sparse coding.
\newblock In {\em Proceedings of the 29th International Conference on Machine
  Learning (ICML)}.

\bibitem[Goodfellow et~al., 2013]{goodfellow2013maxout}
Goodfellow, I.~J., Warde-Farley, D., Mirza, M., Courville, A., and Bengio, Y.
  (2013).
\newblock Maxout networks.
\newblock In {\em International Conference on Machine Learning (ICML)}.

\bibitem[Graves, 2011]{Graves2011}
Graves, A. (2011).
\newblock Practical variational inference for neural networks.
\newblock In {\em Advances in Neural Information Processing Systems (NIPS)},
  pages 2348--2356.

\bibitem[Graves et~al., 2003]{graves+eck+beringer+schmidhuber:2003}
Graves, A., Eck, D., Beringer, N., and Schmidhuber, J. (2003).
\newblock Isolated digit recognition with {LSTM} recurrent networks.
\newblock In {\em First International Workshop on Biologically Inspired
  Approaches to Advanced Information Technology}, Lausanne.

\bibitem[Graves et~al., 2006]{Graves:06icml}
Graves, A., Fernandez, S., Gomez, F.~J., and Schmidhuber, J. (2006).
\newblock Connectionist temporal classification: Labelling unsegmented sequence
  data with recurrent neural nets.
\newblock In {\em ICML'06: Proceedings of the 23rd International Conference on
  Machine Learning}, pages 369--376.

\bibitem[Graves et~al., 2008]{graves:08nips}
Graves, A., Fernandez, S., Liwicki, M., Bunke, H., and Schmidhuber, J. (2008).
\newblock Unconstrained on-line handwriting recognition with recurrent neural
  networks.
\newblock In Platt, J., Koller, D., Singer, Y., and Roweis, S., editors, {\em
  Advances in Neural Information Processing Systems (NIPS) 20}, pages 577--584.
  MIT Press, Cambridge, MA.

\bibitem[Graves and Jaitly, 2014]{graves2014}
Graves, A. and Jaitly, N. (2014).
\newblock Towards end-to-end speech recognition with recurrent neural networks.
\newblock In {\em Proc. 31st International Conference on Machine Learning
  (ICML)}, pages 1764--1772.

\bibitem[Graves et~al., 2009]{Graves:09tpami}
Graves, A., Liwicki, M., Fernandez, S., Bertolami, R., Bunke, H., and
  Schmidhuber, J. (2009).
\newblock A novel connectionist system for improved unconstrained handwriting
  recognition.
\newblock {\em IEEE Transactions on Pattern Analysis and Machine Intelligence},
  31(5).

\bibitem[Graves et~al., 2013]{graves:2013icassp}
Graves, A., Mohamed, A.-R., and Hinton, G.~E. (2013).
\newblock Speech recognition with deep recurrent neural networks.
\newblock In {\em IEEE International Conference on Acoustics, Speech and Signal
  Processing (ICASSP)}, pages 6645--6649. IEEE.

\bibitem[Graves and Schmidhuber, 2005]{graves05nn}
Graves, A. and Schmidhuber, J. (2005).
\newblock Framewise phoneme classification with bidirectional {LSTM} and other
  neural network architectures.
\newblock {\em Neural Networks}, 18(5-6):602--610.

\bibitem[Graves and Schmidhuber, 2009]{graves:2009nips}
Graves, A. and Schmidhuber, J. (2009).
\newblock Offline handwriting recognition with multidimensional recurrent
  neural networks.
\newblock In {\em Advances in Neural Information Processing Systems (NIPS) 21},
  pages 545--552. MIT Press, Cambridge, MA.

\bibitem[Graziano, 2009]{Graziano:book}
Graziano, M. (2009).
\newblock {\em The Intelligent Movement Machine: An Ethological Perspective on
  the Primate Motor System}.
\newblock Oxford University Press, USA.

\bibitem[Griewank, 2012]{Griewank:2012}
Griewank, A. (2012).
\newblock {\em Documenta Mathematica - Extra Volume ISMP}, pages 389--400.

\bibitem[Grondman et~al., 2012]{grondman2012}
Grondman, I., Busoniu, L., Lopes, G. A.~D., and Babuska, R. (2012).
\newblock A survey of actor-critic reinforcement learning: Standard and natural
  policy gradients.
\newblock {\em Systems, Man, and Cybernetics, Part C: Applications and Reviews,
  IEEE Transactions on}, 42(6):1291--1307.

\bibitem[Grossberg, 1969]{Grossberg69a}
Grossberg, S. (1969).
\newblock Some networks that can learn, remember, and reproduce any number of
  complicated space-time patterns, {I.}
\newblock {\em Journal of Mathematics and Mechanics}, 19:53--91.

\bibitem[Grossberg, 1976a]{Grossberg:76a}
Grossberg, S. (1976a).
\newblock Adaptive pattern classification and universal recoding, 1: Parallel
  development and coding of neural feature detectors.
\newblock {\em Biological Cybernetics}, 23:187--202.

\bibitem[Grossberg, 1976b]{Grossberg:76b}
Grossberg, S. (1976b).
\newblock Adaptive pattern classification and universal recoding, 2: Feedback,
  expectation, olfaction, and illusions.
\newblock {\em Biological Cybernetics}, 23.

\bibitem[Gruau et~al., 1996]{gruau:tr96-048}
Gruau, F., Whitley, D., and Pyeatt, L. (1996).
\newblock A comparison between cellular encoding and direct encoding for
  genetic neural networks.
\newblock {NeuroCOLT} Technical Report NC-TR-96-048, {ESPRIT} Working Group in
  Neural and Computational Learning, Neuro{COLT} 8556.

\bibitem[Gr{\"u}nwald et~al., 2005]{gruenwald2005}
Gr{\"u}nwald, P.~D., Myung, I.~J., and Pitt, M.~A. (2005).
\newblock {\em Advances in minimum description length: Theory and
  applications}.
\newblock MIT Press.

\bibitem[Gr{\"u}ttner et~al., 2010]{gruettner2010multi}
Gr{\"u}ttner, M., Sehnke, F., Schaul, T., and Schmidhuber, J. (2010).
\newblock {Multi-Dimensional Deep Memory Atari-Go Players for Parameter
  Exploring Policy Gradients}.
\newblock In {\em Proceedings of the International Conference on Artificial
  Neural Networks ICANN}, pages 114--123. Springer.

\bibitem[Guo et~al., 2014]{atarimcts2014}
Guo, X., Singh, S., Lee, H., Lewis, R., and Wang, X. (2014).
\newblock Deep learning for real-time {Atari} game play using offline
  {Monte-Carlo} tree search planning.
\newblock In {\em Advances in Neural Information Processing Systems 27 (NIPS)}.

\bibitem[Guyon et~al., 1992]{Vapnik:92a}
Guyon, I., Vapnik, V., Boser, B., Bottou, L., and Solla, S.~A. (1992).
\newblock Structural risk minimization for character recognition.
\newblock In Lippman, D.~S., Moody, J.~E., and Touretzky, D.~S., editors, {\em
  Advances in Neural Information Processing Systems (NIPS) 4}, pages 471--479.
  Morgan Kaufmann.

\bibitem[Hadamard, 1908]{hadamard1908memoire}
Hadamard, J. (1908).
\newblock {\em M{\'e}moire sur le probl{\`e}me d'analyse relatif {\`a}
  l'{\'e}quilibre des plaques {\'e}lastiques encastr{\'e}es}.
\newblock M{\'e}moires pr{\'e}sent{\'e}s par divers savants {\`a}
  l'Acad{\'e}mie des sciences de l'Institut de France: {\'E}xtrait. Imprimerie
  nationale.

\bibitem[Hadsell et~al., 2006]{hadsell-chopra-lecun-06}
Hadsell, R., Chopra, S., and LeCun, Y. (2006).
\newblock Dimensionality reduction by learning an invariant mapping.
\newblock In {\em Proc. Computer Vision and Pattern Recognition Conference
  (CVPR'06)}. IEEE Press.

\bibitem[Hagras et~al., 2004]{hagras2004}
Hagras, H., Pounds-Cornish, A., Colley, M., Callaghan, V., and Clarke, G.
  (2004).
\newblock Evolving spiking neural network controllers for autonomous robots.
\newblock In {\em IEEE International Conference on Robotics and Automation
  (ICRA)}, volume~5, pages 4620--4626.

\bibitem[Hansen et~al., 2003]{hansen2003}
Hansen, N., M{\"u}ller, S.~D., and Koumoutsakos, P. (2003).
\newblock Reducing the time complexity of the derandomized evolution strategy
  with covariance matrix adaptation {(CMA-ES)}.
\newblock {\em Evolutionary Computation}, 11(1):1--18.

\bibitem[Hansen and Ostermeier, 2001]{hansenCMA}
Hansen, N. and Ostermeier, A. (2001).
\newblock Completely derandomized self-adaptation in evolution strategies.
\newblock {\em Evolutionary Computation}, 9(2):159--195.

\bibitem[Hanson, 1990]{hanson1990}
Hanson, S.~J. (1990).
\newblock A stochastic version of the delta rule.
\newblock {\em Physica D: Nonlinear Phenomena}, 42(1):265--272.

\bibitem[Hanson and Pratt, 1989]{Hanson:89}
Hanson, S.~J. and Pratt, L.~Y. (1989).
\newblock Comparing biases for minimal network construction with
  back-propagation.
\newblock In Touretzky, D.~S., editor, {\em Advances in Neural Information
  Processing Systems (NIPS) 1}, pages 177--185. San Mateo, CA: Morgan Kaufmann.

\bibitem[Happel and Murre, 1994]{happel1994}
Happel, B.~L. and Murre, J.~M. (1994).
\newblock Design and evolution of modular neural network architectures.
\newblock {\em Neural Networks}, 7(6):985--1004.

\bibitem[Hashem and Schmeiser, 1992]{hashem:1992}
Hashem, S. and Schmeiser, B. (1992).
\newblock Improving model accuracy using optimal linear combinations of trained
  neural networks.
\newblock {\em IEEE Transactions on Neural Networks}, 6:792--794.

\bibitem[Hassibi and Stork, 1993]{Hassibi:93}
Hassibi, B. and Stork, D.~G. (1993).
\newblock Second order derivatives for network pruning: Optimal brain surgeon.
\newblock In Lippman, D.~S., Moody, J.~E., and Touretzky, D.~S., editors, {\em
  Advances in Neural Information Processing Systems 5}, pages 164--171. Morgan
  Kaufmann.

\bibitem[Hastie et~al., 2009]{hastie2009}
Hastie, T., Tibshirani, R., and Friedman, J. (2009).
\newblock {\em The elements of statistical learning}.
\newblock Springer Series in Statistics.

\bibitem[Hastie and Tibshirani, 1990]{Hastie:90}
Hastie, T.~J. and Tibshirani, R.~J. (1990).
\newblock Generalized additive models.
\newblock {\em Monographs on Statisics and Applied Probability}, 43.

\bibitem[Hawkins and George, 2006]{hawkins2006}
Hawkins, J. and George, D. (2006).
\newblock {\em Hierarchical Temporal Memory - Concepts, Theory, and
  Terminology}.
\newblock Numenta Inc.

\bibitem[Haykin, 2001]{haykin2001}
Haykin, S.~S. (2001).
\newblock {\em Kalman filtering and neural networks}.
\newblock Wiley Online Library.

\bibitem[Hebb, 1949]{Hebb:49}
Hebb, D.~O. (1949).
\newblock {\em The Organization of Behavior}.
\newblock Wiley, New York.

\bibitem[Hecht-Nielsen, 1989]{hecht1989}
Hecht-Nielsen, R. (1989).
\newblock Theory of the backpropagation neural network.
\newblock In {\em International Joint Conference on Neural Networks (IJCNN)},
  pages 593--605. IEEE.

\bibitem[Heemskerk, 1995]{heemskerk1995}
Heemskerk, J.~N. (1995).
\newblock Overview of neural hardware.
\newblock {\em Neurocomputers for Brain-Style Processing. Design,
  Implementation and Application}.

\bibitem[Heess et~al., 2012]{heess2012}
Heess, N., Silver, D., and Teh, Y.~W. (2012).
\newblock Actor-critic reinforcement learning with energy-based policies.
\newblock In {\em Proc. European Workshop on Reinforcement Learning}, pages
  43--57.

\bibitem[Heidrich-Meisner and Igel, 2009]{heidrich-meisner:09}
Heidrich-Meisner, V. and Igel, C. (2009).
\newblock Neuroevolution strategies for episodic reinforcement learning.
\newblock {\em Journal of Algorithms}, 64(4):152--168.

\bibitem[Herrero et~al., 2001]{herrero2001}
Herrero, J., Valencia, A., and Dopazo, J. (2001).
\newblock A hierarchical unsupervised growing neural network for clustering
  gene expression patterns.
\newblock {\em Bioinformatics}, 17(2):126--136.

\bibitem[Hertz et~al., 1991]{Hertz:91}
Hertz, J., Krogh, A., and Palmer, R. (1991).
\newblock {\em Introduction to the Theory of Neural Computation}.
\newblock Addison-Wesley, Redwood City.

\bibitem[Hestenes and Stiefel, 1952]{HestenesStiefel:1952}
Hestenes, M.~R. and Stiefel, E. (1952).
\newblock Methods of conjugate gradients for solving linear systems.
\newblock {\em Journal of research of the National Bureau of Standards},
  49:409--436.

\bibitem[Hihi and Bengio, 1996]{hihi:95}
Hihi, S.~E. and Bengio, Y. (1996).
\newblock Hierarchical recurrent neural networks for long-term dependencies.
\newblock In Touretzky, D.~S., Mozer, M.~C., and Hasselmo, M.~E., editors, {\em
  Advances in Neural Information Processing Systems 8}, pages 493--499. MIT
  Press.

\bibitem[Hinton and Salakhutdinov, 2006]{HinSal06}
Hinton, G. and Salakhutdinov, R. (2006).
\newblock Reducing the dimensionality of data with neural networks.
\newblock {\em Science}, 313(5786):504--507.

\bibitem[Hinton, 1989]{hinton1989connectionist}
Hinton, G.~E. (1989).
\newblock Connectionist learning procedures.
\newblock {\em Artificial intelligence}, 40(1):185--234.

\bibitem[Hinton, 2002]{hinton:2002}
Hinton, G.~E. (2002).
\newblock Training products of experts by minimizing contrastive divergence.
\newblock {\em Neural Comp.}, 14(8):1771--1800.

\bibitem[Hinton et~al., 1995]{hinton:95}
Hinton, G.~E., Dayan, P., Frey, B.~J., and Neal, R.~M. (1995).
\newblock The wake-sleep algorithm for unsupervised neural networks.
\newblock {\em Science}, 268:1158--1160.

\bibitem[Hinton et~al., 2012a]{speech2012}
Hinton, G.~E., Deng, L., Yu, D., Dahl, G.~E., Mohamed, A., Jaitly, N., Senior,
  A., Vanhoucke, V., Nguyen, P., Sainath, T.~N., and Kingsbury, B. (2012a).
\newblock Deep neural networks for acoustic modeling in speech recognition: The
  shared views of four research groups.
\newblock {\em IEEE Signal Process. Mag.}, 29(6):82--97.

\bibitem[Hinton and Ghahramani, 1997]{Hinton:97}
Hinton, G.~E. and Ghahramani, Z. (1997).
\newblock Generative models for discovering sparse distributed representations.
\newblock {\em Philosophical Transactions of the Royal Society {\bf B}},
  352:1177--1190.

\bibitem[Hinton et~al., 2006]{hinton:06afast}
Hinton, G.~E., Osindero, S., and Teh, Y.-W. (2006).
\newblock A fast learning algorithm for deep belief nets.
\newblock {\em Neural Computation}, 18(7):1527--1554.

\bibitem[Hinton and Sejnowski, 1986]{HintonSejnowski:86}
Hinton, G.~E. and Sejnowski, T.~E. (1986).
\newblock Learning and relearning in {Boltzmann} machines.
\newblock In {\em Parallel Distributed Processing}, volume~1, pages 282--317.
  MIT Press.

\bibitem[Hinton et~al., 2012b]{Hinton2012}
Hinton, G.~E., Srivastava, N., Krizhevsky, A., Sutskever, I., and
  Salakhutdinov, R.~R. (2012b).
\newblock Improving neural networks by preventing co-adaptation of feature
  detectors.
\newblock Technical Report arXiv:1207.0580.

\bibitem[Hinton and van Camp, 1993]{Hinton:93}
Hinton, G.~E. and van Camp, D. (1993).
\newblock Keeping neural networks simple.
\newblock In {\em Proceedings of the International Conference on Artificial
  Neural Networks, Amsterdam}, pages 11--18. Springer.

\bibitem[Hochreiter, 1991]{Hochreiter:91}
Hochreiter, S. (1991).
\newblock {Untersuchungen zu dynamischen neuronalen Netzen. Diploma thesis,
  Institut f\"{u}r Informatik, Lehrstuhl Prof. Brauer, Technische
  Universit\"{a}t M\"{u}nchen}.
\newblock Advisor: J. Schmidhuber.

\bibitem[Hochreiter et~al., 2001a]{Hochreiter:01book}
Hochreiter, S., Bengio, Y., Frasconi, P., and Schmidhuber, J. (2001a).
\newblock Gradient flow in recurrent nets: the difficulty of learning long-term
  dependencies.
\newblock In Kremer, S.~C. and Kolen, J.~F., editors, {\em A Field Guide to
  Dynamical Recurrent Neural Networks}. IEEE Press.

\bibitem[Hochreiter and Obermayer, 2005]{hochreiter:snowbird}
Hochreiter, S. and Obermayer, K. (2005).
\newblock Sequence classification for protein analysis.
\newblock In {\em Snowbird Workshop}, Snowbird, Utah. Computational and
  Biological Learning Society.

\bibitem[Hochreiter and Schmidhuber, 1996]{Hochreiter:96sintra}
Hochreiter, S. and Schmidhuber, J. (1996).
\newblock Bridging long time lags by weight guessing and {``Long Short-Term
  Memory''}.
\newblock In Silva, F.~L., Principe, J.~C., and Almeida, L.~B., editors, {\em
  Spatiotemporal models in biological and artificial systems}, pages 65--72.
  IOS Press, Amsterdam, Netherlands.
\newblock Serie: Frontiers in Artificial Intelligence and Applications, Volume
  37.

\bibitem[Hochreiter and Schmidhuber, 1997a]{Hochreiter:97nc1}
Hochreiter, S. and Schmidhuber, J. (1997a).
\newblock Flat minima.
\newblock {\em Neural Computation}, 9(1):1--42.

\bibitem[Hochreiter and Schmidhuber, 1997b]{lstm97and95}
Hochreiter, S. and Schmidhuber, J. (1997b).
\newblock {Long Short-Term Memory}.
\newblock {\em Neural Computation}, 9(8):1735--1780.
\newblock Based on TR FKI-207-95, TUM (1995).

\bibitem[Hochreiter and Schmidhuber, 1999]{Hochreiter:99nc}
Hochreiter, S. and Schmidhuber, J. (1999).
\newblock Feature extraction through {LOCOCODE}.
\newblock {\em Neural Computation}, 11(3):679--714.

\bibitem[Hochreiter et~al., 2001b]{Hochreiter:01meta}
Hochreiter, S., Younger, A.~S., and Conwell, P.~R. (2001b).
\newblock Learning to learn using gradient descent.
\newblock In {\em Lecture Notes on Comp. Sci. 2130, Proc. Intl. Conf. on
  Artificial Neural Networks (ICANN-2001)}, pages 87--94. Springer: Berlin,
  Heidelberg.

\bibitem[Hodgkin and Huxley, 1952]{hodgkin1952}
Hodgkin, A.~L. and Huxley, A.~F. (1952).
\newblock A quantitative description of membrane current and its application to
  conduction and excitation in nerve.
\newblock {\em The Journal of Physiology}, 117(4):500.

\bibitem[Hoerzer et~al., 2014]{maass2014}
Hoerzer, G.~M., Legenstein, R., and Maass, W. (2014).
\newblock Emergence of complex computational structures from chaotic neural
  networks through reward-modulated {Hebbian} learning.
\newblock {\em Cerebral Cortex}, 24:677--690.

\bibitem[Holden, 1994]{Holden:94}
Holden, S.~B. (1994).
\newblock {\em On the Theory of Generalization and Self-Structuring in Linearly
  Weighted Connectionist Networks}.
\newblock PhD thesis, Cambridge University, Engineering Department.

\bibitem[Holland, 1975]{Holland:75}
Holland, J.~H. (1975).
\newblock {\em Adaptation in Natural and Artificial Systems}.
\newblock University of Michigan Press, Ann Arbor.

\bibitem[Honavar and Uhr, 1993]{honavar1993}
Honavar, V. and Uhr, L. (1993).
\newblock Generative learning structures and processes for generalized
  connectionist networks.
\newblock {\em Information Sciences}, 70(1):75--108.

\bibitem[Honavar and Uhr, 1988]{honavar1988}
Honavar, V. and Uhr, L.~M. (1988).
\newblock A network of neuron-like units that learns to perceive by generation
  as well as reweighting of its links.
\newblock In Touretzky, D., Hinton, G.~E., and Sejnowski, T., editors, {\em
  Proc. of the 1988 Connectionist Models Summer School}, pages 472--484, San
  Mateo. Morgan Kaufman.

\bibitem[Hopfield, 1982]{Hopfield:82}
Hopfield, J.~J. (1982).
\newblock Neural networks and physical systems with emergent collective
  computational abilities.
\newblock {\em Proc. of the National Academy of Sciences}, 79:2554--2558.

\bibitem[Hornik et~al., 1989]{hornik1989}
Hornik, K., Stinchcombe, M., and White, H. (1989).
\newblock Multilayer feedforward networks are universal approximators.
\newblock {\em Neural Networks}, 2(5):359--366.

\bibitem[Hubel and Wiesel, 1962]{Hubel:62}
Hubel, D.~H. and Wiesel, T. (1962).
\newblock Receptive fields, binocular interaction, and functional architecture
  in the cat's visual cortex.
\newblock {\em Journal of Physiology (London)}, 160:106--154.

\bibitem[Hubel and Wiesel, 1968]{hubel1968}
Hubel, D.~H. and Wiesel, T.~N. (1968).
\newblock Receptive fields and functional architecture of monkey striate
  cortex.
\newblock {\em The Journal of Physiology}, 195(1):215--243.

\bibitem[Huffman, 1952]{Huffman:52}
Huffman, D.~A. (1952).
\newblock A method for construction of minimum-redundancy codes.
\newblock {\em Proceedings IRE}, 40:1098--1101.

\bibitem[Hung et~al., 2005]{poggio2005}
Hung, C.~P., Kreiman, G., Poggio, T., and DiCarlo, J.~J. (2005).
\newblock Fast readout of object identity from macaque inferior temporal
  cortex.
\newblock {\em Science}, 310(5749):863--866.

\bibitem[Hutter, 2002]{Hutter:01fast+}
Hutter, M. (2002).
\newblock The fastest and shortest algorithm for all well-defined problems.
\newblock {\em International Journal of Foundations of Computer Science},
  13(3):431--443.
\newblock (On J. Schmidhuber's SNF grant 20-61847).

\bibitem[Hutter, 2005]{Hutter:05book+}
Hutter, M. (2005).
\newblock {\em Universal Artificial Intelligence: Sequential Decisions based on
  Algorithmic Probability}.
\newblock Springer, Berlin.
\newblock (On J. Schmidhuber's SNF grant 20-61847).

\bibitem[Hyv{\"a}rinen et~al., 1999]{Hyvarinen:99}
Hyv{\"a}rinen, A., Hoyer, P., and Oja, E. (1999).
\newblock Sparse code shrinkage: Denoising by maximum likelihood estimation.
\newblock In Kearns, M., Solla, S.~A., and Cohn, D., editors, {\em Advances in
  Neural Information Processing Systems (NIPS) 12}. MIT Press.

\bibitem[Hyv{\"a}rinen et~al., 2001]{hyvarinen2001}
Hyv{\"a}rinen, A., Karhunen, J., and Oja, E. (2001).
\newblock {\em Independent component analysis}.
\newblock John Wiley \& Sons.

\bibitem[{ICPR 2012 Contest on Mitosis Detection in Breast Cancer Histological
  Images}, 2012]{icpr12}
{ICPR 2012 Contest on Mitosis Detection in Breast Cancer Histological Images}
  (2012).
\newblock {IPAL Laboratory and TRIBVN Company and Pitie-Salpetriere Hospital
  and CIALAB of Ohio State Univ., http://ipal.cnrs.fr/ICPR2012/}.

\bibitem[Igel, 2003]{igel:cec03}
Igel, C. (2003).
\newblock Neuroevolution for reinforcement learning using evolution strategies.
\newblock In Reynolds, R., Abbass, H., Tan, K.~C., Mckay, B., Essam, D., and
  Gedeon, T., editors, {\em Congress on Evolutionary Computation (CEC 2003)},
  volume~4, pages 2588--2595. IEEE.

\bibitem[Igel and H\"usken, 2003]{igel:01}
Igel, C. and H\"usken, M. (2003).
\newblock Empirical evaluation of the improved {R}prop learning algorithm.
\newblock {\em Neurocomputing}, 50(C):105--123.

\bibitem[Ikeda et~al., 1976]{ikeda1976}
Ikeda, S., Ochiai, M., and Sawaragi, Y. (1976).
\newblock Sequential {GMDH} algorithm and its application to river flow
  prediction.
\newblock {\em IEEE Transactions on Systems, Man and Cybernetics},
  (7):473--479.

\bibitem[Indermuhle et~al., 2012]{indermuhle2012mode}
Indermuhle, E., Frinken, V., and Bunke, H. (2012).
\newblock Mode detection in online handwritten documents using {BLSTM} neural
  networks.
\newblock In {\em Frontiers in Handwriting Recognition (ICFHR), 2012
  International Conference on}, pages 302--307. IEEE.

\bibitem[Indermuhle et~al., 2011]{indermuhle2011keyword}
Indermuhle, E., Frinken, V., Fischer, A., and Bunke, H. (2011).
\newblock Keyword spotting in online handwritten documents containing text and
  non-text using {BLSTM} neural networks.
\newblock In {\em Document Analysis and Recognition (ICDAR), 2011 International
  Conference on}, pages 73--77. IEEE.

\bibitem[Indiveri et~al., 2011]{indiveri2011}
Indiveri, G., Linares-Barranco, B., Hamilton, T.~J., Van~Schaik, A.,
  Etienne-Cummings, R., Delbruck, T., Liu, S.-C., Dudek, P., H{\"a}fliger, P.,
  Renaud, S., et~al. (2011).
\newblock Neuromorphic silicon neuron circuits.
\newblock {\em Frontiers in Neuroscience}, 5(73).

\bibitem[Ivakhnenko, 1968]{ivakhnenko1968}
Ivakhnenko, A.~G. (1968).
\newblock The group method of data handling -- a rival of the method of
  stochastic approximation.
\newblock {\em Soviet Automatic Control}, 13(3):43--55.

\bibitem[Ivakhnenko, 1971]{ivakhnenko1971}
Ivakhnenko, A.~G. (1971).
\newblock Polynomial theory of complex systems.
\newblock {\em IEEE Transactions on Systems, Man and Cybernetics},
  (4):364--378.

\bibitem[Ivakhnenko, 1995]{ivakhnenko1995}
Ivakhnenko, A.~G. (1995).
\newblock The review of problems solvable by algorithms of the group method of
  data handling {(GMDH)}.
\newblock {\em Pattern Recognition and Image Analysis / Raspoznavaniye Obrazov
  I Analiz Izobrazhenii}, 5:527--535.

\bibitem[Ivakhnenko and Lapa, 1965]{ivakhnenko1965}
Ivakhnenko, A.~G. and Lapa, V.~G. (1965).
\newblock {\em Cybernetic Predicting Devices}.
\newblock CCM Information Corporation.

\bibitem[Ivakhnenko et~al., 1967]{ivakhnenko1967}
Ivakhnenko, A.~G., Lapa, V.~G., and McDonough, R.~N. (1967).
\newblock {\em Cybernetics and forecasting techniques}.
\newblock American Elsevier, NY.

\bibitem[Izhikevich et~al., 2003]{izhikevich2003}
Izhikevich, E.~M. et~al. (2003).
\newblock Simple model of spiking neurons.
\newblock {\em IEEE Transactions on Neural Networks}, 14(6):1569--1572.

\bibitem[Jaakkola et~al., 1995]{Jaakkola:95}
Jaakkola, T., Singh, S.~P., and Jordan, M.~I. (1995).
\newblock Reinforcement learning algorithm for partially observable {Markov}
  decision problems.
\newblock In Tesauro, G., Touretzky, D.~S., and Leen, T.~K., editors, {\em
  Advances in Neural Information Processing Systems (NIPS) 7}, pages 345--352.
  MIT Press.

\bibitem[Jackel et~al., 1990]{jackel-90}
Jackel, L., Boser, B., Graf, H.-P., Denker, J., LeCun, Y., Henderson, D.,
  Matan, O., Howard, R., and Baird, H. (1990).
\newblock {VLSI} implementation of electronic neural networks: and example in
  character recognition.
\newblock In IEEE, editor, {\em IEEE International Conference on Systems, Man,
  and Cybernetics}, pages 320--322, Los Angeles, CA.

\bibitem[Jacob et~al., 1994]{lindenmayer94}
Jacob, C., Lindenmayer, A., and Rozenberg, G. (1994).
\newblock {Genetic L-System Programming}.
\newblock In {\em Parallel Problem Solving from Nature III}, Lecture Notes in
  Computer Science.

\bibitem[Jacobs, 1988]{Jacobs:88}
Jacobs, R.~A. (1988).
\newblock Increased rates of convergence through learning rate adaptation.
\newblock {\em Neural Networks}, 1(4):295--307.

\bibitem[Jaeger, 2001]{Jaeger2001a}
Jaeger, H. (2001).
\newblock The "echo state" approach to analysing and training recurrent neural
  networks.
\newblock Technical Report GMD Report 148, German National Research Center for
  Information Technology.

\bibitem[Jaeger, 2004]{Jaeger:04}
Jaeger, H. (2004).
\newblock Harnessing nonlinearity: Predicting chaotic systems and saving energy
  in wireless communication.
\newblock {\em Science}, 304:78--80.

\bibitem[Jain and Seung, 2009]{seung2009}
Jain, V. and Seung, S. (2009).
\newblock Natural image denoising with convolutional networks.
\newblock In Koller, D., Schuurmans, D., Bengio, Y., and Bottou, L., editors,
  {\em Advances in Neural Information Processing Systems (NIPS) 21}, pages
  769--776. Curran Associates, Inc.

\bibitem[Jameson, 1991]{Jameson:91}
Jameson, J. (1991).
\newblock Delayed reinforcement learning with multiple time scale hierarchical
  backpropagated adaptive critics.
\newblock In {\em Neural Networks for Control}.

\bibitem[Ji et~al., 2013]{ji2013}
Ji, S., Xu, W., Yang, M., and Yu, K. (2013).
\newblock 3{D} convolutional neural networks for human action recognition.
\newblock {\em IEEE Transactions on Pattern Analysis and Machine Intelligence},
  35(1):221--231.

\bibitem[Jim et~al., 1995]{giles95}
Jim, K., Giles, C.~L., and Horne, B.~G. (1995).
\newblock Effects of noise on convergence and generalization in recurrent
  networks.
\newblock In Tesauro, G., Touretzky, D., and Leen, T., editors, {\em Advances
  in Neural Information Processing Systems (NIPS) 7}, page 649. San Mateo, CA:
  Morgan Kaufmann.

\bibitem[Jin et~al., 2010]{jin2010}
Jin, X., Lujan, M., Plana, L.~A., Davies, S., Temple, S., and Furber, S.~B.
  (2010).
\newblock Modeling spiking neural networks on {SpiNNaker}.
\newblock {\em Computing in Science \& Engineering}, 12(5):91--97.

\bibitem[Jodogne and Piater, 2007]{Jodogne07}
Jodogne, S.~R. and Piater, J.~H. (2007).
\newblock Closed-loop learning of visual control policies.
\newblock {\em J. Artificial Intelligence Research}, 28:349--391.

\bibitem[Jones and Palmer, 1987]{jones1987}
Jones, J.~P. and Palmer, L.~A. (1987).
\newblock An evaluation of the two-dimensional {Gabor} filter model of simple
  receptive fields in cat striate cortex.
\newblock {\em Journal of Neurophysiology}, 58(6):1233--1258.

\bibitem[Jordan, 1986]{Jordan:86}
Jordan, M.~I. (1986).
\newblock Serial order: A parallel distributed processing approach.
\newblock Technical Report ICS Report 8604, Institute for Cognitive Science,
  University of California, San Diego.

\bibitem[Jordan, 1988]{Jordan:88}
Jordan, M.~I. (1988).
\newblock Supervised learning and systems with excess degrees of freedom.
\newblock Technical Report COINS TR 88-27, Massachusetts Institute of
  Technology.

\bibitem[Jordan, 1997]{jordan1997}
Jordan, M.~I. (1997).
\newblock Serial order: A parallel distributed processing approach.
\newblock {\em Advances in Psychology}, 121:471--495.

\bibitem[Jordan and Rumelhart, 1990]{JordanRumelhart:90}
Jordan, M.~I. and Rumelhart, D.~E. (1990).
\newblock Supervised learning with a distal teacher.
\newblock Technical Report Occasional Paper \#40, Center for Cog. Sci.,
  Massachusetts Institute of Technology.

\bibitem[Jordan and Sejnowski, 2001]{jordan2001}
Jordan, M.~I. and Sejnowski, T.~J. (2001).
\newblock {\em Graphical models: Foundations of neural computation}.
\newblock MIT Press.

\bibitem[Joseph, 1961]{joseph1961}
Joseph, R.~D. (1961).
\newblock {\em Contributions to perceptron theory}.
\newblock PhD thesis, Cornell Univ.

\bibitem[Juang, 2004]{juang2004}
Juang, C.-F. (2004).
\newblock A hybrid of genetic algorithm and particle swarm optimization for
  recurrent network design.
\newblock {\em Systems, Man, and Cybernetics, Part B: Cybernetics, IEEE
  Transactions on}, 34(2):997--1006.

\bibitem[Judd, 1990]{judd1990}
Judd, J.~S. (1990).
\newblock {\em Neural network design and the complexity of learning}.
\newblock Neural network modeling and connectionism. MIT Press.

\bibitem[Jutten and Herault, 1991]{Jutten:91}
Jutten, C. and Herault, J. (1991).
\newblock Blind separation of sources, part {I}: {A}n adaptive algorithm based
  on neuromimetic architecture.
\newblock {\em Signal Processing}, 24(1):1--10.

\bibitem[Kaelbling et~al., 1995]{Kaelbling:95}
Kaelbling, L.~P., Littman, M.~L., and Cassandra, A.~R. (1995).
\newblock Planning and acting in partially observable stochastic domains.
\newblock Technical report, Brown University, Providence RI.

\bibitem[Kaelbling et~al., 1996]{Kaelbling:96}
Kaelbling, L.~P., Littman, M.~L., and Moore, A.~W. (1996).
\newblock Reinforcement learning: a survey.
\newblock {\em Journal of AI research}, 4:237--285.

\bibitem[Kak et~al., 2010]{kak2010}
Kak, S., Chen, Y., and Wang, L. (2010).
\newblock Data mining using surface and deep agents based on neural networks.
\newblock {\em AMCIS 2010 Proceedings}.

\bibitem[Kalinke and Lehmann, 1998]{kalinke98computation}
Kalinke, Y. and Lehmann, H. (1998).
\newblock Computation in recurrent neural networks: From counters to iterated
  function systems.
\newblock In Antoniou, G. and Slaney, J., editors, {\em {Advanced Topics in
  Artificial Intelligence, Proceedings of the 11th Australian Joint Conference
  on Artificial Intelligence}}, volume 1502 of {\em LNAI}, Berlin, Heidelberg.
  Springer.

\bibitem[Kalman, 1960]{kalman1960}
Kalman, R.~E. (1960).
\newblock A new approach to linear filtering and prediction problems.
\newblock {\em Journal of Basic Engineering}, 82(1):35--45.

\bibitem[Karhunen and Joutsensalo, 1995]{karhunen1995}
Karhunen, J. and Joutsensalo, J. (1995).
\newblock Generalizations of principal component analysis, optimization
  problems, and neural networks.
\newblock {\em Neural Networks}, 8(4):549--562.

\bibitem[Karpathy et~al., 2014]{karpathy2014}
Karpathy, A., Toderici, G., Shetty, S., Leung, T., Sukthankar, R., and Fei-Fei,
  L. (2014).
\newblock Large-scale video classification with convolutional neural networks.
\newblock In {\em IEEE Conference on Computer Vision and Pattern Recognition
  (CVPR)}.

\bibitem[Kasabov, 2014]{kasabov2014}
Kasabov, N.~K. (2014).
\newblock Neucube: A spiking neural network architecture for mapping, learning
  and understanding of spatio-temporal brain data.
\newblock {\em Neural Networks}.

\bibitem[Kelley, 1960]{Kelley:1960}
Kelley, H.~J. (1960).
\newblock Gradient theory of optimal flight paths.
\newblock {\em ARS Journal}, 30(10):947--954.

\bibitem[Kempter et~al., 1999]{kempter1999}
Kempter, R., Gerstner, W., and Van~Hemmen, J.~L. (1999).
\newblock Hebbian learning and spiking neurons.
\newblock {\em Physical Review E}, 59(4):4498.

\bibitem[Kerlirzin and Vallet, 1993]{Kerlirzin:93}
Kerlirzin, P. and Vallet, F. (1993).
\newblock Robustness in multilayer perceptrons.
\newblock {\em Neural Computation}, 5(1):473--482.

\bibitem[Khan et~al., 2010]{khan2010}
Khan, M.~M., Khan, G.~M., and Miller, J.~F. (2010).
\newblock Evolution of neural networks using {Cartesian Genetic Programming}.
\newblock In {\em IEEE Congress on Evolutionary Computation (CEC)}, pages 1--8.

\bibitem[Khan et~al., 2008]{khan2008}
Khan, M.~M., Lester, D.~R., Plana, L.~A., Rast, A., Jin, X., Painkras, E., and
  Furber, S.~B. (2008).
\newblock {SpiNNaker:} mapping neural networks onto a massively-parallel chip
  multiprocessor.
\newblock In {\em International Joint Conference on Neural Networks (IJCNN)},
  pages 2849--2856. IEEE.

\bibitem[Khan et~al., 2014]{khan2014}
Khan, S.~H., Bennamoun, M., Sohel, F., and Togneri, R. (2014).
\newblock Automatic feature learning for robust shadow detection.
\newblock In {\em IEEE Conference on Computer Vision and Pattern Recognition
  CVPR}.

\bibitem[Kimura et~al., 1997]{kimura1997}
Kimura, H., Miyazaki, K., and Kobayashi, S. (1997).
\newblock Reinforcement learning in {POMDPs} with function approximation.
\newblock In {\em ICML}, volume~97, pages 152--160.

\bibitem[Kistler et~al., 1997]{kistler1997}
Kistler, W.~M., Gerstner, W., and van Hemmen, J.~L. (1997).
\newblock Reduction of the {Hodgkin-Huxley} equations to a single-variable
  threshold model.
\newblock {\em Neural Computation}, 9(5):1015--1045.

\bibitem[Kitano, 1990]{kitano90}
Kitano, H. (1990).
\newblock Designing neural networks using genetic algorithms with graph
  generation system.
\newblock {\em Complex Systems}, 4:461--476.

\bibitem[Klampfl and Maass, 2013]{maass2013}
Klampfl, S. and Maass, W. (2013).
\newblock Emergence of dynamic memory traces in cortical microcircuit models
  through {STDP}.
\newblock {\em The Journal of Neuroscience}, 33(28):11515--11529.

\bibitem[Klapper-Rybicka et~al., 2001]{Klapper:01}
Klapper-Rybicka, M., Schraudolph, N.~N., and Schmidhuber, J. (2001).
\newblock Unsupervised learning in {LSTM} recurrent neural networks.
\newblock In {\em Lecture Notes on Comp. Sci. 2130, Proc. Intl. Conf. on
  Artificial Neural Networks (ICANN-2001)}, pages 684--691. Springer: Berlin,
  Heidelberg.

\bibitem[Kobatake and Tanaka, 1994]{Tanaka:94}
Kobatake, E. and Tanaka, K. (1994).
\newblock Neuronal selectivities to complex object features in the ventral
  visual pathway of the macaque cerebral cortex.
\newblock {\em J. Neurophysiol.}, 71:856--867.

\bibitem[Kohl and Stone, 2004]{stoneICRA04}
Kohl, N. and Stone, P. (2004).
\newblock Policy gradient reinforcement learning for fast quadrupedal
  locomotion.
\newblock In {\em Robotics and Automation, 2004. Proceedings. ICRA'04. 2004
  IEEE International Conference on}, volume~3, pages 2619--2624. IEEE.

\bibitem[Kohonen, 1972]{kohonen1972}
Kohonen, T. (1972).
\newblock Correlation matrix memories.
\newblock {\em Computers, IEEE Transactions on}, 100(4):353--359.

\bibitem[Kohonen, 1982]{Kohonen:82}
Kohonen, T. (1982).
\newblock Self-organized formation of topologically correct feature maps.
\newblock {\em Biological Cybernetics}, 43(1):59--69.

\bibitem[Kohonen, 1988]{Kohonen:88}
Kohonen, T. (1988).
\newblock {\em Self-Organization and Associative Memory}.
\newblock Springer, second edition.

\bibitem[Koikkalainen and Oja, 1990]{koikkalainen1990}
Koikkalainen, P. and Oja, E. (1990).
\newblock Self-organizing hierarchical feature maps.
\newblock In {\em International Joint Conference on Neural Networks (IJCNN)},
  pages 279--284. IEEE.

\bibitem[Kolmogorov, 1965a]{Kolmogorov:57}
Kolmogorov, A.~N. (1965a).
\newblock On the representation of continuous functions of several variables by
  superposition of continuous functions of one variable and addition.
\newblock {\em Doklady Akademii. Nauk USSR,}, 114:679--681.

\bibitem[Kolmogorov, 1965b]{Kolmogorov:65}
Kolmogorov, A.~N. (1965b).
\newblock Three approaches to the quantitative definition of information.
\newblock {\em Problems of Information Transmission}, 1:1--11.

\bibitem[Kompella et~al., 2012]{DBLP:journals/neco/KompellaLS12}
Kompella, V.~R., Luciw, M.~D., and Schmidhuber, J. (2012).
\newblock Incremental slow feature analysis: Adaptive low-complexity slow
  feature updating from high-dimensional input streams.
\newblock {\em Neural Computation}, 24(11):2994--3024.

\bibitem[Kondo, 1998]{kondo1998}
Kondo, T. (1998).
\newblock {GMDH} neural network algorithm using the heuristic self-organization
  method and its application to the pattern identification problem.
\newblock In {\em Proceedings of the 37th SICE Annual Conference SICE'98},
  pages 1143--1148. IEEE.

\bibitem[Kondo and Ueno, 2008]{kondo2008}
Kondo, T. and Ueno, J. (2008).
\newblock Multi-layered {GMDH}-type neural network self-selecting optimum
  neural network architecture and its application to 3-dimensional medical
  image recognition of blood vessels.
\newblock {\em International Journal of Innovative Computing, Information and
  Control}, 4(1):175--187.

\bibitem[Kord{\'\i}k et~al., 2003]{kordik2003}
Kord{\'\i}k, P., N{\'a}plava, P., Snorek, M., and Genyk-Berezovskyj, M. (2003).
\newblock Modified {GMDH} method and models quality evaluation by
  visualization.
\newblock {\em Control Systems and Computers}, 2:68--75.

\bibitem[Korkin et~al., 1997]{cbm97}
Korkin, M., de~Garis, H., Gers, F., and Hemmi, H. (1997).
\newblock {CBM (CAM-Brain Machine)} - a hardware tool which evolves a neural
  net module in a fraction of a second and runs a million neuron artificial
  brain in real time.

\bibitem[Kosko, 1990]{kosko1990}
Kosko, B. (1990).
\newblock Unsupervised learning in noise.
\newblock {\em IEEE Transactions on Neural Networks}, 1(1):44--57.

\bibitem[Koutn{\'i}k et~al., 2013]{koutnik:gecco13}
Koutn{\'i}k, J., Cuccu, G., Schmidhuber, J., and Gomez, F. (July 2013).
\newblock Evolving large-scale neural networks for vision-based reinforcement
  learning.
\newblock In {\em Proceedings of the Genetic and Evolutionary Computation
  Conference (GECCO)}, pages 1061--1068, Amsterdam. ACM.

\bibitem[Koutn{\'i}k et~al., 2010]{koutnik:gecco10}
Koutn{\'i}k, J., Gomez, F., and Schmidhuber, J. (2010).
\newblock Evolving neural networks in compressed weight space.
\newblock In {\em Proceedings of the 12th Annual Conference on Genetic and
  Evolutionary Computation}, pages 619--626.

\bibitem[Koutn{\'i}k et~al., 2014]{icml2014}
Koutn{\'i}k, J., Greff, K., Gomez, F., and Schmidhuber, J. (2014).
\newblock {A Clockwork RNN}.
\newblock In {\em Proceedings of the 31th International Conference on Machine
  Learning (ICML)}, volume~32, pages 1845--1853.
\newblock arXiv:1402.3511 [cs.NE].

\bibitem[Koza, 1992]{Koza:92}
Koza, J.~R. (1992).
\newblock {\em Genetic Programming -- On the Programming of Computers by Means
  of Natural Selection}.
\newblock MIT Press.

\bibitem[Kramer, 1991]{Kramer:91}
Kramer, M. (1991).
\newblock Nonlinear principal component analysis using autoassociative neural
  networks.
\newblock {\em AIChE Journal}, 37:233--243.

\bibitem[Kremer and Kolen, 2001]{kremer2001}
Kremer, S.~C. and Kolen, J.~F. (2001).
\newblock {\em Field guide to dynamical recurrent networks}.
\newblock Wiley-IEEE Press.

\bibitem[Kriegeskorte et~al., 2008]{kriegeskorte2008}
Kriegeskorte, N., Mur, M., Ruff, D.~A., Kiani, R., Bodurka, J., Esteky, H.,
  Tanaka, K., and Bandettini, P.~A. (2008).
\newblock Matching categorical object representations in inferior temporal
  cortex of man and monkey.
\newblock {\em Neuron}, 60(6):1126--1141.

\bibitem[Krizhevsky et~al., 2012]{Krizhevsky:2012}
Krizhevsky, A., Sutskever, I., and Hinton, G.~E. (2012).
\newblock Imagenet classification with deep convolutional neural networks.
\newblock In {\em Advances in Neural Information Processing Systems (NIPS
  2012)}, page~4.

\bibitem[Krogh and Hertz, 1992]{Krogh:92}
Krogh, A. and Hertz, J.~A. (1992).
\newblock A simple weight decay can improve generalization.
\newblock In Lippman, D.~S., Moody, J.~E., and Touretzky, D.~S., editors, {\em
  Advances in Neural Information Processing Systems 4}, pages 950--957. Morgan
  Kaufmann.

\bibitem[Kruger et~al., 2013]{kruger2013}
Kruger, N., Janssen, P., Kalkan, S., Lappe, M., Leonardis, A., Piater, J.,
  Rodriguez-Sanchez, A., and Wiskott, L. (2013).
\newblock Deep hierarchies in the primate visual cortex: What can we learn for
  computer vision?
\newblock {\em IEEE Transactions on Pattern Analysis and Machine Intelligence},
  35(8):1847--1871.

\bibitem[Kullback and Leibler, 1951]{kullback1951}
Kullback, S. and Leibler, R.~A. (1951).
\newblock On information and sufficiency.
\newblock {\em The Annals of Mathematical Statistics}, pages 79--86.

\bibitem[Kurzweil, 2012]{kurzweil2012}
Kurzweil, R. (2012).
\newblock {\em How to Create a Mind: The Secret of Human Thought Revealed}.

\bibitem[Lagoudakis and Parr, 2003]{03-LspiLagoudakis}
Lagoudakis, M.~G. and Parr, R. (2003).
\newblock Least-squares policy iteration.
\newblock {\em JMLR}, 4:1107--1149.

\bibitem[Lampinen and Oja, 1992]{lampinen1992}
Lampinen, J. and Oja, E. (1992).
\newblock Clustering properties of hierarchical self-organizing maps.
\newblock {\em Journal of Mathematical Imaging and Vision}, 2(2-3):261--272.

\bibitem[Lang et~al., 1990]{Lang:90}
Lang, K., Waibel, A., and Hinton, G.~E. (1990).
\newblock A time-delay neural network architecture for isolated word
  recognition.
\newblock {\em Neural Networks}, 3:23--43.

\bibitem[Lange and Riedmiller, 2010]{lange}
Lange, S. and Riedmiller, M. (2010).
\newblock Deep auto-encoder neural networks in reinforcement learning.
\newblock In {\em Neural Networks (IJCNN), The 2010 International Joint
  Conference on}, pages 1--8.

\bibitem[Lapedes and Farber, 1986]{Lapedes:86a}
Lapedes, A. and Farber, R. (1986).
\newblock A self-optimizing, nonsymmetrical neural net for content addressable
  memory and pattern recognition.
\newblock {\em Physica D}, 22:247--259.

\bibitem[Laplace, 1774]{laplace1774}
Laplace, P. (1774).
\newblock M{\'e}moire sur la probabilit{\'e} des causes par les
  {\'e}v{\`e}nements.
\newblock {\em M{\'e}moires de l'Academie Royale des Sciences Present{\'e}s par
  Divers Savan}, 6:621--656.

\bibitem[Larraanaga and Lozano, 2001]{Larraanaga2001}
Larraanaga, P. and Lozano, J.~A. (2001).
\newblock {\em Estimation of Distribution Algorithms: A New Tool for
  Evolutionary Computation}.
\newblock Kluwer Academic Publishers, Norwell, MA, USA.

\bibitem[Le et~al., 2012]{ng2012}
Le, Q.~V., Ranzato, M., Monga, R., Devin, M., Corrado, G., Chen, K., Dean, J.,
  and Ng, A.~Y. (2012).
\newblock Building high-level features using large scale unsupervised learning.
\newblock In {\em Proc. ICML'12}.

\bibitem[LeCun, 1985]{LeCun:85}
LeCun, Y. (1985).
\newblock Une proc\'{e}dure d'apprentissage pour r\'{e}seau \`{a} seuil
  asym\'{e}trique.
\newblock {\em Proceedings of Cognitiva 85, Paris}, pages 599--604.

\bibitem[LeCun, 1988]{lecun-88}
LeCun, Y. (1988).
\newblock A theoretical framework for back-propagation.
\newblock In Touretzky, D., Hinton, G., and Sejnowski, T., editors, {\em
  Proceedings of the 1988 Connectionist Models Summer School}, pages 21--28,
  CMU, Pittsburgh, Pa. Morgan Kaufmann.

\bibitem[LeCun et~al., 1989]{LeCun:89}
LeCun, Y., Boser, B., Denker, J.~S., Henderson, D., Howard, R.~E., Hubbard, W.,
  and Jackel, L.~D. (1989).
\newblock Back-propagation applied to handwritten zip code recognition.
\newblock {\em Neural Computation}, 1(4):541--551.

\bibitem[LeCun et~al., 1990a]{LeCun:90}
LeCun, Y., Boser, B., Denker, J.~S., Henderson, D., Howard, R.~E., Hubbard, W.,
  and Jackel, L.~D. (1990a).
\newblock Handwritten digit recognition with a back-propagation network.
\newblock In Touretzky, D.~S., editor, {\em Advances in Neural Information
  Processing Systems 2}, pages 396--404. Morgan Kaufmann.

\bibitem[LeCun et~al., 1998]{LeCun:98}
LeCun, Y., Bottou, L., Bengio, Y., and Haffner, P. (1998).
\newblock Gradient-based learning applied to document recognition.
\newblock {\em Proceedings of the IEEE}, 86(11):2278--2324.

\bibitem[LeCun et~al., 1990b]{LeCun:90a}
LeCun, Y., Denker, J.~S., and Solla, S.~A. (1990b).
\newblock Optimal brain damage.
\newblock In Touretzky, D.~S., editor, {\em Advances in Neural Information
  Processing Systems 2}, pages 598--605. Morgan Kaufmann.

\bibitem[LeCun et~al., 2006]{LeCun:06}
LeCun, Y., Muller, U., Cosatto, E., and Flepp, B. (2006).
\newblock Off-road obstacle avoidance through end-to-end learning.
\newblock In {\em Advances in Neural Information Processing Systems (NIPS
  2005)}.

\bibitem[LeCun et~al., 1993]{lecun-simard-pearlmutter-93}
LeCun, Y., Simard, P., and Pearlmutter, B. (1993).
\newblock Automatic learning rate maximization by on-line estimation of the
  {Hessian}'s eigenvectors.
\newblock In Hanson, S., Cowan, J., and Giles, L., editors, {\em Advances in
  Neural Information Processing Systems (NIPS 1992)}, volume~5. Morgan Kaufmann
  Publishers, San Mateo, CA.

\bibitem[Lee et~al., 2007a]{sparse2007ng}
Lee, H., Battle, A., Raina, R., and Ng, A.~Y. (2007a).
\newblock Efficient sparse coding algorithms.
\newblock In {\em Advances in Neural Information Processing Systems (NIPS) 19},
  pages 801--808.

\bibitem[Lee et~al., 2007b]{lee2007sparse}
Lee, H., Ekanadham, C., and Ng, A.~Y. (2007b).
\newblock Sparse deep belief net model for visual area {V2}.
\newblock In {\em Advances in Neural Information Processing Systems (NIPS)},
  volume~7, pages 873--880.

\bibitem[Lee et~al., 2009a]{lee:2009}
Lee, H., Grosse, R., Ranganath, R., and Ng, A.~Y. (2009a).
\newblock Convolutional deep belief networks for scalable unsupervised learning
  of hierarchical representations.
\newblock In {\em Proceedings of the 26th International Conference on Machine
  Learning (ICML)}, pages 609--616.

\bibitem[Lee et~al., 2009b]{lee2009audio}
Lee, H., Pham, P.~T., Largman, Y., and Ng, A.~Y. (2009b).
\newblock Unsupervised feature learning for audio classification using
  convolutional deep belief networks.
\newblock In {\em Proc. NIPS}, volume~9, pages 1096--1104.

\bibitem[Lee, 1996]{lee-learning:96}
Lee, L. (1996).
\newblock Learning of context-free languages: A survey of the literature.
\newblock Technical Report TR-12-96, Center for Research in Computing
  Technology, Harvard University, Cambridge, Massachusetts.

\bibitem[Lee and Kil, 1991]{lee1991}
Lee, S. and Kil, R.~M. (1991).
\newblock A {Gaussian} potential function network with hierarchically
  self-organizing learning.
\newblock {\em Neural Networks}, 4(2):207--224.

\bibitem[Legendre, 1805]{legendre1805}
Legendre, A.~M. (1805).
\newblock {\em Nouvelles m{\'e}thodes pour la d{\'e}termination des orbites des
  cometes}.
\newblock F. Didot.

\bibitem[Legenstein et~al., 2010]{Legenstein2010}
Legenstein, R., Wilbert, N., and Wiskott, L. (2010).
\newblock Reinforcement learning on slow features of high-dimensional input
  streams.
\newblock {\em PLoS Computational Biology}, 6(8).

\bibitem[Legenstein and Maass, 2002]{maass2002wire}
Legenstein, R.~A. and Maass, W. (2002).
\newblock Neural circuits for pattern recognition with small total wire length.
\newblock {\em Theor. Comput. Sci.}, 287(1):239--249.

\bibitem[Leibniz, 1676]{leibniz:1676}
Leibniz, G.~W. (1676).
\newblock Memoir using the chain rule (cited in {TMME} 7:2\&3 p 321-332, 2010).

\bibitem[Leibniz, 1684]{leibniz1684}
Leibniz, G.~W. (1684).
\newblock Nova methodus pro maximis et minimis, itemque tangentibus, quae nec
  fractas, nec irrationales quantitates moratur, et singulare pro illis calculi
  genus.
\newblock {\em Acta Eruditorum}, pages 467--473.

\bibitem[Lenat, 1983]{Lenat:83}
Lenat, D.~B. (1983).
\newblock Theory formation by heuristic search.
\newblock {\em Machine Learning}, 21.

\bibitem[Lenat and Brown, 1984]{Lenat:84}
Lenat, D.~B. and Brown, J.~S. (1984).
\newblock {Why AM an EURISKO appear to work}.
\newblock {\em Artificial Intelligence}, 23(3):269--294.

\bibitem[Lennie and Movshon, 2005]{lennie2005}
Lennie, P. and Movshon, J.~A. (2005).
\newblock Coding of color and form in the geniculostriate visual pathway.
\newblock {\em Journal of the Optical Society of America A}, 22(10):2013--2033.

\bibitem[Levenberg, 1944]{levenberg1944}
Levenberg, K. (1944).
\newblock A method for the solution of certain problems in least squares.
\newblock {\em Quarterly of applied mathematics}, 2:164--168.

\bibitem[Levin et~al., 1994]{Levin:94}
Levin, A.~U., Leen, T.~K., and Moody, J.~E. (1994).
\newblock Fast pruning using principal components.
\newblock In {\em Advances in Neural Information Processing Systems 6},
  page~35. Morgan Kaufmann.

\bibitem[Levin and Narendra, 1995]{levin1995}
Levin, A.~U. and Narendra, K.~S. (1995).
\newblock Control of nonlinear dynamical systems using neural networks. ii.
  observability, identification, and control.
\newblock {\em IEEE Transactions on Neural Networks}, 7(1):30--42.

\bibitem[Levin, 1973a]{Levin:73a}
Levin, L.~A. (1973a).
\newblock On the notion of a random sequence.
\newblock {\em Soviet Math. Dokl.}, 14(5):1413--1416.

\bibitem[Levin, 1973b]{Levin:73}
Levin, L.~A. (1973b).
\newblock Universal sequential search problems.
\newblock {\em Problems of Information Transmission}, 9(3):265--266.

\bibitem[Lewicki and Olshausen, 1998]{Lewicki:98b}
Lewicki, M.~S. and Olshausen, B.~A. (1998).
\newblock Inferring sparse, overcomplete image codes using an efficient coding
  framework.
\newblock In Jordan, M.~I., Kearns, M.~J., and Solla, S.~A., editors, {\em
  Advances in Neural Information Processing Systems (NIPS) 10}, pages 815--821.

\bibitem[L'H\^{o}pital, 1696]{de1716analyse}
L'H\^{o}pital, G. F.~A. (1696).
\newblock {\em Analyse des infiniment petits, pour l'intelligence des lignes
  courbes}.
\newblock Paris: L'Imprimerie Royale.

\bibitem[Li and Vit\'{a}nyi, 1997]{LiVitanyi:97}
Li, M. and Vit\'{a}nyi, P. M.~B. (1997).
\newblock {\em An Introduction to {Kolmogorov} Complexity and its Applications
  (2nd edition)}.
\newblock Springer.

\bibitem[Li et~al., 2014]{shuiwang2014}
Li, R., Zhang, W., Suk, H.-I., Wang, L., Li, J., Shen, D., and Ji, S. (2014).
\newblock Deep learning based imaging data completion for improved brain
  disease diagnosis.
\newblock In {\em Proc. MICCAI}. Springer.

\bibitem[Lin, 1993]{Lin:93}
Lin, L. (1993).
\newblock {\em Reinforcement Learning for Robots Using Neural Networks}.
\newblock PhD thesis, Carnegie Mellon University, Pittsburgh.

\bibitem[Lin et~al., 1996]{Lin:96}
Lin, T., Horne, B., Tino, P., and Giles, C. (1996).
\newblock Learning long-term dependencies in {NARX} recurrent neural networks.
\newblock {\em IEEE Transactions on Neural Networks}, 7(6):1329--1338.

\bibitem[Lindenmayer, 1968]{lindenmayer68}
Lindenmayer, A. (1968).
\newblock Mathematical models for cellular interaction in development.
\newblock {\em J. Theoret. Biology}, 18:280--315.

\bibitem[Lindst\"{a}dt, 1993]{Steffi:93cmss}
Lindst\"{a}dt, S. (1993).
\newblock Comparison of two unsupervised neural network models for redundancy
  reduction.
\newblock In Mozer, M.~C., Smolensky, P., Touretzky, D.~S., Elman, J.~L., and
  Weigend, A.~S., editors, {\em Proc. of the 1993 Connectionist Models Summer
  School}, pages 308--315. Hillsdale, NJ: Erlbaum Associates.

\bibitem[Linnainmaa, 1970]{Linnainmaa:1970}
Linnainmaa, S. (1970).
\newblock The representation of the cumulative rounding error of an algorithm
  as a {Taylor} expansion of the local rounding errors.
\newblock Master's thesis, Univ. Helsinki.

\bibitem[Linnainmaa, 1976]{Linnainmaa:1976}
Linnainmaa, S. (1976).
\newblock Taylor expansion of the accumulated rounding error.
\newblock {\em BIT Numerical Mathematics}, 16(2):146--160.

\bibitem[Linsker, 1988]{Linsker:88}
Linsker, R. (1988).
\newblock Self-organization in a perceptual network.
\newblock {\em IEEE Computer}, 21:105--117.

\bibitem[Littman et~al., 1995]{Littman:95}
Littman, M.~L., Cassandra, A.~R., and Kaelbling, L.~P. (1995).
\newblock Learning policies for partially observable environments: Scaling up.
\newblock In Prieditis, A. and Russell, S., editors, {\em Machine Learning:
  Proceedings of the Twelfth International Conference}, pages 362--370. Morgan
  Kaufmann Publishers, San Francisco, CA.

\bibitem[Liu et~al., 2001]{liu2001}
Liu, S.-C., Kramer, J., Indiveri, G., Delbr{\"u}ck, T., Burg, T., Douglas, R.,
  et~al. (2001).
\newblock Orientation-selective {aVLSI} spiking neurons.
\newblock {\em Neural Networks}, 14(6-7):629--643.

\bibitem[Ljung, 1998]{ljung1998}
Ljung, L. (1998).
\newblock {\em System identification}.
\newblock Springer.

\bibitem[Logothetis et~al., 1995]{logothetis1995}
Logothetis, N.~K., Pauls, J., and Poggio, T. (1995).
\newblock Shape representation in the inferior temporal cortex of monkeys.
\newblock {\em Current Biology}, 5(5):552--563.

\bibitem[Loiacono et~al., 2011]{torcs-manual:2011}
Loiacono, D., Cardamone, L., and Lanzi, P.~L. (2011).
\newblock Simulated car racing championship competition software manual.
\newblock Technical report, Dipartimento di Elettronica e Informazione,
  Politecnico di Milano, Italy.

\bibitem[Loiacono et~al., 2009]{wcci:torcs:09}
Loiacono, D., Lanzi, P.~L., Togelius, J., Onieva, E., Pelta, D.~A., Butz,
  M.~V., L\"{o}nneker, T.~D., Cardamone, L., Perez, D., S\'{a}ez, Y., Preuss,
  M., and Quadflieg, J. (2009).
\newblock The 2009 simulated car racing championship.

\bibitem[Lowe, 1999]{Lowe:1999}
Lowe, D. (1999).
\newblock Object recognition from local scale-invariant features.
\newblock In {\em The Proceedings of the Seventh IEEE International Conference
  on Computer Vision (ICCV)}, volume~2, pages 1150--1157.

\bibitem[Lowe, 2004]{Lowe:04}
Lowe, D. (2004).
\newblock Distinctive image features from scale-invariant key-points.
\newblock {\em Intl. Journal of Computer Vision}, 60:91--110.

\bibitem[Luciw et~al., 2013]{luciwkomp13}
Luciw, M., Kompella, V.~R., Kazerounian, S., and Schmidhuber, J. (2013).
\newblock An intrinsic value system for developing multiple invariant
  representations with incremental slowness learning.
\newblock {\em Frontiers in Neurorobotics}, 7(9).

\bibitem[Lusci et~al., 2013]{lusci2013}
Lusci, A., Pollastri, G., and Baldi, P. (2013).
\newblock Deep architectures and deep learning in chemoinformatics: the
  prediction of aqueous solubility for drug-like molecules.
\newblock {\em Journal of Chemical Information and Modeling}, 53(7):1563--1575.

\bibitem[Maas et~al., 2013]{Maas2013}
Maas, A.~L., Hannun, A.~Y., and Ng, A.~Y. (2013).
\newblock Rectifier nonlinearities improve neural network acoustic models.
\newblock In {\em International Conference on Machine Learning (ICML)}.

\bibitem[Maass, 1996]{maass1996}
Maass, W. (1996).
\newblock Lower bounds for the computational power of networks of spiking
  neurons.
\newblock {\em Neural Computation}, 8(1):1--40.

\bibitem[Maass, 1997]{maass1997}
Maass, W. (1997).
\newblock Networks of spiking neurons: the third generation of neural network
  models.
\newblock {\em Neural Networks}, 10(9):1659--1671.

\bibitem[Maass, 2000]{Maass2000}
Maass, W. (2000).
\newblock On the computational power of winner-take-all.
\newblock {\em Neural Computation}, 12:2519--2535.

\bibitem[Maass et~al., 2002]{maass2002}
Maass, W., Natschl{\"a}ger, T., and Markram, H. (2002).
\newblock Real-time computing without stable states: A new framework for neural
  computation based on perturbations.
\newblock {\em Neural Computation}, 14(11):2531--2560.

\bibitem[MacKay, 1992]{MacKay:92b}
MacKay, D. J.~C. (1992).
\newblock A practical {Bayesian} framework for backprop networks.
\newblock {\em Neural Computation}, 4:448--472.

\bibitem[MacKay and Miller, 1990]{MacKay:90}
MacKay, D. J.~C. and Miller, K.~D. (1990).
\newblock Analysis of {Linsker's} simulation of {Hebbian} rules.
\newblock {\em Neural Computation}, 2:173--187.

\bibitem[Maclin and Shavlik, 1993]{maclin1993}
Maclin, R. and Shavlik, J.~W. (1993).
\newblock Using knowledge-based neural networks to improve algorithms: Refining
  the {Chou-Fasman} algorithm for protein folding.
\newblock {\em Machine Learning}, 11(2-3):195--215.

\bibitem[Maclin and Shavlik, 1995]{maclin1995}
Maclin, R. and Shavlik, J.~W. (1995).
\newblock Combining the predictions of multiple classifiers: Using competitive
  learning to initialize neural networks.
\newblock In {\em Proc. IJCAI}, pages 524--531.

\bibitem[Madala and Ivakhnenko, 1994]{madala1994}
Madala, H.~R. and Ivakhnenko, A.~G. (1994).
\newblock {\em Inductive learning algorithms for complex systems modeling}.
\newblock CRC Press, Boca Raton.

\bibitem[Madani et~al., 2003]{madani2003}
Madani, O., Hanks, S., and Condon, A. (2003).
\newblock On the undecidability of probabilistic planning and related
  stochastic optimization problems.
\newblock {\em Artificial Intelligence}, 147(1):5--34.

\bibitem[Maei and Sutton, 2010]{10-GqLambda}
Maei, H.~R. and Sutton, R.~S. (2010).
\newblock {GQ($\lambda$)}: A general gradient algorithm for temporal-difference
  prediction learning with eligibility traces.
\newblock In {\em Proceedings of the Third Conference on Artificial General
  Intelligence}, volume~1, pages 91--96.

\bibitem[Maex and Orban, 1996]{maex1996}
Maex, R. and Orban, G. (1996).
\newblock Model circuit of spiking neurons generating directional selectivity
  in simple cells.
\newblock {\em Journal of Neurophysiology}, 75(4):1515--1545.

\bibitem[Mahadevan, 1996]{Mahadevan:96}
Mahadevan, S. (1996).
\newblock Average reward reinforcement learning: Foundations, algorithms, and
  empirical results.
\newblock {\em Machine Learning}, 22:159.

\bibitem[Malik and Perona, 1990]{malik1990}
Malik, J. and Perona, P. (1990).
\newblock Preattentive texture discrimination with early vision mechanisms.
\newblock {\em Journal of the Optical Society of America A}, 7(5):923--932.

\bibitem[Maniezzo, 1994]{maniezzo1994}
Maniezzo, V. (1994).
\newblock Genetic evolution of the topology and weight distribution of neural
  networks.
\newblock {\em IEEE Transactions on Neural Networks}, 5(1):39--53.

\bibitem[Manolios and Fanelli, 1994]{Manolios:94}
Manolios, P. and Fanelli, R. (1994).
\newblock First-order recurrent neural networks and deterministic finite state
  automata.
\newblock {\em Neural Computation}, 6:1155--1173.

\bibitem[Marchi et~al., 2014]{marchi2014}
Marchi, E., Ferroni, G., Eyben, F., Gabrielli, L., Squartini, S., and Schuller,
  B. (2014).
\newblock Multi-resolution linear prediction based features for audio onset
  detection with bidirectional {LSTM} neural networks.
\newblock In {\em Proc. 39th IEEE International Conference on Acoustics,
  Speech, and Signal Processing, ICASSP 2014, Florence, Italy}, pages
  2183--2187.

\bibitem[Markram, 2012]{markram2012}
Markram, H. (2012).
\newblock The human brain project.
\newblock {\em Scientific American}, 306(6):50--55.

\bibitem[Marquardt, 1963]{marquardt1963}
Marquardt, D.~W. (1963).
\newblock An algorithm for least-squares estimation of nonlinear parameters.
\newblock {\em Journal of the Society for Industrial \& Applied Mathematics},
  11(2):431--441.

\bibitem[Martens, 2010]{icml2010_094}
Martens, J. (2010).
\newblock Deep learning via {Hessian}-free optimization.
\newblock In F{\"u}rnkranz, J. and Joachims, T., editors, {\em Proceedings of
  the 27th International Conference on Machine Learning (ICML-10)}, pages
  735--742, Haifa, Israel. Omnipress.

\bibitem[Martens and Sutskever, 2011]{Martens:2011hessfree}
Martens, J. and Sutskever, I. (2011).
\newblock Learning recurrent neural networks with {Hessian}-free optimization.
\newblock In {\em Proceedings of the 28th International Conference on Machine
  Learning (ICML)}, pages 1033--1040.

\bibitem[Martinetz et~al., 1990]{Ritter:90}
Martinetz, T.~M., Ritter, H.~J., and Schulten, K.~J. (1990).
\newblock Three-dimensional neural net for learning visuomotor coordination of
  a robot arm.
\newblock {\em IEEE Transactions on Neural Networks}, 1(1):131--136.

\bibitem[Masci et~al., 2013]{masci:2013icip}
Masci, J., Giusti, A., Ciresan, D.~C., Fricout, G., and Schmidhuber, J. (2013).
\newblock A fast learning algorithm for image segmentation with max-pooling
  convolutional networks.
\newblock In {\em International Conference on Image Processing (ICIP13)}, pages
  2713--2717.

\bibitem[Matsuoka, 1992]{Matsuoka:92}
Matsuoka, K. (1992).
\newblock Noise injection into inputs in back-propagation learning.
\newblock {\em IEEE Transactions on Systems, Man, and Cybernetics},
  22(3):436--440.

\bibitem[Mayer et~al., 2008]{mayer2008}
Mayer, H., Gomez, F., Wierstra, D., Nagy, I., Knoll, A., and Schmidhuber, J.
  (2008).
\newblock A system for robotic heart surgery that learns to tie knots using
  recurrent neural networks.
\newblock {\em Advanced Robotics}, 22(13-14):1521--1537.

\bibitem[McCallum, 1996]{McCallum:96}
McCallum, R.~A. (1996).
\newblock Learning to use selective attention and short-term memory in
  sequential tasks.
\newblock In Maes, P., Mataric, M., Meyer, J.-A., Pollack, J., and Wilson,
  S.~W., editors, {\em From Animals to Animats 4: Proceedings of the Fourth
  International Conference on Simulation of Adaptive Behavior, Cambridge, MA},
  pages 315--324. MIT Press, Bradford Books.

\bibitem[McCulloch and Pitts, 1943]{mcculloch:43}
McCulloch, W. and Pitts, W. (1943).
\newblock A logical calculus of the ideas immanent in nervous activity.
\newblock {\em Bulletin of Mathematical Biophysics}, 7:115--133.

\bibitem[Melnik et~al., 2000]{Melnik2000}
Melnik, O., Levy, S.~D., and Pollack, J.~B. (2000).
\newblock {RAAM} for infinite context-free languages.
\newblock In {\em Proc. IJCNN (5)}, pages 585--590.

\bibitem[Memisevic and Hinton, 2010]{memisevic2010}
Memisevic, R. and Hinton, G.~E. (2010).
\newblock Learning to represent spatial transformations with factored
  higher-order {Boltzmann} machines.
\newblock {\em Neural Computation}, 22(6):1473--1492.

\bibitem[Menache et~al., 2002]{menache2002}
Menache, I., Mannor, S., and Shimkin, N. (2002).
\newblock Q-cut -- dynamic discovery of sub-goals in reinforcement learning.
\newblock In {\em Proc. ECML'02}, pages 295--306.

\bibitem[Merolla et~al., 2014]{merolla2014}
Merolla, P.~A., Arthur, J.~V., Alvarez-Icaza, R., Cassidy, A.~S., Sawada, J.,
  Akopyan, F., Jackson, B.~L., Imam, N., Guo, C., Nakamura, Y., Brezzo, B., Vo,
  I., Esser, S.~K., Appuswamy, R., Taba, B., Amir, A., Flickner, M.~D., Risk,
  W.~P., Manohar, R., and Modha, D.~S. (2014).
\newblock A million spiking-neuron integrated circuit with a scalable
  communication network and interface.
\newblock {\em Science}, 345(6197):668--673.

\bibitem[Mesnil et~al., 2011]{transfer2011}
Mesnil, G., Dauphin, Y., Glorot, X., Rifai, S., Bengio, Y., Goodfellow, I.,
  Lavoie, E., Muller, X., Desjardins, G., Warde-Farley, D., Vincent, P.,
  Courville, A., and Bergstra, J. (2011).
\newblock Unsupervised and transfer learning challenge: a deep learning
  approach.
\newblock In {\em JMLR W\&CP: Proc. Unsupervised and Transfer Learning},
  volume~7.

\bibitem[Meuleau et~al., 1999]{meuleau:icuai99}
Meuleau, N., Peshkin, L., Kim, K.~E., and Kaelbling, L.~P. (1999).
\newblock Learning finite state controllers for partially observable
  environments.
\newblock In {\em 15th International Conference of Uncertainty in AI}, pages
  427--436.

\bibitem[Miglino et~al., 1995]{miglino95evolving}
Miglino, O., Lund, H., and Nolfi, S. (1995).
\newblock Evolving mobile robots in simulated and real environments.
\newblock {\em Artificial Life}, 2(4):417--434.

\bibitem[Miller et~al., 1989]{miller:icga89}
Miller, G., Todd, P., and Hedge, S. (1989).
\newblock Designing neural networks using genetic algorithms.
\newblock In {\em Proceedings of the 3rd International Conference on Genetic
  Algorithms}, pages 379--384. Morgan Kauffman.

\bibitem[Miller and Harding, 2009]{miller2009}
Miller, J.~F. and Harding, S.~L. (2009).
\newblock Cartesian genetic programming.
\newblock In {\em Proceedings of the 11th Annual Conference Companion on
  Genetic and Evolutionary Computation Conference: Late Breaking Papers}, pages
  3489--3512. ACM.

\bibitem[Miller and Thomson, 2000]{miller2000}
Miller, J.~F. and Thomson, P. (2000).
\newblock Cartesian genetic programming.
\newblock In {\em Genetic Programming}, pages 121--132. Springer.

\bibitem[Miller, 1994]{Miller:94}
Miller, K.~D. (1994).
\newblock A model for the development of simple cell receptive fields and the
  ordered arrangement of orientation columns through activity-dependent
  competition between on- and off-center inputs.
\newblock {\em Journal of Neuroscience}, 14(1):409--441.

\bibitem[Miller et~al., 1995]{miller1995}
Miller, W.~T., Werbos, P.~J., and Sutton, R.~S. (1995).
\newblock {\em Neural networks for control}.
\newblock MIT Press.

\bibitem[Minai and Williams, 1994]{Minai:94}
Minai, A.~A. and Williams, R.~D. (1994).
\newblock Perturbation response in feedforward networks.
\newblock {\em Neural Networks}, 7(5):783--796.

\bibitem[Minsky, 1963]{Minsky:63}
Minsky, M. (1963).
\newblock Steps toward artificial intelligence.
\newblock In Feigenbaum, E. and Feldman, J., editors, {\em Computers and
  Thought}, pages 406--450. McGraw-Hill, New York.

\bibitem[Minsky and Papert, 1969]{MinskyPapert:69}
Minsky, M. and Papert, S. (1969).
\newblock {\em Perceptrons}.
\newblock Cambridge, MA: MIT Press.

\bibitem[Minton et~al., 1989]{minton1989}
Minton, S., Carbonell, J.~G., Knoblock, C.~A., Kuokka, D.~R., Etzioni, O., and
  Gil, Y. (1989).
\newblock Explanation-based learning: A problem solving perspective.
\newblock {\em Artificial Intelligence}, 40(1):63--118.

\bibitem[Mitchell, 1997]{Mitchell:97}
Mitchell, T. (1997).
\newblock {\em Machine Learning}.
\newblock McGraw Hill.

\bibitem[Mitchell et~al., 1986]{mitchell1986}
Mitchell, T.~M., Keller, R.~M., and Kedar-Cabelli, S.~T. (1986).
\newblock Explanation-based generalization: A unifying view.
\newblock {\em Machine Learning}, 1(1):47--80.

\bibitem[Mnih et~al., 2013]{atari2013}
Mnih, V., Kavukcuoglu, K., Silver, D., Graves, A., Antonoglou, I., Wierstra,
  D., and Riedmiller, M. (Dec 2013).
\newblock Playing {Atari} with deep reinforcement learning.
\newblock Technical Report arXiv:1312.5602 [cs.LG], Deepmind Technologies.

\bibitem[Mohamed and Hinton, 2010]{mohamed2010}
Mohamed, A. and Hinton, G.~E. (2010).
\newblock Phone recognition using restricted {Boltzmann} machines.
\newblock In {\em IEEE International Conference on Acoustics, Speech and Signal
  Processing (ICASSP)}, pages 4354--4357.

\bibitem[Molgedey and Schuster, 1994]{Molgedey:94}
Molgedey, L. and Schuster, H.~G. (1994).
\newblock Separation of independent signals using time-delayed correlations.
\newblock {\em Phys. Reviews Letters}, 72(23):3634--3637.

\bibitem[M$\o$ller, 1993]{Moller:93}
M$\o$ller, M.~F. (1993).
\newblock Exact calculation of the product of the {H}essian matrix of
  feed-forward network error functions and a vector in {O}({N}) time.
\newblock Technical Report PB-432, Computer Science Department, Aarhus
  University, Denmark.

\bibitem[Montana and Davis, 1989]{montana1989}
Montana, D.~J. and Davis, L. (1989).
\newblock Training feedforward neural networks using genetic algorithms.
\newblock In {\em Proceedings of the 11th International Joint Conference on
  Artificial Intelligence (IJCAI) - Volume 1}, IJCAI'89, pages 762--767, San
  Francisco, CA, USA. Morgan Kaufmann Publishers Inc.

\bibitem[Montavon et~al., 2012]{tricksofthetrade:2012}
Montavon, G., Orr, G., and M{\"u}ller, K. (2012).
\newblock {\em Neural Networks: Tricks of the Trade}.
\newblock Number LNCS 7700 in Lecture Notes in Computer Science Series.
  Springer Verlag.

\bibitem[Moody, 1989]{Moody:89}
Moody, J.~E. (1989).
\newblock Fast learning in multi-resolution hierarchies.
\newblock In Touretzky, D.~S., editor, {\em Advances in Neural Information
  Processing Systems (NIPS) 1}, pages 29--39. Morgan Kaufmann.

\bibitem[Moody, 1992]{Moody:92}
Moody, J.~E. (1992).
\newblock The effective number of parameters: An analysis of generalization and
  regularization in nonlinear learning systems.
\newblock In Lippman, D.~S., Moody, J.~E., and Touretzky, D.~S., editors, {\em
  Advances in Neural Information Processing Systems (NIPS) 4}, pages 847--854.
  Morgan Kaufmann.

\bibitem[Moody and Utans, 1994]{Moody:94a}
Moody, J.~E. and Utans, J. (1994).
\newblock Architecture selection strategies for neural networks: Application to
  corporate bond rating prediction.
\newblock In Refenes, A.~N., editor, {\em Neural Networks in the Capital
  Markets}. John Wiley \& Sons.

\bibitem[Moore and Atkeson, 1995]{partigame}
Moore, A. and Atkeson, C. (1995).
\newblock The parti-game algorithm for variable resolution reinforcement
  learning in multidimensional state-spaces.
\newblock {\em Machine Learning}, 21(3):199--233.

\bibitem[Moore and Atkeson, 1993]{Moore:93}
Moore, A. and Atkeson, C.~G. (1993).
\newblock Prioritized sweeping: Reinforcement learning with less data and less
  time.
\newblock {\em Machine Learning}, 13:103--130.

\bibitem[Moriarty, 1997]{moriarty:phd}
Moriarty, D.~E. (1997).
\newblock {\em Symbiotic Evolution of Neural Networks in Sequential Decision
  Tasks}.
\newblock PhD thesis, Department of Computer Sciences, The University of Texas
  at Austin.

\bibitem[Moriarty and Miikkulainen, 1996]{moriarty:ml96}
Moriarty, D.~E. and Miikkulainen, R. (1996).
\newblock Efficient reinforcement learning through symbiotic evolution.
\newblock {\em Machine Learning}, 22:11--32.

\bibitem[Morimoto and Doya, 2000]{Doya:00}
Morimoto, J. and Doya, K. (2000).
\newblock Robust reinforcement learning.
\newblock In Leen, T.~K., Dietterich, T.~G., and Tresp, V., editors, {\em
  Advances in Neural Information Processing Systems (NIPS) 13}, pages
  1061--1067. MIT Press.

\bibitem[Mosteller and Tukey, 1968]{Mosteller:68}
Mosteller, F. and Tukey, J.~W. (1968).
\newblock Data analysis, including statistics.
\newblock In Lindzey, G. and Aronson, E., editors, {\em Handbook of Social
  Psychology, Vol. 2}. Addison-Wesley.

\bibitem[Mozer, 1989]{Mozer:89focus}
Mozer, M.~C. (1989).
\newblock A focused back-propagation algorithm for temporal sequence
  recognition.
\newblock {\em Complex Systems}, 3:349--381.

\bibitem[Mozer, 1991]{Mozer:91nips}
Mozer, M.~C. (1991).
\newblock Discovering discrete distributed representations with iterative
  competitive learning.
\newblock In Lippmann, R.~P., Moody, J.~E., and Touretzky, D.~S., editors, {\em
  Advances in Neural Information Processing Systems 3}, pages 627--634. Morgan
  Kaufmann.

\bibitem[Mozer, 1992]{Mozer:92nips}
Mozer, M.~C. (1992).
\newblock Induction of multiscale temporal structure.
\newblock In Lippman, D.~S., Moody, J.~E., and Touretzky, D.~S., editors, {\em
  Advances in Neural Information Processing Systems (NIPS) 4}, pages 275--282.
  Morgan Kaufmann.

\bibitem[Mozer and Smolensky, 1989]{Mozer:89a}
Mozer, M.~C. and Smolensky, P. (1989).
\newblock Skeletonization: A technique for trimming the fat from a network via
  relevance assessment.
\newblock In Touretzky, D.~S., editor, {\em Advances in Neural Information
  Processing Systems (NIPS) 1}, pages 107--115. Morgan Kaufmann.

\bibitem[Muller et~al., 1995]{muller1995}
Muller, U.~A., Gunzinger, A., and Guggenb{\"u}hl, W. (1995).
\newblock Fast neural net simulation with a {DSP} processor array.
\newblock {\em IEEE Transactions on Neural Networks}, 6(1):203--213.

\bibitem[Munro, 1987]{Munro:87}
Munro, P.~W. (1987).
\newblock A dual back-propagation scheme for scalar reinforcement learning.
\newblock {\em Proceedings of the Ninth Annual Conference of the Cognitive
  Science Society, Seattle, WA}, pages 165--176.

\bibitem[Murray and Edwards, 1993]{Murray:93}
Murray, A.~F. and Edwards, P.~J. (1993).
\newblock Synaptic weight noise during {MLP} learning enhances fault-tolerance,
  generalisation and learning trajectory.
\newblock In S.~J.~Hanson, J. D.~C. and Giles, C.~L., editors, {\em Advances in
  Neural Information Processing Systems (NIPS) 5}, pages 491--498. San Mateo,
  CA: Morgan Kaufmann.

\bibitem[Nadal and Parga, 1994]{Nadal:94}
Nadal, J.-P. and Parga, N. (1994).
\newblock Non-linear neurons in the low noise limit: a factorial code maximises
  information transfer.
\newblock {\em Network}, 5:565--581.

\bibitem[Nagumo et~al., 1962]{nagumo1962}
Nagumo, J., Arimoto, S., and Yoshizawa, S. (1962).
\newblock An active pulse transmission line simulating nerve axon.
\newblock {\em Proceedings of the IRE}, 50(10):2061--2070.

\bibitem[Nair and Hinton, 2010]{Nair2010}
Nair, V. and Hinton, G.~E. (2010).
\newblock {Rectified linear units improve restricted Boltzmann machines}.
\newblock In {\em International Conference on Machine Learning (ICML)}.

\bibitem[Narendra and Parthasarathy, 1990]{narendra1990}
Narendra, K.~S. and Parthasarathy, K. (1990).
\newblock Identification and control of dynamical systems using neural
  networks.
\newblock {\em Neural Networks, IEEE Transactions on}, 1(1):4--27.

\bibitem[Narendra and Thathatchar, 1974]{Narendra:74}
Narendra, K.~S. and Thathatchar, M. A.~L. (1974).
\newblock Learning automata -- a survey.
\newblock {\em IEEE Transactions on Systems, Man, and Cybernetics}, 4:323--334.

\bibitem[Neal, 1995]{neal1995}
Neal, R.~M. (1995).
\newblock {\em Bayesian learning for neural networks}.
\newblock PhD thesis, University of Toronto.

\bibitem[Neal, 2006]{neal2006b}
Neal, R.~M. (2006).
\newblock Classification with {Bayesian} neural networks.
\newblock In Quinonero-Candela, J., Magnini, B., Dagan, I., and D'Alche-Buc,
  F., editors, {\em Machine Learning Challenges. Evaluating Predictive
  Uncertainty, Visual Object Classification, and Recognising Textual
  Entailment}, volume 3944 of {\em Lecture Notes in Computer Science}, pages
  28--32. Springer.

\bibitem[Neal and Zhang, 2006]{neal2006}
Neal, R.~M. and Zhang, J. (2006).
\newblock High dimensional classification with {Bayesian} neural networks and
  {Dirichlet} diffusion trees.
\newblock In Guyon, I., Gunn, S., Nikravesh, M., and Zadeh, L.~A., editors,
  {\em Feature Extraction: Foundations and Applications, Studies in Fuzziness
  and Soft Computing}, pages 265--295. Springer.

\bibitem[Neftci et~al., 2014]{nefti2014}
Neftci, E., Das, S., Pedroni, B., Kreutz-Delgado, K., and Cauwenberghs, G.
  (2014).
\newblock Event-driven contrastive divergence for spiking neuromorphic systems.
\newblock {\em Frontiers in Neuroscience}, 7(272).

\bibitem[Neil and Liu, 2014]{neil2014}
Neil, D. and Liu, S.-C. (2014).
\newblock Minitaur, an event-driven {FPGA}-based spiking network accelerator.
\newblock {\em IEEE Transactions on Very Large Scale Integration (VLSI)
  Systems}, PP(99):1--8.

\bibitem[Nessler et~al., 2013]{nessler2013}
Nessler, B., Pfeiffer, M., Buesing, L., and Maass, W. (2013).
\newblock Bayesian computation emerges in generic cortical microcircuits
  through spike-timing-dependent plasticity.
\newblock {\em PLoS Computational Biology}, 9(4):e1003037.

\bibitem[Neti et~al., 1992]{Neti:92}
Neti, C., Schneider, M.~H., and Young, E.~D. (1992).
\newblock Maximally fault tolerant neural networks.
\newblock In {\em IEEE Transactions on Neural Networks}, volume~3, pages
  14--23.

\bibitem[Neuneier and Zimmermann, 1996]{DBLP:conf/nips/NeuneierZ96}
Neuneier, R. and Zimmermann, H.-G. (1996).
\newblock How to train neural networks.
\newblock In Orr, G.~B. and M{\"u}ller, K.-R., editors, {\em Neural Networks:
  Tricks of the Trade}, volume 1524 of {\em Lecture Notes in Computer Science},
  pages 373--423. Springer.

\bibitem[Newton, 1687]{newton1687}
Newton, I. (1687).
\newblock {\em Philosophiae naturalis principia mathematica}.
\newblock William Dawson \& Sons Ltd., London.

\bibitem[Nguyen and Widrow, 1989]{NguyenWidrow:89}
Nguyen, N. and Widrow, B. (1989).
\newblock The truck backer-upper: An example of self learning in neural
  networks.
\newblock In {\em Proceedings of the International Joint Conference on Neural
  Networks}, pages 357--363. IEEE Press.

\bibitem[Nilsson, 1980]{Nilsson:80}
Nilsson, N.~J. (1980).
\newblock {\em Principles of artificial intelligence}.
\newblock Morgan Kaufmann, San Francisco, CA, USA.

\bibitem[Nolfi et~al., 1994a]{nolfi:alife4}
Nolfi, S., Floreano, D., Miglino, O., and Mondada, F. (1994a).
\newblock How to evolve autonomous robots: {D}ifferent approaches in
  evolutionary robotics.
\newblock In Brooks, R.~A. and Maes, P., editors, {\em Fourth International
  Workshop on the Synthesis and Simulation of Living Systems (Artificial Life
  {IV})}, pages 190--197. MIT.

\bibitem[Nolfi et~al., 1994b]{nolfi1994}
Nolfi, S., Parisi, D., and Elman, J.~L. (1994b).
\newblock Learning and evolution in neural networks.
\newblock {\em Adaptive Behavior}, 3(1):5--28.

\bibitem[Nowak et~al., 2006]{nowak2006}
Nowak, E., Jurie, F., and Triggs, B. (2006).
\newblock Sampling strategies for bag-of-features image classification.
\newblock In {\em Proc. ECCV 2006}, pages 490--503. Springer.

\bibitem[Nowlan and Hinton, 1992]{Nowlan:92}
Nowlan, S.~J. and Hinton, G.~E. (1992).
\newblock Simplifying neural networks by soft weight sharing.
\newblock {\em Neural Computation}, 4:173--193.

\bibitem[O'Connor et~al., 2013]{oconnor2013}
O'Connor, P., Neil, D., Liu, S.-C., Delbruck, T., and Pfeiffer, M. (2013).
\newblock Real-time classification and sensor fusion with a spiking deep belief
  network.
\newblock {\em Frontiers in Neuroscience}, 7(178).

\bibitem[Oh and Jung, 2004]{gpu2004}
Oh, K.-S. and Jung, K. (2004).
\newblock {GPU} implementation of neural networks.
\newblock {\em Pattern Recognition}, 37(6):1311--1314.

\bibitem[Oja, 1989]{Oja:89}
Oja, E. (1989).
\newblock Neural networks, principal components, and subspaces.
\newblock {\em International Journal of Neural Systems}, 1(1):61--68.

\bibitem[Oja, 1991]{Oja:91}
Oja, E. (1991).
\newblock Data compression, feature extraction, and autoassociation in
  feedforward neural networks.
\newblock In Kohonen, T., M\"{a}kisara, K., Simula, O., and Kangas, J.,
  editors, {\em Artificial Neural Networks}, volume~1, pages 737--745. Elsevier
  Science Publishers B.V., North-Holland.

\bibitem[Olshausen and Field, 1996]{Olshausen:96}
Olshausen, B.~A. and Field, D.~J. (1996).
\newblock Emergence of simple-cell receptive field properties by learning a
  sparse code for natural images.
\newblock {\em Nature}, 381(6583):607--609.

\bibitem[Omlin and Giles, 1996]{Omlin:96}
Omlin, C. and Giles, C.~L. (1996).
\newblock Extraction of rules from discrete-time recurrent neural networks.
\newblock {\em Neural Networks}, 9(1):41--52.

\bibitem[Oquab et~al., 2013]{oquab2013}
Oquab, M., Bottou, L., Laptev, I., and Sivic, J. (2013).
\newblock Learning and transferring mid-level image representations using
  convolutional neural networks.
\newblock Technical Report hal-00911179.

\bibitem[O'Reilly, 2003]{oreilly:2003}
O'Reilly, R. (2003).
\newblock Making working memory work: A computational model of learning in the
  prefrontal cortex and basal ganglia.
\newblock Technical Report ICS-03-03, ICS.

\bibitem[O'Reilly, 1996]{oreilly1996}
O'Reilly, R.~C. (1996).
\newblock Biologically plausible error-driven learning using local activation
  differences: The generalized recirculation algorithm.
\newblock {\em Neural Computation}, 8(5):895--938.

\bibitem[Orr and M{\"u}ller, 1998]{orr1998neural}
Orr, G. and M{\"u}ller, K. (1998).
\newblock {\em Neural Networks: Tricks of the Trade}.
\newblock Number LNCS 1524 in Lecture Notes in Computer Science Series.
  Springer Verlag.

\bibitem[Ostrovskii et~al., 1971]{ostrovskii:1971}
Ostrovskii, G.~M., Volin, Y.~M., and Borisov, W.~W. (1971).
\newblock {{\"U}ber die Berechnung von Ableitungen}.
\newblock {\em Wiss. Z. Tech. Hochschule f{\"u}r Chemie}, 13:382--384.

\bibitem[Otsuka, 2010]{otsuka2010phd}
Otsuka, M. (2010).
\newblock {\em Goal-Oriented Representation of the External World: A
  Free-Energy-Based Approach}.
\newblock PhD thesis, Nara Institute of Science and Technology.

\bibitem[Otsuka et~al., 2010]{otsuka2010}
Otsuka, M., Yoshimoto, J., and Doya, K. (2010).
\newblock Free-energy-based reinforcement learning in a partially observable
  environment.
\newblock In {\em Proc. ESANN}.

\bibitem[Otte et~al., 2012]{otte2012local}
Otte, S., Krechel, D., Liwicki, M., and Dengel, A. (2012).
\newblock Local feature based online mode detection with recurrent neural
  networks.
\newblock In {\em Proceedings of the 2012 International Conference on Frontiers
  in Handwriting Recognition}, pages 533--537. IEEE Computer Society.

\bibitem[Oudeyer et~al., 2013]{Oudeyer:12intrinsic}
Oudeyer, P.-Y., Baranes, A., and Kaplan, F. (2013).
\newblock Intrinsically motivated learning of real world sensorimotor skills
  with developmental constraints.
\newblock In Baldassarre, G. and Mirolli, M., editors, {\em Intrinsically
  Motivated Learning in Natural and Artificial Systems}. Springer.

\bibitem[O’Reilly et~al., 2013]{oreilly2013}
O’Reilly, R.~C., Wyatte, D., Herd, S., Mingus, B., and Jilk, D.~J. (2013).
\newblock Recurrent processing during object recognition.
\newblock {\em Frontiers in Psychology}, 4:124.

\bibitem[Pachitariu and Sahani, 2013]{pachitariu2013regularization}
Pachitariu, M. and Sahani, M. (2013).
\newblock Regularization and nonlinearities for neural language models: when
  are they needed?
\newblock {\em arXiv preprint arXiv:1301.5650}.

\bibitem[Palm, 1980]{Palm:80}
Palm, G. (1980).
\newblock On associative memory.
\newblock {\em Biological Cybernetics}, 36.

\bibitem[Palm, 1992]{Palm:92}
Palm, G. (1992).
\newblock On the information storage capacity of local learning rules.
\newblock {\em Neural Computation}, 4(2):703--711.

\bibitem[Pan and Yang, 2010]{transfer2010}
Pan, S.~J. and Yang, Q. (2010).
\newblock A survey on transfer learning.
\newblock {\em IEEE Transactions on Knowledge and Data Engineering},
  22(10):1345--1359.

\bibitem[Parekh et~al., 2000]{parekh2000}
Parekh, R., Yang, J., and Honavar, V. (2000).
\newblock Constructive neural network learning algorithms for multi-category
  pattern classification.
\newblock {\em IEEE Transactions on Neural Networks}, 11(2):436--451.

\bibitem[Parker, 1985]{Parker:85}
Parker, D.~B. (1985).
\newblock Learning-logic.
\newblock Technical Report TR-47, Center for Comp. Research in Economics and
  Management Sci., MIT.

\bibitem[Pascanu et~al., 2013a]{pascanu2013construct}
Pascanu, R., Gulcehre, C., Cho, K., and Bengio, Y. (2013a).
\newblock How to construct deep recurrent neural networks.
\newblock {\em arXiv preprint arXiv:1312.6026}.

\bibitem[Pascanu et~al., 2013b]{pascanu2013}
Pascanu, R., Mikolov, T., and Bengio, Y. (2013b).
\newblock On the difficulty of training recurrent neural networks.
\newblock In {\em ICML'13: JMLR: W\&CP volume 28}.

\bibitem[Pasemann et~al., 1999]{pasemann99}
Pasemann, F., Steinmetz, U., and Dieckman, U. (1999).
\newblock Evolving structure and function of neurocontrollers.
\newblock In Angeline, P.~J., Michalewicz, Z., Schoenauer, M., Yao, X., and
  Zalzala, A., editors, {\em Proceedings of the Congress on Evolutionary
  Computation}, volume~3, pages 1973--1978, Mayflower Hotel, Washington D.C.,
  USA. IEEE Press.

\bibitem[Pearlmutter, 1989]{Pearlmutter:89}
Pearlmutter, B.~A. (1989).
\newblock Learning state space trajectories in recurrent neural networks.
\newblock {\em Neural Computation}, 1(2):263--269.

\bibitem[Pearlmutter, 1994]{Pearlmutter:93}
Pearlmutter, B.~A. (1994).
\newblock Fast exact multiplication by the {Hessian}.
\newblock {\em Neural Computation}, 6(1):147--160.

\bibitem[Pearlmutter, 1995]{Pearlmutter:95}
Pearlmutter, B.~A. (1995).
\newblock Gradient calculations for dynamic recurrent neural networks: A
  survey.
\newblock {\em IEEE Transactions on Neural Networks}, 6(5):1212--1228.

\bibitem[Pearlmutter and Hinton, 1986]{PearlmutterHinton:86}
Pearlmutter, B.~A. and Hinton, G.~E. (1986).
\newblock G-maximization: An unsupervised learning procedure for discovering
  regularities.
\newblock In Denker, J.~S., editor, {\em Neural Networks for Computing:
  American Institute of Physics Conference Proceedings 151}, volume~2, pages
  333--338.

\bibitem[Peng and Williams, 1996]{Peng:96}
Peng, J. and Williams, R.~J. (1996).
\newblock Incremental multi-step {Q}-learning.
\newblock {\em Machine Learning}, 22:283--290.

\bibitem[P\'{e}rez-Ortiz et~al., 2003]{Perez:02}
P\'{e}rez-Ortiz, J.~A., Gers, F.~A., Eck, D., and Schmidhuber, J. (2003).
\newblock Kalman filters improve {LSTM} network performance in problems
  unsolvable by traditional recurrent nets.
\newblock {\em Neural Networks}, (16):241--250.

\bibitem[Perrett et~al., 1992]{perrett1992}
Perrett, D., Hietanen, J., Oram, M., Benson, P., and Rolls, E. (1992).
\newblock Organization and functions of cells responsive to faces in the
  temporal cortex [and discussion].
\newblock {\em Philosophical Transactions of the Royal Society of London.
  Series B: Biological Sciences}, 335(1273):23--30.

\bibitem[Perrett et~al., 1982]{perrett1982}
Perrett, D., Rolls, E., and Caan, W. (1982).
\newblock Visual neurones responsive to faces in the monkey temporal cortex.
\newblock {\em Experimental Brain Research}, 47(3):329--342.

\bibitem[Peters, 2010]{peters2010}
Peters, J. (2010).
\newblock Policy gradient methods.
\newblock {\em Scholarpedia}, 5(11):3698.

\bibitem[Peters and Schaal, 2008a]{peters2008neurocomputing}
Peters, J. and Schaal, S. (2008a).
\newblock Natural actor-critic.
\newblock {\em Neurocomputing}, 71:1180--1190.

\bibitem[Peters and Schaal, 2008b]{peters2008neuralnetworks}
Peters, J. and Schaal, S. (2008b).
\newblock {Reinforcement learning of motor skills with policy gradients}.
\newblock {\em Neural Network}, 21(4):682--697.

\bibitem[Pham et~al., 2013]{Pham2013}
Pham, V., Kermorvant, C., and Louradour, J. (2013).
\newblock {Dropout Improves Recurrent Neural Networks for Handwriting
  Recognition}.
\newblock {\em arXiv preprint arXiv:1312.4569}.

\bibitem[Pineda, 1987]{Pineda:87}
Pineda, F.~J. (1987).
\newblock Generalization of back-propagation to recurrent neural networks.
\newblock {\em Physical Review Letters}, 19(59):2229--2232.

\bibitem[Plate, 1993]{Plate:93}
Plate, T.~A. (1993).
\newblock Holographic recurrent networks.
\newblock In S.~J.~Hanson, J. D.~C. and Giles, C.~L., editors, {\em Advances in
  Neural Information Processing Systems (NIPS) 5}, pages 34--41. Morgan
  Kaufmann.

\bibitem[Plumbley, 1991]{Plumbley:91}
Plumbley, M.~D. (1991).
\newblock {On information theory and unsupervised neural networks.
  Dissertation, published as technical report CUED/F-INFENG/TR.78, Engineering
  Department, Cambridge University}.

\bibitem[Pollack, 1988]{pollack1988implications}
Pollack, J.~B. (1988).
\newblock Implications of recursive distributed representations.
\newblock In {\em Proc. NIPS}, pages 527--536.

\bibitem[Pollack, 1990]{Pollack:90}
Pollack, J.~B. (1990).
\newblock Recursive distributed representation.
\newblock {\em Artificial Intelligence}, 46:77--105.

\bibitem[Pontryagin et~al., 1961]{PONTRYAGIN61A}
Pontryagin, L.~S., Boltyanskii, V.~G., Gamrelidze, R.~V., and Mishchenko, E.~F.
  (1961).
\newblock {\em The Mathematical Theory of Optimal Processes}.

\bibitem[Poon and Domingos, 2011]{domingos2011}
Poon, H. and Domingos, P. (2011).
\newblock Sum-product networks: A new deep architecture.
\newblock In {\em IEEE International Conference on Computer Vision (ICCV)
  Workshops}, pages 689--690. IEEE.

\bibitem[Post, 1936]{Post:36}
Post, E.~L. (1936).
\newblock Finite combinatory processes-formulation 1.
\newblock {\em The Journal of Symbolic Logic}, 1(3):103--105.

\bibitem[Prasoon et~al., 2013]{prasoon:13}
Prasoon, A., Petersen, K., Igel, C., Lauze, F., Dam, E., and Nielsen, M.
  (2013).
\newblock Voxel classification based on triplanar convolutional neural networks
  applied to cartilage segmentation in knee {MRI}.
\newblock In {\em Medical Image Computing and Computer Assisted Intervention
  (MICCAI)}, volume 8150 of {\em LNCS}, pages 246--253. Springer.

\bibitem[Precup et~al., 1998]{Precup:MTimeNIPS98}
Precup, D., Sutton, R.~S., and Singh, S. (1998).
\newblock Multi-time models for temporally abstract planning.
\newblock In {\em Advances in Neural Information Processing Systems (NIPS)},
  pages 1050--1056. Morgan Kaufmann.

\bibitem[Prokhorov, 2010]{prokhorov2010}
Prokhorov, D. (2010).
\newblock A convolutional learning system for object classification in {3-D
  LIDAR} data.
\newblock {\em IEEE Transactions on Neural Networks}, 21(5):858--863.

\bibitem[Prokhorov et~al., 2001]{prokhorov2001}
Prokhorov, D., Puskorius, G., and Feldkamp, L. (2001).
\newblock Dynamical neural networks for control.
\newblock In Kolen, J. and Kremer, S., editors, {\em A field guide to dynamical
  recurrent networks}, pages 23--78. IEEE Press.

\bibitem[Prokhorov and Wunsch, 1997]{prwu97}
Prokhorov, D. and Wunsch, D. (1997).
\newblock Adaptive critic design.
\newblock {\em IEEE Transactions on Neural Networks}, 8(5):997--1007.

\bibitem[Prokhorov et~al., 2002]{prokhorov2002meta}
Prokhorov, D.~V., Feldkamp, L.~A., and Tyukin, I.~Y. (2002).
\newblock Adaptive behavior with fixed weights in {RNN}: an overview.
\newblock In {\em Proceedings of the IEEE International Joint Conference on
  Neural Networks (IJCNN)}, pages 2018--2023.

\bibitem[Puskorius and Feldkamp, 1994]{Puskorius:94}
Puskorius, G.~V. and Feldkamp, L.~A. (1994).
\newblock Neurocontrol of nonlinear dynamical systems with {Kalman} filter
  trained recurrent networks.
\newblock {\em IEEE Transactions on Neural Networks}, 5(2):279--297.

\bibitem[Raiko et~al., 2012]{raiko2012}
Raiko, T., Valpola, H., and LeCun, Y. (2012).
\newblock Deep learning made easier by linear transformations in perceptrons.
\newblock In {\em International Conference on Artificial Intelligence and
  Statistics}, pages 924--932.

\bibitem[Raina et~al., 2009]{raina2009large}
Raina, R., Madhavan, A., and Ng, A. (2009).
\newblock Large-scale deep unsupervised learning using graphics processors.
\newblock In {\em Proceedings of the 26th Annual International Conference on
  Machine Learning (ICML)}, pages 873--880. ACM.

\bibitem[Ramacher et~al., 1993]{ramacher93}
Ramacher, U., Raab, W., Anlauf, J., Hachmann, U., Beichter, J., Bruels, N.,
  Wesseling, M., Sicheneder, E., Maenner, R., Glaess, J., and Wurz, A. (1993).
\newblock Multiprocessor and memory architecture of the neurocomputer
  {SYNAPSE-1}.
\newblock {\em International Journal of Neural Systems}, 4(4):333--336.

\bibitem[Ranzato et~al., 2006]{ranzato-06}
Ranzato, M., Poultney, C., Chopra, S., and LeCun, Y. (2006).
\newblock Efficient learning of sparse representations with an energy-based
  model.
\newblock In et~al., J.~P., editor, {\em Advances in Neural Information
  Processing Systems (NIPS 2006)}. MIT Press.

\bibitem[Ranzato et~al., 2007]{ranzato-cvpr-07}
Ranzato, M.~A., Huang, F., Boureau, Y., and LeCun, Y. (2007).
\newblock Unsupervised learning of invariant feature hierarchies with
  applications to object recognition.
\newblock In {\em Proc. Computer Vision and Pattern Recognition Conference
  (CVPR'07)}, pages 1--8. IEEE Press.

\bibitem[Rauber et~al., 2002]{rauber2002}
Rauber, A., Merkl, D., and Dittenbach, M. (2002).
\newblock The growing hierarchical self-organizing map: exploratory analysis of
  high-dimensional data.
\newblock {\em IEEE Transactions on Neural Networks}, 13(6):1331--1341.

\bibitem[Razavian et~al., 2014]{razavian2014}
Razavian, A.~S., Azizpour, H., Sullivan, J., and Carlsson, S. (2014).
\newblock {CNN} features off-the-shelf: an astounding baseline for recognition.
\newblock {\em arXiv preprint arXiv:1403.6382}.

\bibitem[Rechenberg, 1971]{Rechenberg:71}
Rechenberg, I. (1971).
\newblock {Evolutionsstrategie - Optimierung technischer Systeme nach
  Prinzipien der biologischen Evolution. Dissertation}.
\newblock Published 1973 by Fromman-Holzboog.

\bibitem[Redlich, 1993]{Redlich:93a}
Redlich, A.~N. (1993).
\newblock Redundancy reduction as a strategy for unsupervised learning.
\newblock {\em Neural Computation}, 5:289--304.

\bibitem[Refenes et~al., 1994]{Refenes:94}
Refenes, N.~A., Zapranis, A., and Francis, G. (1994).
\newblock Stock performance modeling using neural networks: a comparative study
  with regression models.
\newblock {\em Neural Networks}, 7(2):375--388.

\bibitem[Rezende and Gerstner, 2014]{rezende2014}
Rezende, D.~J. and Gerstner, W. (2014).
\newblock Stochastic variational learning in recurrent spiking networks.
\newblock {\em Frontiers in Computational Neuroscience}, 8:38.

\bibitem[Riedmiller, 2005]{nfq}
Riedmiller, M. (2005).
\newblock Neural fitted {Q} iteration---first experiences with a data efficient
  neural reinforcement learning method.
\newblock In {\em Proc. ECML-2005}, pages 317--328. Springer-Verlag Berlin
  Heidelberg.

\bibitem[Riedmiller and Braun, 1993]{rprop93}
Riedmiller, M. and Braun, H. (1993).
\newblock A direct adaptive method for faster backpropagation learning: The
  {Rprop} algorithm.
\newblock In {\em Proc. IJCNN}, pages 586--591. IEEE Press.

\bibitem[Riedmiller et~al., 2012]{rieijcnn12}
Riedmiller, M., Lange, S., and Voigtlaender, A. (2012).
\newblock Autonomous reinforcement learning on raw visual input data in a real
  world application.
\newblock In {\em International Joint Conference on Neural Networks (IJCNN)},
  pages 1--8, Brisbane, Australia.

\bibitem[Riesenhuber and Poggio, 1999]{riesenhuber:1999}
Riesenhuber, M. and Poggio, T. (1999).
\newblock Hierarchical models of object recognition in cortex.
\newblock {\em Nat. Neurosci.}, 2(11):1019--1025.

\bibitem[Rifai et~al., 2011]{vincent2011}
Rifai, S., Vincent, P., Muller, X., Glorot, X., and Bengio, Y. (2011).
\newblock Contractive auto-encoders: Explicit invariance during feature
  extraction.
\newblock In {\em Proceedings of the 28th International Conference on Machine
  Learning (ICML-11)}, pages 833--840.

\bibitem[Ring et~al., 2011]{ring:icdl2011}
Ring, M., Schaul, T., and Schmidhuber, J. (2011).
\newblock The two-dimensional organization of behavior.
\newblock In {\em Proceedings of the First Joint Conference on Development
  Learning and on Epigenetic Robotics ICDL-EPIROB}, Frankfurt.

\bibitem[Ring, 1991]{Ring:91}
Ring, M.~B. (1991).
\newblock Incremental development of complex behaviors through automatic
  construction of sensory-motor hierarchies.
\newblock In Birnbaum, L. and Collins, G., editors, {\em Machine Learning:
  Proceedings of the Eighth International Workshop}, pages 343--347. Morgan
  Kaufmann.

\bibitem[Ring, 1993]{Ring:93}
Ring, M.~B. (1993).
\newblock Learning sequential tasks by incrementally adding higher orders.
\newblock In S.~J.~Hanson, J. D.~C. and Giles, C.~L., editors, {\em Advances in
  Neural Information Processing Systems 5}, pages 115--122. Morgan Kaufmann.

\bibitem[Ring, 1994]{Ring:94}
Ring, M.~B. (1994).
\newblock {\em Continual Learning in Reinforcement Environments}.
\newblock PhD thesis, University of Texas at Austin, Austin, Texas 78712.

\bibitem[Risi and Stanley, 2012]{risi2012}
Risi, S. and Stanley, K.~O. (2012).
\newblock A unified approach to evolving plasticity and neural geometry.
\newblock In {\em International Joint Conference on Neural Networks (IJCNN)},
  pages 1--8. IEEE.

\bibitem[Rissanen, 1986]{Rissanen:86}
Rissanen, J. (1986).
\newblock Stochastic complexity and modeling.
\newblock {\em The Annals of Statistics}, 14(3):1080--1100.

\bibitem[Ritter and Kohonen, 1989]{ritter1989}
Ritter, H. and Kohonen, T. (1989).
\newblock Self-organizing semantic maps.
\newblock {\em Biological Cybernetics}, 61(4):241--254.

\bibitem[Robinson and Fallside, 1987]{RobinsonFallside:87tr}
Robinson, A.~J. and Fallside, F. (1987).
\newblock The utility driven dynamic error propagation network.
\newblock Technical Report CUED/F-INFENG/TR.1, Cambridge University Engineering
  Department.

\bibitem[Robinson and Fallside, 1989]{RobinsonFallside:89}
Robinson, T. and Fallside, F. (1989).
\newblock Dynamic reinforcement driven error propagation networks with
  application to game playing.
\newblock In {\em Proceedings of the 11th Conference of the Cognitive Science
  Society, Ann Arbor}, pages 836--843.

\bibitem[Rodriguez and Wiles, 1998]{Rodriguez+Wiles:1998:nips10}
Rodriguez, P. and Wiles, J. (1998).
\newblock Recurrent neural networks can learn to implement symbol-sensitive
  counting.
\newblock In {\em Advances in Neural Information Processing Systems (NIPS)},
  volume~10, pages 87--93. The {MIT} Press.

\bibitem[Rodriguez et~al., 1999]{Rodriguez:1999CS}
Rodriguez, P., Wiles, J., and Elman, J. (1999).
\newblock A recurrent neural network that learns to count.
\newblock {\em Connection Science}, 11(1):5--40.

\bibitem[Roggen et~al., 2003]{roggen2003}
Roggen, D., Hofmann, S., Thoma, Y., and Floreano, D. (2003).
\newblock Hardware spiking neural network with run-time reconfigurable
  connectivity in an autonomous robot.
\newblock In {\em Proc. NASA/DoD Conference on Evolvable Hardware, 2003}, pages
  189--198. IEEE.

\bibitem[Rohwer, 1989]{Rohwer:89}
Rohwer, R. (1989).
\newblock The `moving targets' training method.
\newblock In Kindermann, J. and Linden, A., editors, {\em Proceedings of
  `Distributed Adaptive Neural Information Processing', St.Augustin,
  24.-25.5,}. Oldenbourg.

\bibitem[Rosenblatt, 1958]{rosenblatt1958}
Rosenblatt, F. (1958).
\newblock The perceptron: a probabilistic model for information storage and
  organization in the brain.
\newblock {\em Psychological review}, 65(6):386.

\bibitem[Rosenblatt, 1962]{Rosenblatt:62}
Rosenblatt, F. (1962).
\newblock {\em Principles of Neurodynamics}.
\newblock Spartan, New York.

\bibitem[Roux et~al., 2013]{icpr12report}
Roux, L., Racoceanu, D., Lomenie, N., Kulikova, M., Irshad, H., Klossa, J.,
  Capron, F., Genestie, C., Naour, G.~L., and Gurcan, M.~N. (2013).
\newblock Mitosis detection in breast cancer histological images - an {ICPR
  2012} contest.
\newblock {\em J. Pathol. Inform.}, 4:8.

\bibitem[Rubner and Schulten, 1990]{RubnerSchulten:90}
Rubner, J. and Schulten, K. (1990).
\newblock Development of feature detectors by self-organization: A network
  model.
\newblock {\em Biological Cybernetics}, 62:193--199.

\bibitem[R{\"u}ckstie{\ss} et~al., 2008]{rueckstiess2008b}
R{\"u}ckstie{\ss}, T., Felder, M., and Schmidhuber, J. (2008).
\newblock {State-Dependent Exploration} for policy gradient methods.
\newblock In et~al., W.~D., editor, {\em European Conference on Machine
  Learning (ECML) and Principles and Practice of Knowledge Discovery in
  Databases 2008, Part II, LNAI 5212}, pages 234--249.

\bibitem[Rumelhart et~al., 1986]{Rumelhart:86}
Rumelhart, D.~E., Hinton, G.~E., and Williams, R.~J. (1986).
\newblock Learning internal representations by error propagation.
\newblock In Rumelhart, D.~E. and McClelland, J.~L., editors, {\em Parallel
  Distributed Processing}, volume~1, pages 318--362. MIT Press.

\bibitem[Rumelhart and Zipser, 1986]{RumelhartZipser:86}
Rumelhart, D.~E. and Zipser, D. (1986).
\newblock Feature discovery by competitive learning.
\newblock In {\em Parallel Distributed Processing}, pages 151--193. MIT Press.

\bibitem[Rummery and Niranjan, 1994]{Rummery:94}
Rummery, G. and Niranjan, M. (1994).
\newblock On-line {Q}-learning using connectionist sytems.
\newblock Technical Report {CUED/F}-{INFENG}-{TR} 166, Cambridge University,
  UK.

\bibitem[Russell et~al., 1995]{russell1995}
Russell, S.~J., Norvig, P., Canny, J.~F., Malik, J.~M., and Edwards, D.~D.
  (1995).
\newblock {\em Artificial Intelligence: a Modern Approach}, volume~2.
\newblock Englewood Cliffs: Prentice Hall.

\bibitem[Saito and Nakano, 1997]{Saito:1997}
Saito, K. and Nakano, R. (1997).
\newblock Partial {BFGS} update and efficient step-length calculation for
  three-layer neural networks.
\newblock {\em Neural Computation}, 9(1):123--141.

\bibitem[Sak et~al., 2014a]{sak2014large}
Sak, H., Senior, A., and Beaufays, F. (2014a).
\newblock {Long Short-Term Memory} recurrent neural network architectures for
  large scale acoustic modeling.
\newblock In {\em Proc. Interspeech}.

\bibitem[Sak et~al., 2014b]{sak2014}
Sak, H., Vinyals, O., Heigold, G., Senior, A., McDermott, E., Monga, R., and
  Mao, M. (2014b).
\newblock Sequence discriminative distributed training of {Long Short-Term
  Memory} recurrent neural networks.
\newblock In {\em Proc. Interspeech}.

\bibitem[Salakhutdinov and Hinton, 2009]{salakhutdinov2009}
Salakhutdinov, R. and Hinton, G. (2009).
\newblock Semantic hashing.
\newblock {\em Int. J. Approx. Reasoning}, 50(7):969--978.

\bibitem[Sallans and Hinton, 2004]{sallans2004}
Sallans, B. and Hinton, G. (2004).
\newblock Reinforcement learning with factored states and actions.
\newblock {\em Journal of Machine Learning Research}, 5:1063--1088.

\bibitem[Sa\l{}ustowicz and Schmidhuber, 1997]{Salustowicz:97ecj}
Sa\l{}ustowicz, R.~P. and Schmidhuber, J. (1997).
\newblock Probabilistic incremental program evolution.
\newblock {\em Evolutionary Computation}, 5(2):123--141.

\bibitem[Samejima et~al., 2003]{SamejimaDoyaKawato}
Samejima, K., Doya, K., and Kawato, M. (2003).
\newblock Inter-module credit assignment in modular reinforcement learning.
\newblock {\em Neural Networks}, 16(7):985--994.

\bibitem[Samuel, 1959]{Samuel:59}
Samuel, A.~L. (1959).
\newblock Some studies in machine learning using the game of checkers.
\newblock {\em IBM Journal on Research and Development}, 3:210--229.

\bibitem[Sanger, 1989]{Sanger:89}
Sanger, T.~D. (1989).
\newblock An optimality principle for unsupervised learning.
\newblock In Touretzky, D.~S., editor, {\em Advances in Neural Information
  Processing Systems (NIPS) 1}, pages 11--19. Morgan Kaufmann.

\bibitem[Santamar\'ia et~al., 1997]{Santamaria:97}
Santamar\'ia, J.~C., Sutton, R.~S., and Ram, A. (1997).
\newblock Experiments with reinforcement learning in problems with continuous
  state and action spaces.
\newblock {\em Adaptive Behavior}, 6(2):163--217.

\bibitem[Saravanan and Fogel, 1995]{saravanan:ieeeexpert95}
Saravanan, N. and Fogel, D.~B. (1995).
\newblock Evolving neural control systems.
\newblock {\em IEEE Expert}, pages 23--27.

\bibitem[Saund, 1994]{Saund:94}
Saund, E. (1994).
\newblock Unsupervised learning of mixtures of multiple causes in binary data.
\newblock In Cowan, J.~D., Tesauro, G., and Alspector, J., editors, {\em
  Advances in Neural Information Processing Systems (NIPS) 6}, pages 27--34.
  Morgan Kaufmann.

\bibitem[Schaback and Werner, 1992]{schaback1992}
Schaback, R. and Werner, H. (1992).
\newblock {\em {Numerische Mathematik}}, volume~4.
\newblock Springer.

\bibitem[Sch{\"a}fer et~al., 2006]{DBLP:conf/icann/SchaferUZ06}
Sch{\"a}fer, A.~M., Udluft, S., and Zimmermann, H.-G. (2006).
\newblock Learning long term dependencies with recurrent neural networks.
\newblock In Kollias, S.~D., Stafylopatis, A., Duch, W., and Oja, E., editors,
  {\em ICANN (1)}, volume 4131 of {\em Lecture Notes in Computer Science},
  pages 71--80. Springer.

\bibitem[Schapire, 1990]{Schapire:90}
Schapire, R.~E. (1990).
\newblock The strength of weak learnability.
\newblock {\em Machine Learning}, 5:197--227.

\bibitem[Schaul and Schmidhuber, 2010]{scholarpedia2010}
Schaul, T. and Schmidhuber, J. (2010).
\newblock Metalearning.
\newblock {\em Scholarpedia}, 6(5):4650.

\bibitem[Schaul et~al., 2013]{Schaul2012}
Schaul, T., Zhang, S., and LeCun, Y. (2013).
\newblock {No more pesky learning rates}.
\newblock In {\em Proc. 30th International Conference on Machine Learning
  (ICML)}.

\bibitem[Schemmel et~al., 2006]{schemmel2006}
Schemmel, J., Grubl, A., Meier, K., and Mueller, E. (2006).
\newblock Implementing synaptic plasticity in a {VLSI} spiking neural network
  model.
\newblock In {\em International Joint Conference on Neural Networks (IJCNN)},
  pages 1--6. IEEE.

\bibitem[Scherer et~al., 2010]{scherer:2010}
Scherer, D., M{\"u}ller, A., and Behnke, S. (2010).
\newblock Evaluation of pooling operations in convolutional architectures for
  object recognition.
\newblock In {\em Proc. International Conference on Artificial Neural Networks
  (ICANN)}, pages 92--101.

\bibitem[Schmidhuber, 1987]{schmidhuber87}
Schmidhuber, J. (1987).
\newblock {Evolutionary principles in self-referential learning, or on learning
  how to learn: the meta-meta-... hook. Diploma thesis, Inst. f. Inf., Tech.
  Univ. Munich}.
\newblock http://www.idsia.ch/\~{ }juergen/diploma.html.

\bibitem[Schmidhuber, 1989a]{Schmidhuber:89-1}
Schmidhuber, J. (1989a).
\newblock Accelerated learning in back-propagation nets.
\newblock In Pfeifer, R., Schreter, Z., Fogelman, Z., and Steels, L., editors,
  {\em Connectionism in Perspective}, pages 429 -- 438. Amsterdam: Elsevier,
  North-Holland.

\bibitem[Schmidhuber, 1989b]{Schmidhuber:89cs}
Schmidhuber, J. (1989b).
\newblock A local learning algorithm for dynamic feedforward and recurrent
  networks.
\newblock {\em Connection Science}, 1(4):403--412.

\bibitem[Schmidhuber, 1990a]{schmidhuber1990}
Schmidhuber, J. (1990a).
\newblock {Dynamische neuronale Netze und das fundamentale raumzeitliche
  Lernproblem. {\em (Dynamic neural nets and the fundamental spatio-temporal
  credit assignment problem.)} Dissertation, Inst. f. Inf., Tech. Univ.
  Munich}.

\bibitem[Schmidhuber, 1990b]{Schmidhuber:90cmss}
Schmidhuber, J. (1990b).
\newblock Learning algorithms for networks with internal and external feedback.
\newblock In Touretzky, D.~S., Elman, J.~L., Sejnowski, T.~J., and Hinton,
  G.~E., editors, {\em Proc. of the 1990 Connectionist Models Summer School},
  pages 52--61. Morgan Kaufmann.

\bibitem[Schmidhuber, 1990c]{heat90-96}
Schmidhuber, J. (1990c).
\newblock The {Neural Heat Exchanger}.
\newblock {Talks at TU Munich (1990), University of Colorado at Boulder (1992),
  and Z. Li's NIPS*94 workshop on unsupervised learning. Also published at the
  {\em Intl. Conference on Neural Information Processing} (ICONIP'96), vol. 1,
  pages 194-197, 1996}.

\bibitem[Schmidhuber, 1990d]{Schmidhuber:90sandiego}
Schmidhuber, J. (1990d).
\newblock An on-line algorithm for dynamic reinforcement learning and planning
  in reactive environments.
\newblock In {\em Proc. IEEE/INNS International Joint Conference on Neural
  Networks, San Diego}, volume~2, pages 253--258.

\bibitem[Schmidhuber, 1991a]{Schmidhuber:91singaporecur}
Schmidhuber, J. (1991a).
\newblock Curious model-building control systems.
\newblock In {\em Proceedings of the International Joint Conference on Neural
  Networks, Singapore}, volume~2, pages 1458--1463. IEEE press.

\bibitem[Schmidhuber, 1991b]{Schmidhuber:91icannsubgoals}
Schmidhuber, J. (1991b).
\newblock Learning to generate sub-goals for action sequences.
\newblock In Kohonen, T., M\"{a}kisara, K., Simula, O., and Kangas, J.,
  editors, {\em Artificial Neural Networks}, pages 967--972. Elsevier Science
  Publishers B.V., North-Holland.

\bibitem[Schmidhuber, 1991c]{Schmidhuber:91nips}
Schmidhuber, J. (1991c).
\newblock Reinforcement learning in {M}arkovian and non-{M}arkovian
  environments.
\newblock In Lippman, D.~S., Moody, J.~E., and Touretzky, D.~S., editors, {\em
  Advances in Neural Information Processing Systems 3 (NIPS 3)}, pages
  500--506. Morgan Kaufmann.

\bibitem[Schmidhuber, 1992a]{Schmidhuber:92ncn3}
Schmidhuber, J. (1992a).
\newblock A fixed size storage {$O(n^3)$} time complexity learning algorithm
  for fully recurrent continually running networks.
\newblock {\em Neural Computation}, 4(2):243--248.

\bibitem[Schmidhuber, 1992b]{chunker91and92}
Schmidhuber, J. (1992b).
\newblock Learning complex, extended sequences using the principle of history
  compression.
\newblock {\em Neural Computation}, 4(2):234--242.
\newblock (Based on TR FKI-148-91, TUM, 1991).

\bibitem[Schmidhuber, 1992c]{Schmidhuber:92ncfactorial}
Schmidhuber, J. (1992c).
\newblock Learning factorial codes by predictability minimization.
\newblock {\em Neural Computation}, 4(6):863--879.

\bibitem[Schmidhuber, 1993a]{Schmidhuber:93selfrefann}
Schmidhuber, J. (1993a).
\newblock An introspective network that can learn to run its own weight change
  algorithm.
\newblock In {\em Proc. of the Intl. Conf. on Artificial Neural Networks,
  Brighton}, pages 191--195. IEE.

\bibitem[Schmidhuber, 1993b]{schmidhuber1993}
Schmidhuber, J. (1993b).
\newblock {Netzwerkarchitekturen, Zielfunktionen und Kettenregel. {\em (Network
  architectures, objective functions, and chain rule.)} Habilitation Thesis,
  Inst. f. Inf., Tech. Univ. Munich}.

\bibitem[Schmidhuber, 1997]{Schmidhuber:97nn+}
Schmidhuber, J. (1997).
\newblock Discovering neural nets with low {Kolmogorov} complexity and high
  generalization capability.
\newblock {\em Neural Networks}, 10(5):857--873.

\bibitem[Schmidhuber, 2002]{Schmidhuber:02colt}
Schmidhuber, J. (2002).
\newblock The {Speed Prior:} a new simplicity measure yielding near-optimal
  computable predictions.
\newblock In Kivinen, J. and Sloan, R.~H., editors, {\em Proceedings of the
  15th Annual Conference on Computational Learning Theory (COLT 2002)}, Lecture
  Notes in Artificial Intelligence, pages 216--228. Springer, Sydney,
  Australia.

\bibitem[Schmidhuber, 2004]{Schmidhuber:04oops}
Schmidhuber, J. (2004).
\newblock Optimal ordered problem solver.
\newblock {\em Machine Learning}, 54:211--254.

\bibitem[Schmidhuber, 2006a]{Schmidhuber:06cs}
Schmidhuber, J. (2006a).
\newblock Developmental robotics, optimal artificial curiosity, creativity,
  music, and the fine arts.
\newblock {\em Connection Science}, 18(2):173--187.

\bibitem[Schmidhuber, 2006b]{Schmidhuber:05gmai}
Schmidhuber, J. (2006b).
\newblock G\"{o}del machines: Fully self-referential optimal universal
  self-improvers.
\newblock In Goertzel, B. and Pennachin, C., editors, {\em Artificial General
  Intelligence}, pages 199--226. Springer Verlag.
\newblock Variant available as arXiv:cs.LO/0309048.

\bibitem[Schmidhuber, 2007]{schmidhuber2007pro}
Schmidhuber, J. (2007).
\newblock Prototype resilient, self-modeling robots.
\newblock {\em Science}, 316(5825):688.

\bibitem[Schmidhuber, 2012]{Schmidhuber:12slimnn}
Schmidhuber, J. (2012).
\newblock Self-delimiting neural networks.
\newblock Technical Report IDSIA-08-12, arXiv:1210.0118v1 [cs.NE], The Swiss AI
  Lab IDSIA.

\bibitem[Schmidhuber, 2013a]{mydeep2013}
Schmidhuber, J. (2013a).
\newblock My first {Deep Learning} system of 1991 $+$ {Deep Learning} timeline
  1962-2013.
\newblock Technical Report arXiv:1312.5548v1 [cs.NE], The Swiss AI Lab IDSIA.

\bibitem[Schmidhuber, 2013b]{Schmidhuber:13powerplay}
Schmidhuber, J. (2013b).
\newblock {{\sc PowerPlay}: Training an Increasingly General Problem Solver by
  Continually Searching for the Simplest Still Unsolvable Problem}.
\newblock {\em Frontiers in Psychology}.

\bibitem[Schmidhuber et~al., 2011]{schmidhuber2011agi}
Schmidhuber, J., Ciresan, D., Meier, U., Masci, J., and Graves, A. (2011).
\newblock On fast deep nets for {AGI} vision.
\newblock In {\em {Proc. Fourth Conference on Artificial General Intelligence
  (AGI), Google, Mountain View, CA}}, pages 243--246.

\bibitem[Schmidhuber et~al., 1996]{Schmidhuber:96ncedges}
Schmidhuber, J., Eldracher, M., and Foltin, B. (1996).
\newblock Semilinear predictability minimization produces well-known feature
  detectors.
\newblock {\em Neural Computation}, 8(4):773--786.

\bibitem[Schmidhuber and Huber, 1991]{SchmidhuberHuber:91}
Schmidhuber, J. and Huber, R. (1991).
\newblock Learning to generate artificial fovea trajectories for target
  detection.
\newblock {\em International Journal of Neural Systems}, 2(1 \& 2):135--141.

\bibitem[Schmidhuber et~al., 1993]{SchmidhuberMozerPrelinger:93}
Schmidhuber, J., Mozer, M.~C., and Prelinger, D. (1993).
\newblock Continuous history compression.
\newblock In H\"{u}ning, H., Neuhauser, S., Raus, M., and Ritschel, W.,
  editors, {\em Proc. of Intl. Workshop on Neural Networks, RWTH Aachen}, pages
  87--95. Augustinus.

\bibitem[Schmidhuber and Prelinger, 1992]{siamese92and93}
Schmidhuber, J. and Prelinger, D. (1992).
\newblock Discovering predictable classifications.
\newblock Technical Report CU-CS-626-92, Dept. of Comp. Sci., University of
  Colorado at Boulder.
\newblock Published in {\em Neural Computation} 5(4):625-635 (1993).

\bibitem[Schmidhuber and Wahnsiedler, 1992]{SchmidhuberWahnsiedler:92sab}
Schmidhuber, J. and Wahnsiedler, R. (1992).
\newblock Planning simple trajectories using neural subgoal generators.
\newblock In Meyer, J.~A., Roitblat, H.~L., and Wilson, S.~W., editors, {\em
  Proc. of the 2nd International Conference on Simulation of Adaptive
  Behavior}, pages 196--202. MIT Press.

\bibitem[Schmidhuber et~al., 2007]{Schmidhuber:07nc}
Schmidhuber, J., Wierstra, D., Gagliolo, M., and Gomez, F.~J. (2007).
\newblock Training recurrent networks by {Evolino}.
\newblock {\em Neural Computation}, 19(3):757--779.

\bibitem[Schmidhuber et~al., 1997a]{Schmidhuber:97ssa}
Schmidhuber, J., Zhao, J., and Schraudolph, N. (1997a).
\newblock Reinforcement learning with self-modifying policies.
\newblock In Thrun, S. and Pratt, L., editors, {\em Learning to learn}, pages
  293--309. Kluwer.

\bibitem[Schmidhuber et~al., 1997b]{Schmidhuber:97bias}
Schmidhuber, J., Zhao, J., and Wiering, M. (1997b).
\newblock Shifting inductive bias with success-story algorithm, adaptive
  {Levin} search, and incremental self-improvement.
\newblock {\em Machine Learning}, 28:105--130.

\bibitem[Sch{\"o}lkopf et~al., 1998]{advkernel}
Sch{\"o}lkopf, B., Burges, C. J.~C., and Smola, A.~J., editors (1998).
\newblock {\em Advances in Kernel Methods - Support Vector Learning}.
\newblock {MIT} Press, Cambridge, MA.

\bibitem[Schraudolph and Sejnowski, 1993]{Schraudolph:93}
Schraudolph, N. and Sejnowski, T.~J. (1993).
\newblock Unsupervised discrimination of clustered data via optimization of
  binary information gain.
\newblock In Hanson, S.~J., Cowan, J.~D., and Giles, C.~L., editors, {\em
  Advances in Neural Information Processing Systems}, volume~5, pages 499--506.
  Morgan Kaufmann, San Mateo.

\bibitem[Schraudolph, 2002]{schraudolph02}
Schraudolph, N.~N. (2002).
\newblock Fast curvature matrix-vector products for second-order gradient
  descent.
\newblock {\em Neural Computation}, 14(7):1723--1738.

\bibitem[Schraudolph and Sejnowski, 1996]{Schraudolph:96}
Schraudolph, N.~N. and Sejnowski, T.~J. (1996).
\newblock Tempering backpropagation networks: Not all weights are created
  equal.
\newblock In Touretzky, D.~S., Mozer, M.~C., and Hasselmo, M.~E., editors, {\em
  Advances in Neural Information Processing Systems (NIPS)}, volume~8, pages
  563--569. The {MIT} Press, Cambridge, MA.

\bibitem[Schrauwen et~al., 2007]{schrauwen2007}
Schrauwen, B., Verstraeten, D., and Van~Campenhout, J. (2007).
\newblock An overview of reservoir computing: theory, applications and
  implementations.
\newblock In {\em Proceedings of the 15th European Symposium on Artificial
  Neural Networks. p. 471-482 2007}, pages 471--482.

\bibitem[Schuster, 1992]{Schuster:92}
Schuster, H.~G. (1992).
\newblock Learning by maximization the information transfer through nonlinear
  noisy neurons and ``noise breakdown''.
\newblock {\em Phys. Rev. {\bf A}}, 46(4):2131--2138.

\bibitem[Schuster, 1999]{schuster99thesis}
Schuster, M. (1999).
\newblock {\em On supervised learning from sequential data with applications
  for speech recognition}.
\newblock PhD thesis, Nara Institute of Science and Technolog, Kyoto, Japan.

\bibitem[Schuster and Paliwal, 1997]{schuster97bidirectional}
Schuster, M. and Paliwal, K.~K. (1997).
\newblock Bidirectional recurrent neural networks.
\newblock {\em IEEE Transactions on Signal Processing}, 45:2673--2681.

\bibitem[Schwartz, 1993]{Schwartz:93}
Schwartz, A. (1993).
\newblock A reinforcement learning method for maximizing undiscounted rewards.
\newblock In {\em Proc. ICML}, pages 298--305.

\bibitem[Schwefel, 1974]{Schwefel:74}
Schwefel, H.~P. (1974).
\newblock {Numerische Optimierung von Computer-Modellen. Dissertation}.
\newblock Published 1977 by Birkh\"{a}user, Basel.

\bibitem[{Segmentation of Neuronal Structures in EM Stacks Challenge},
  2012]{isbi12}
{Segmentation of Neuronal Structures in EM Stacks Challenge} (2012).
\newblock {IEEE International Symposium on Biomedical Imaging (ISBI),
  http://tinyurl.com/d2fgh7g}.

\bibitem[Sehnke et~al., 2010]{sehnke2009parameter}
Sehnke, F., Osendorfer, C., R{\"u}ckstie{\ss}, T., Graves, A., Peters, J., and
  Schmidhuber, J. (2010).
\newblock Parameter-exploring policy gradients.
\newblock {\em Neural Networks}, 23(4):551--559.

\bibitem[Sermanet et~al., 2013]{sermanet2013}
Sermanet, P., Eigen, D., Zhang, X., Mathieu, M., Fergus, R., and LeCun, Y.
  (2013).
\newblock {OverFeat}: Integrated recognition, localization and detection using
  convolutional networks.
\newblock {\em arXiv preprint arXiv:1312.6229}.

\bibitem[Sermanet and LeCun, 2011]{sermanet-ijcnn-11}
Sermanet, P. and LeCun, Y. (2011).
\newblock Traffic sign recognition with multi-scale convolutional networks.
\newblock In {\em Proceedings of International Joint Conference on Neural
  Networks (IJCNN'11)}, pages 2809--2813.

\bibitem[Serrano-Gotarredona et~al., 2009]{serrano2009}
Serrano-Gotarredona, R., Oster, M., Lichtsteiner, P., Linares-Barranco, A.,
  Paz-Vicente, R., G{\'o}mez-Rodr{\'\i}guez, F., Camu{\~n}as-Mesa, L., Berner,
  R., Rivas-P{\'e}rez, M., Delbruck, T., et~al. (2009).
\newblock Caviar: A 45k neuron, 5m synapse, 12g connects/s {AER} hardware
  sensory--processing--learning--actuating system for high-speed visual object
  recognition and tracking.
\newblock {\em IEEE Transactions on Neural Networks}, 20(9):1417--1438.

\bibitem[Serre et~al., 2002]{serre2002}
Serre, T., Riesenhuber, M., Louie, J., and Poggio, T. (2002).
\newblock On the role of object-specific features for real world object
  recognition in biological vision.
\newblock In {\em Biologically Motivated Computer Vision}, pages 387--397.

\bibitem[Seung, 2003]{seung2003}
Seung, H.~S. (2003).
\newblock Learning in spiking neural networks by reinforcement of stochastic
  synaptic transmission.
\newblock {\em Neuron}, 40(6):1063--1073.

\bibitem[Shan and Cottrell, 2014]{shan2014}
Shan, H. and Cottrell, G. (2014).
\newblock Efficient visual coding: From retina to {V2}.
\newblock In {\em Proc. International Conference on Learning Representations
  (ICLR)}.
\newblock arXiv preprint arXiv:1312.6077.

\bibitem[Shan et~al., 2007]{shan2007}
Shan, H., Zhang, L., and Cottrell, G.~W. (2007).
\newblock Recursive {ICA}.
\newblock {\em Advances in Neural Information Processing Systems (NIPS)},
  19:1273.

\bibitem[Shanno, 1970]{shanno1970}
Shanno, D.~F. (1970).
\newblock Conditioning of quasi-{Newton} methods for function minimization.
\newblock {\em Mathematics of computation}, 24(111):647--656.

\bibitem[Shannon, 1948]{Shannon:48}
Shannon, C.~E. (1948).
\newblock A mathematical theory of communication (parts {I} and {II}).
\newblock {\em Bell System Technical Journal}, XXVII:379--423.

\bibitem[Shao et~al., 2014]{shao2014}
Shao, L., Wu, D., and Li, X. (2014).
\newblock Learning deep and wide: A spectral method for learning deep networks.
\newblock {\em IEEE Transactions on Neural Networks and Learning Systems}.

\bibitem[Shavlik, 1994]{shavlik1994}
Shavlik, J.~W. (1994).
\newblock Combining symbolic and neural learning.
\newblock {\em Machine Learning}, 14(3):321--331.

\bibitem[Shavlik and Towell, 1989]{shavlik1989}
Shavlik, J.~W. and Towell, G.~G. (1989).
\newblock Combining explanation-based and neural learning: An algorithm and
  empirical results.
\newblock {\em Connection Science}, 1(3):233--255.

\bibitem[Siegelmann, 1992]{siegelmann93foundations}
Siegelmann, H. (1992).
\newblock {\em Theoretical Foundations of Recurrent Neural Networks}.
\newblock PhD thesis, Rutgers, New Brunswick Rutgers, The State of New Jersey.

\bibitem[Siegelmann and Sontag, 1991]{siegelmann91turing}
Siegelmann, H.~T. and Sontag, E.~D. (1991).
\newblock Turing computability with neural nets.
\newblock {\em Applied Mathematics Letters}, 4(6):77--80.

\bibitem[Silva and Almeida, 1990]{SilvaAlmeida:1990}
Silva, F.~M. and Almeida, L.~B. (1990).
\newblock Speeding up back-propagation.
\newblock In Eckmiller, R., editor, {\em Advanced Neural Computers}, pages
  151--158, Amsterdam. Elsevier.

\bibitem[S\'{\i}ma, 1994]{sima1994}
S\'{\i}ma, J. (1994).
\newblock Loading deep networks is hard.
\newblock {\em Neural Computation}, 6(5):842--850.

\bibitem[S\'{\i}ma, 2002]{sima2002}
S\'{\i}ma, J. (2002).
\newblock Training a single sigmoidal neuron is hard.
\newblock {\em Neural Computation}, 14(11):2709--2728.

\bibitem[Simard et~al., 2003]{simard:2003}
Simard, P., Steinkraus, D., and Platt, J. (2003).
\newblock Best practices for convolutional neural networks applied to visual
  document analysis.
\newblock In {\em Seventh International Conference on Document Analysis and
  Recognition}, pages 958--963.

\bibitem[Sims, 1994]{Sims:1994:EVC}
Sims, K. (1994).
\newblock Evolving virtual creatures.
\newblock In Glassner, A., editor, {\em Proceedings of SIGGRAPH '94 (Orlando,
  Florida, July 1994)}, Computer Graphics Proceedings, Annual Conference, pages
  15--22. ACM SIGGRAPH, ACM Press.
\newblock ISBN 0-89791-667-0.

\bibitem[Simsek and Barto, 2008]{simsek2008skill}
Simsek, {\"O}. and Barto, A.~G. (2008).
\newblock Skill characterization based on betweenness.
\newblock In {\em NIPS'08}, pages 1497--1504.

\bibitem[Singh et~al., 2005]{Singh:05nips}
Singh, S., Barto, A.~G., and Chentanez, N. (2005).
\newblock Intrinsically motivated reinforcement learning.
\newblock In {\em {Advances in Neural Information Processing Systems 17
  (NIPS)}}. MIT Press, Cambridge, MA.

\bibitem[Singh, 1994]{Singh:94R}
Singh, S.~P. (1994).
\newblock Reinforcement learning algorithms for average-payoff {Markovian}
  decision processes.
\newblock In {\em National Conference on Artificial Intelligence}, pages
  700--705.

\bibitem[Smith, 1980]{smith80}
Smith, S.~F. (1980).
\newblock {\em A Learning System Based on Genetic Adaptive Algorithms,}.
\newblock PhD thesis, Univ. Pittsburgh.

\bibitem[Smolensky, 1986]{smolensky86}
Smolensky, P. (1986).
\newblock Parallel distributed processing: Explorations in the microstructure
  of cognition, vol. 1.
\newblock chapter Information Processing in Dynamical Systems: Foundations of
  Harmony Theory, pages 194--281. MIT Press, Cambridge, MA, USA.

\bibitem[Solla, 1988]{Solla:88}
Solla, S.~A. (1988).
\newblock Accelerated learning in layered neural networks.
\newblock {\em Complex Systems}, 2:625--640.

\bibitem[Solomonoff, 1964]{Solomonoff:64}
Solomonoff, R.~J. (1964).
\newblock A formal theory of inductive inference. {Part I}.
\newblock {\em Information and Control}, 7:1--22.

\bibitem[Solomonoff, 1978]{Solomonoff:78}
Solomonoff, R.~J. (1978).
\newblock Complexity-based induction systems.
\newblock {\em IEEE Transactions on Information Theory}, IT-24(5):422--432.

\bibitem[Soloway, 1986]{soloway1986}
Soloway, E. (1986).
\newblock Learning to program $=$ learning to construct mechanisms and
  explanations.
\newblock {\em Communications of the ACM}, 29(9):850--858.

\bibitem[Song et~al., 2000]{song2000}
Song, S., Miller, K.~D., and Abbott, L.~F. (2000).
\newblock Competitive {Hebbian} learning through spike-timing-dependent
  synaptic plasticity.
\newblock {\em Nature Neuroscience}, 3(9):919--926.

\bibitem[Speelpenning, 1980]{SPEELPENNING80A}
Speelpenning, B. (1980).
\newblock {\em Compiling Fast Partial Derivatives of Functions Given by
  Algorithms}.
\newblock PhD thesis, Department of Computer Science, University of Illinois,
  Urbana-Champaign.

\bibitem[Srivastava et~al., 2013]{srivastava2013compete}
Srivastava, R.~K., Masci, J., Kazerounian, S., Gomez, F., and Schmidhuber, J.
  (2013).
\newblock Compete to compute.
\newblock In {\em Advances in Neural Information Processing Systems (NIPS)},
  pages 2310--2318.

\bibitem[Stallkamp et~al., 2011]{stallkamp:11}
Stallkamp, J., Schlipsing, M., Salmen, J., and Igel, C. (2011).
\newblock The {G}erman traffic sign recognition benchmark: {A} multi-class
  classification competition.
\newblock In {\em International Joint Conference on Neural Networks (IJCNN
  2011)}, pages 1453--1460. IEEE Press.

\bibitem[Stallkamp et~al., 2012]{stallkamp:12}
Stallkamp, J., Schlipsing, M., Salmen, J., and Igel, C. (2012).
\newblock Man vs.~computer: {B}enchmarking machine learning algorithms for
  traffic sign recognition.
\newblock {\em Neural Networks}, 32:323--332.

\bibitem[Stanley et~al., 2009]{stanley09}
Stanley, K.~O., D'Ambrosio, D.~B., and Gauci, J. (2009).
\newblock A hypercube-based encoding for evolving large-scale neural networks.
\newblock {\em Artificial Life}, 15(2):185--212.

\bibitem[Stanley and Miikkulainen, 2002]{stanley:ec02}
Stanley, K.~O. and Miikkulainen, R. (2002).
\newblock Evolving neural networks through augmenting topologies.
\newblock {\em Evolutionary Computation}, 10:99--127.

\bibitem[Steijvers and Grunwald, 1996]{steijvers96recurrent}
Steijvers, M. and Grunwald, P. (1996).
\newblock A recurrent network that performs a contextsensitive prediction task.
\newblock In {\em Proceedings of the 18th Annual Conference of the Cognitive
  Science Society}. Erlbaum.

\bibitem[Steil, 2007]{steil2007}
Steil, J.~J. (2007).
\newblock Online reservoir adaptation by intrinsic plasticity for
  backpropagation--decorrelation and echo state learning.
\newblock {\em Neural Networks}, 20(3):353--364.

\bibitem[Stemmler, 1996]{stemmler1996}
Stemmler, M. (1996).
\newblock A single spike suffices: the simplest form of stochastic resonance in
  model neurons.
\newblock {\em Network: Computation in Neural Systems}, 7(4):687--716.

\bibitem[Stoianov and Zorzi, 2012]{stoianov2012}
Stoianov, I. and Zorzi, M. (2012).
\newblock Emergence of a 'visual number sense' in hierarchical generative
  models.
\newblock {\em Nature Neuroscience}, 15(2):194--6.

\bibitem[Stone, 1974]{Stone:74}
Stone, M. (1974).
\newblock Cross-validatory choice and assessment of statistical predictions.
\newblock {\em Roy. Stat. Soc.}, 36:111--147.

\bibitem[Stoop et~al., 2000]{stoop2000}
Stoop, R., Schindler, K., and Bunimovich, L. (2000).
\newblock When pyramidal neurons lock, when they respond chaotically, and when
  they like to synchronize.
\newblock {\em Neuroscience research}, 36(1):81--91.

\bibitem[Stratonovich, 1960]{stratonovich1960}
Stratonovich, R. (1960).
\newblock Conditional {Markov} processes.
\newblock {\em Theory of Probability And Its Applications}, 5(2):156--178.

\bibitem[Sun et~al., 1993a]{Sun:93}
Sun, G., Chen, H., and Lee, Y. (1993a).
\newblock Time warping invariant neural networks.
\newblock In S.~J.~Hanson, J. D.~C. and Giles, C.~L., editors, {\em Advances in
  Neural Information Processing Systems (NIPS) 5}, pages 180--187. Morgan
  Kaufmann.

\bibitem[Sun et~al., 1993b]{Sun93:abRNN}
Sun, G.~Z., Giles, C.~L., Chen, H.~H., and Lee, Y.~C. (1993b).
\newblock The neural network pushdown automaton: Model, stack and learning
  simulations.
\newblock Technical Report CS-TR-3118, University of Maryland, College Park.

\bibitem[Sun et~al., 2013]{sun:gecco13}
Sun, Y., Gomez, F., Schaul, T., and Schmidhuber, J. (2013).
\newblock {A Linear Time Natural Evolution Strategy for Non-Separable
  Functions}.
\newblock In {\em Proceedings of the Genetic and Evolutionary Computation
  Conference}, page~61, Amsterdam, NL. ACM.

\bibitem[Sun et~al., 2009]{Sun2009a}
Sun, Y., Wierstra, D., Schaul, T., and Schmidhuber, J. (2009).
\newblock Efficient natural evolution strategies.
\newblock In {\em Proc. 11th Genetic and Evolutionary Computation Conference
  (GECCO)}, pages 539--546.

\bibitem[Sutskever et~al., 2008]{sutskever2008}
Sutskever, I., Hinton, G.~E., and Taylor, G.~W. (2008).
\newblock The recurrent temporal restricted {Boltzmann} machine.
\newblock In {\em NIPS}, volume~21, page 2008.

\bibitem[Sutskever et~al., 2014]{sutskever2014}
Sutskever, I., Vinyals, O., and Le, Q.~V. (2014).
\newblock Sequence to sequence learning with neural networks.
\newblock Technical Report arXiv:1409.3215 [cs.CL], Google.
\newblock NIPS'2014.

\bibitem[Sutton and Barto, 1998]{Sutton:98}
Sutton, R. and Barto, A. (1998).
\newblock {\em Reinforcement learning: An introduction}.
\newblock Cambridge, MA, MIT Press.

\bibitem[Sutton et~al., 1999a]{Sutton:99}
Sutton, R.~S., McAllester, D.~A., Singh, S.~P., and Mansour, Y. (1999a).
\newblock Policy gradient methods for reinforcement learning with function
  approximation.
\newblock In {\em Advances in Neural Information Processing Systems (NIPS) 12},
  pages 1057--1063.

\bibitem[Sutton et~al., 1999b]{sutton1999between}
Sutton, R.~S., Precup, D., and Singh, S.~P. (1999b).
\newblock Between {MDP}s and semi-{MDP}s: A framework for temporal abstraction
  in reinforcement learning.
\newblock {\em Artif. Intell.}, 112(1-2):181--211.

\bibitem[Sutton et~al., 2008]{09-Gtd}
Sutton, R.~S., Szepesv{\'a}ri, C., and Maei, H.~R. (2008).
\newblock A convergent {O(n)} algorithm for off-policy temporal-difference
  learning with linear function approximation.
\newblock In {\em Advances in Neural Information Processing Systems (NIPS'08)},
  volume~21, pages 1609--1616.

\bibitem[Szab{\'o} et~al., 2006]{szabo2006}
Szab{\'o}, Z., P{\'o}czos, B., and L{\H{o}}rincz, A. (2006).
\newblock Cross-entropy optimization for independent process analysis.
\newblock In {\em Independent Component Analysis and Blind Signal Separation},
  pages 909--916. Springer.

\bibitem[Szegedy et~al., 2014]{szegedy2014}
Szegedy, C., Liu, W., Jia, Y., Sermanet, P., Reed, S., Anguelov, D., Erhan, D.,
  Vanhoucke, V., and Rabinovich, A. (2014).
\newblock Going deeper with convolutions.
\newblock Technical Report arXiv:1409.4842 [cs.CV], Google.

\bibitem[Szegedy et~al., 2013]{Szegedy2013}
Szegedy, C., Toshev, A., and Erhan, D. (2013).
\newblock Deep neural networks for object detection.
\newblock pages 2553--2561.

\bibitem[Taylor et~al., 2011]{taylor2011}
Taylor, G.~W., Spiro, I., Bregler, C., and Fergus, R. (2011).
\newblock Learning invariance through imitation.
\newblock In {\em Conference on Computer Vision and Pattern Recognition
  (CVPR)}, pages 2729--2736. IEEE.

\bibitem[Tegge et~al., 2009]{tegge2009}
Tegge, A.~N., Wang, Z., Eickholt, J., and Cheng, J. (2009).
\newblock {NNcon}: improved protein contact map prediction using {2D}-recursive
  neural networks.
\newblock {\em Nucleic Acids Research}, 37(Suppl 2):W515--W518.

\bibitem[Teichmann et~al., 2012]{teichmann2012}
Teichmann, M., Wiltschut, J., and Hamker, F. (2012).
\newblock Learning invariance from natural images inspired by observations in
  the primary visual cortex.
\newblock {\em Neural Computation}, 24(5):1271--1296.

\bibitem[Teller, 1994]{Teller:94}
Teller, A. (1994).
\newblock The evolution of mental models.
\newblock In Kenneth E.~Kinnear, J., editor, {\em Advances in Genetic
  Programming}, pages 199--219. MIT Press.

\bibitem[Tenenberg et~al., 1993]{TenenbergKarlssonWhitehead}
Tenenberg, J., Karlsson, J., and Whitehead, S. (1993).
\newblock Learning via task decomposition.
\newblock In Meyer, J.~A., Roitblat, H., and Wilson, S., editors, {\em From
  Animals to Animats 2: Proceedings of the Second International Conference on
  Simulation of Adaptive Behavior}, pages 337--343. MIT Press.

\bibitem[Tesauro, 1994]{Tesauro:94}
Tesauro, G. (1994).
\newblock {TD}-gammon, a self-teaching backgammon program, achieves
  master-level play.
\newblock {\em Neural Computation}, 6(2):215--219.

\bibitem[Tieleman and Hinton, 2012]{Tieleman2012}
Tieleman, T. and Hinton, G. (2012).
\newblock {Lecture 6.5---RmsProp: Divide the gradient by a running average of
  its recent magnitude}.
\newblock COURSERA: Neural Networks for Machine Learning.

\bibitem[Tikhonov et~al., 1977]{tikhonov1977}
Tikhonov, A.~N., Arsenin, V.~I., and John, F. (1977).
\newblock {\em Solutions of ill-posed problems}.
\newblock Winston.

\bibitem[Ting and Witten, 1997]{ting1997}
Ting, K.~M. and Witten, I.~H. (1997).
\newblock Stacked generalization: when does it work?
\newblock In {\em in Proc. International Joint Conference on Artificial
  Intelligence (IJCAI)}.

\bibitem[Ti\v{n}o and Hammer, 2004]{Tino03NC}
Ti\v{n}o, P. and Hammer, B. (2004).
\newblock Architectural bias in recurrent neural networks: Fractal analysis.
\newblock {\em Neural Computation}, 15(8):1931--1957.

\bibitem[Tonkes and Wiles, 1997]{tonkes97learning}
Tonkes, B. and Wiles, J. (1997).
\newblock Learning a context-free task with a recurrent neural network: An
  analysis of stability.
\newblock In {\em Proceedings of the Fourth Biennial Conference of the
  Australasian Cognitive Science Society}.

\bibitem[Towell and Shavlik, 1994]{towell1994}
Towell, G.~G. and Shavlik, J.~W. (1994).
\newblock Knowledge-based artificial neural networks.
\newblock {\em Artificial Intelligence}, 70(1):119--165.

\bibitem[Tsitsiklis and van Roy, 1996]{Tsitsiklis:96}
Tsitsiklis, J.~N. and van Roy, B. (1996).
\newblock Feature-based methods for large scale dynamic programming.
\newblock {\em Machine Learning}, 22(1-3):59--94.

\bibitem[Tsodyks et~al., 1998]{tsodyks1998}
Tsodyks, M., Pawelzik, K., and Markram, H. (1998).
\newblock Neural networks with dynamic synapses.
\newblock {\em Neural Computation}, 10(4):821--835.

\bibitem[Tsodyks et~al., 1996]{tsodyks1996}
Tsodyks, M.~V., Skaggs, W.~E., Sejnowski, T.~J., and McNaughton, B.~L. (1996).
\newblock Population dynamics and theta rhythm phase precession of hippocampal
  place cell firing: a spiking neuron model.
\newblock {\em Hippocampus}, 6(3):271--280.

\bibitem[Turaga et~al., 2010]{turaga2010}
Turaga, S.~C., Murray, J.~F., Jain, V., Roth, F., Helmstaedter, M., Briggman,
  K., Denk, W., and Seung, H.~S. (2010).
\newblock Convolutional networks can learn to generate affinity graphs for
  image segmentation.
\newblock {\em Neural Computation}, 22(2):511--538.

\bibitem[Turing, 1936]{Turing:36}
Turing, A.~M. (1936).
\newblock On computable numbers, with an application to the
  {Entscheidungsproblem}.
\newblock {\em Proceedings of the London Mathematical Society, Series 2},
  41:230--267.

\bibitem[Turner and Miller, 2013]{turner2013}
Turner, A.~J. and Miller, J.~F. (2013).
\newblock Cartesian {Genetic Programming} encoded artificial neural networks: A
  comparison using three benchmarks.
\newblock In {\em Proceedings of the Conference on Genetic and Evolutionary
  Computation (GECCO)}, pages 1005--1012.

\bibitem[Ueda, 2000]{Ueda2000}
Ueda, N. (2000).
\newblock {Optimal linear combination of neural networks for improving
  classification performance}.
\newblock {\em IEEE Transactions on Pattern Analysis and Machine Intelligence},
  22(2):207--215.

\bibitem[Urlbe, 1999]{urlbe1999}
Urlbe, A.~P. (1999).
\newblock {\em Structure-adaptable digital neural networks}.
\newblock PhD thesis, Universidad del Valle.

\bibitem[Utgoff and Stracuzzi, 2002]{utgoff2002}
Utgoff, P.~E. and Stracuzzi, D.~J. (2002).
\newblock Many-layered learning.
\newblock {\em Neural Computation}, 14(10):2497--2529.

\bibitem[Vahed and Omlin, 2004]{Omlin:04}
Vahed, A. and Omlin, C.~W. (2004).
\newblock A machine learning method for extracting symbolic knowledge from
  recurrent neural networks.
\newblock {\em Neural Computation}, 16(1):59--71.

\bibitem[Vaillant et~al., 1994]{vaillant-monrocq-lecun-94}
Vaillant, R., Monrocq, C., and LeCun, Y. (1994).
\newblock Original approach for the localisation of objects in images.
\newblock {\em IEE Proc on Vision, Image, and Signal Processing},
  141(4):245--250.

\bibitem[van~den Berg and Whiteson, 2013]{vandenberg2013}
van~den Berg, T. and Whiteson, S. (2013).
\newblock Critical factors in the performance of {HyperNEAT}.
\newblock In {\em GECCO 2013: Proceedings of the Genetic and Evolutionary
  Computation Conference}, pages 759--766.

\bibitem[van Hasselt, 2012]{hasselt2012}
van Hasselt, H. (2012).
\newblock Reinforcement learning in continuous state and action spaces.
\newblock In Wiering, M. and van Otterlo, M., editors, {\em Reinforcement
  Learning}, pages 207--251. Springer.

\bibitem[Vapnik, 1992]{Vapnik:92}
Vapnik, V. (1992).
\newblock Principles of risk minimization for learning theory.
\newblock In Lippman, D.~S., Moody, J.~E., and Touretzky, D.~S., editors, {\em
  Advances in Neural Information Processing Systems (NIPS) 4}, pages 831--838.
  Morgan Kaufmann.

\bibitem[Vapnik, 1995]{Vapnik:95}
Vapnik, V. (1995).
\newblock {\em The Nature of Statistical Learning Theory}.
\newblock Springer, New York.

\bibitem[Versino and Gambardella, 1996]{versino1996}
Versino, C. and Gambardella, L.~M. (1996).
\newblock Learning fine motion by using the hierarchical extended {Kohonen}
  map.
\newblock In {\em Proc. Intl. Conf. on Artificial Neural Networks (ICANN)},
  pages 221--226. Springer.

\bibitem[Veta et~al., 2013]{miccai13}
Veta, M., Viergever, M., Pluim, J., Stathonikos, N., and van Diest, P.~J.
  (2013).
\newblock {MICCAI 2013 Grand Challenge on Mitosis Detection}.

\bibitem[Vieira and Barradas, 2003]{vieira2003}
Vieira, A. and Barradas, N. (2003).
\newblock A training algorithm for classification of high-dimensional data.
\newblock {\em Neurocomputing}, 50:461--472.

\bibitem[Viglione, 1970]{viglione1970}
Viglione, S. (1970).
\newblock Applications of pattern recognition technology.
\newblock In Mendel, J.~M. and Fu, K.~S., editors, {\em Adaptive, Learning, and
  Pattern Recognition Systems}. Academic Press.

\bibitem[Vincent et~al., 2008]{vincent:2008}
Vincent, P., Hugo, L., Bengio, Y., and Manzagol, P.-A. (2008).
\newblock Extracting and composing robust features with denoising autoencoders.
\newblock In {\em Proceedings of the 25th international conference on Machine
  learning}, ICML '08, pages 1096--1103, New York, NY, USA. ACM.

\bibitem[Vlassis et~al., 2012]{vlassis2012}
Vlassis, N., Littman, M.~L., and Barber, D. (2012).
\newblock On the computational complexity of stochastic controller optimization
  in {POMDPs}.
\newblock {\em ACM Transactions on Computation Theory}, 4(4):12.

\bibitem[Vogl et~al., 1988]{Vogl:88}
Vogl, T., Mangis, J., Rigler, A., Zink, W., and Alkon, D. (1988).
\newblock Accelerating the convergence of the back-propagation method.
\newblock {\em Biological Cybernetics}, 59:257--263.

\bibitem[von~der Malsburg, 1973]{malsburg1973}
von~der Malsburg, C. (1973).
\newblock Self-organization of orientation sensitive cells in the striate
  cortex.
\newblock {\em Kybernetik}, 14(2):85--100.

\bibitem[Waldinger and Lee, 1969]{waldinger69}
Waldinger, R.~J. and Lee, R. C.~T. (1969).
\newblock {PROW:} a step toward automatic program writing.
\newblock In Walker, D.~E. and Norton, L.~M., editors, {\em Proceedings of the
  1st International Joint Conference on Artificial Intelligence (IJCAI)}, pages
  241--252. Morgan Kaufmann.

\bibitem[Wallace and Boulton, 1968]{Wallace:68}
Wallace, C.~S. and Boulton, D.~M. (1968).
\newblock An information theoretic measure for classification.
\newblock {\em Computer Journal}, 11(2):185--194.

\bibitem[Wan, 1994]{wan1994}
Wan, E.~A. (1994).
\newblock Time series prediction by using a connectionist network with internal
  delay lines.
\newblock In Weigend, A.~S. and Gershenfeld, N.~A., editors, {\em Time series
  prediction: Forecasting the future and understanding the past}, pages
  265--295. Addison-Wesley.

\bibitem[Wang et~al., 1994]{Wang:94}
Wang, C., Venkatesh, S.~S., and Judd, J.~S. (1994).
\newblock Optimal stopping and effective machine complexity in learning.
\newblock In {\em Advances in Neural Information Processing Systems (NIPS'6)},
  pages 303--310. Morgan Kaufmann.

\bibitem[Wang and Manning, 2013]{wang2013fast}
Wang, S. and Manning, C. (2013).
\newblock Fast dropout training.
\newblock In {\em Proceedings of the 30th International Conference on Machine
  Learning (ICML-13)}, pages 118--126.

\bibitem[Watanabe, 1992]{Watanabe:92}
Watanabe, O. (1992).
\newblock {\em {Kolmogorov} complexity and computational complexity}.
\newblock EATCS Monographs on Theoretical Computer Science, Springer.

\bibitem[Watanabe, 1985]{Watanabe:85}
Watanabe, S. (1985).
\newblock {\em Pattern Recognition: Human and Mechanical}.
\newblock Willey, New York.

\bibitem[Watkins, 1989]{Watkins:89}
Watkins, C. J. C.~H. (1989).
\newblock {\em Learning from Delayed Rewards}.
\newblock PhD thesis, King's College, Oxford.

\bibitem[Watkins and Dayan, 1992]{WatkinsDayan:92}
Watkins, C. J. C.~H. and Dayan, P. (1992).
\newblock Q-learning.
\newblock {\em Machine Learning}, 8:279--292.

\bibitem[Watrous and Kuhn, 1992]{Watrous:92nips}
Watrous, R.~L. and Kuhn, G.~M. (1992).
\newblock Induction of finite-state automata using second-order recurrent
  networks.
\newblock In Moody, J.~E., Hanson, S.~J., and Lippman, R.~P., editors, {\em
  Advances in Neural Information Processing Systems 4}, pages 309--316. Morgan
  Kaufmann.

\bibitem[Waydo and Koch, 2008]{koch2008}
Waydo, S. and Koch, C. (2008).
\newblock Unsupervised learning of individuals and categories from images.
\newblock {\em Neural Computation}, 20(5):1165--1178.

\bibitem[Weigend and Gershenfeld, 1993]{weigend1993}
Weigend, A.~S. and Gershenfeld, N.~A. (1993).
\newblock Results of the time series prediction competition at the {Santa Fe
  Institute}.
\newblock In {\em Neural Networks, 1993., IEEE International Conference on},
  pages 1786--1793. IEEE.

\bibitem[Weigend et~al., 1991]{Weigend:91}
Weigend, A.~S., Rumelhart, D.~E., and Huberman, B.~A. (1991).
\newblock Generalization by weight-elimination with application to forecasting.
\newblock In Lippmann, R.~P., Moody, J.~E., and Touretzky, D.~S., editors, {\em
  Advances in Neural Information Processing Systems (NIPS) 3}, pages 875--882.
  San Mateo, CA: Morgan Kaufmann.

\bibitem[Weiss, 1994]{Weiss:94a}
Weiss, G. (1994).
\newblock Hierarchical chunking in classifier systems.
\newblock In {\em Proceedings of the 12th National Conference on Artificial
  Intelligence}, volume~2, pages 1335--1340. AAAI Press/The MIT Press.

\bibitem[Weng et~al., 1992]{weng1992}
Weng, J., Ahuja, N., and Huang, T.~S. (1992).
\newblock Cresceptron: a self-organizing neural network which grows adaptively.
\newblock In {\em International Joint Conference on Neural Networks (IJCNN)},
  volume~1, pages 576--581. IEEE.

\bibitem[Weng et~al., 1997]{weng1997}
Weng, J.~J., Ahuja, N., and Huang, T.~S. (1997).
\newblock Learning recognition and segmentation using the cresceptron.
\newblock {\em International Journal of Computer Vision}, 25(2):109--143.

\bibitem[Werbos, 1974]{Werbos:74}
Werbos, P.~J. (1974).
\newblock {\em Beyond Regression: New Tools for Prediction and Analysis in the
  Behavioral Sciences}.
\newblock PhD thesis, Harvard University.

\bibitem[Werbos, 1981]{Werbos:81sensitivity}
Werbos, P.~J. (1981).
\newblock Applications of advances in nonlinear sensitivity analysis.
\newblock In {\em Proceedings of the 10th IFIP Conference, 31.8 - 4.9, NYC},
  pages 762--770.

\bibitem[Werbos, 1987]{Werbos:87}
Werbos, P.~J. (1987).
\newblock Building and understanding adaptive systems: A statistical/numerical
  approach to factory automation and brain research.
\newblock {\em IEEE Transactions on Systems, Man, and Cybernetics}, 17.

\bibitem[Werbos, 1988]{Werbos:88gasmarket}
Werbos, P.~J. (1988).
\newblock Generalization of backpropagation with application to a recurrent gas
  market model.
\newblock {\em Neural Networks}, 1.

\bibitem[Werbos, 1989a]{Werbos:89neurocontrol}
Werbos, P.~J. (1989a).
\newblock Backpropagation and neurocontrol: A review and prospectus.
\newblock In {\em IEEE/INNS International Joint Conference on Neural Networks,
  Washington, D.C.}, volume~1, pages 209--216.

\bibitem[Werbos, 1989b]{Werbos:89identification}
Werbos, P.~J. (1989b).
\newblock Neural networks for control and system identification.
\newblock In {\em Proceedings of IEEE/CDC Tampa, Florida}.

\bibitem[Werbos, 1992]{Werbos:92sticky}
Werbos, P.~J. (1992).
\newblock Neural networks, system identification, and control in the chemical
  industries.
\newblock In D.~A.~White, D. A.~S., editor, {\em Handbook of Intelligent
  Control: Neural, Fuzzy, and Adaptive Approaches}, pages 283--356. Thomson
  Learning.

\bibitem[Werbos, 2006]{werbos2006backwards}
Werbos, P.~J. (2006).
\newblock Backwards differentiation in {AD} and neural nets: Past links and new
  opportunities.
\newblock In {\em Automatic Differentiation: Applications, Theory, and
  Implementations}, pages 15--34. Springer.

\bibitem[West and Saad, 1995]{westsaad:96}
West, A. H.~L. and Saad, D. (1995).
\newblock Adaptive back-propagation in on-line learning of multilayer networks.
\newblock In Touretzky, D.~S., Mozer, M., and Hasselmo, M.~E., editors, {\em
  NIPS}, pages 323--329. MIT Press.

\bibitem[White, 1989]{White:89}
White, H. (1989).
\newblock Learning in artificial neural networks: A statistical perspective.
\newblock {\em Neural Computation}, 1(4):425--464.

\bibitem[Whitehead, 1992]{Whitehead:92}
Whitehead, S. (1992).
\newblock {\em Reinforcement Learning for the adaptive control of perception
  and action}.
\newblock PhD thesis, University of Rochester.

\bibitem[Whiteson, 2012]{whiteson2012}
Whiteson, S. (2012).
\newblock Evolutionary computation for reinforcement learning.
\newblock In Wiering, M. and van Otterlo, M., editors, {\em Reinforcement
  Learning}, pages 325--355. Springer, Berlin, Germany.

\bibitem[Whiteson et~al., 2005]{stoneML05}
Whiteson, S., Kohl, N., Miikkulainen, R., and Stone, P. (2005).
\newblock Evolving keepaway soccer players through task decomposition.
\newblock {\em Machine Learning}, 59(1):5--30.

\bibitem[Whiteson and Stone, 2006]{whiteson2006}
Whiteson, S. and Stone, P. (2006).
\newblock Evolutionary function approximation for reinforcement learning.
\newblock {\em Journal of Machine Learning Research}, 7:877--917.

\bibitem[Widrow and Hoff, 1962]{adaline62}
Widrow, B. and Hoff, M. (1962).
\newblock Associative storage and retrieval of digital information in networks
  of adaptive neurons.
\newblock {\em Biological Prototypes and Synthetic Systems}, 1:160.

\bibitem[Widrow et~al., 1994]{widrow94}
Widrow, B., Rumelhart, D.~E., and Lehr, M.~A. (1994).
\newblock Neural networks: Applications in industry, business and science.
\newblock {\em Commun. ACM}, 37(3):93--105.

\bibitem[Wieland, 1991]{wieland1991}
Wieland, A.~P. (1991).
\newblock Evolving neural network controllers for unstable systems.
\newblock In {\em International Joint Conference on Neural Networks (IJCNN)},
  volume~2, pages 667--673. IEEE.

\bibitem[Wiering and Schmidhuber, 1996]{Wiering:96levin}
Wiering, M. and Schmidhuber, J. (1996).
\newblock Solving {POMDPs} with {L}evin search and {EIRA}.
\newblock In Saitta, L., editor, {\em Machine Learning: Proceedings of the
  Thirteenth International Conference}, pages 534--542. Morgan Kaufmann
  Publishers, San Francisco, CA.

\bibitem[Wiering and Schmidhuber, 1998a]{Wiering:97ab}
Wiering, M. and Schmidhuber, J. (1998a).
\newblock {HQ}-learning.
\newblock {\em Adaptive Behavior}, 6(2):219--246.

\bibitem[Wiering and van Otterlo, 2012]{wiering2012}
Wiering, M. and van Otterlo, M. (2012).
\newblock {\em Reinforcement Learning}.
\newblock Springer.

\bibitem[Wiering and Schmidhuber, 1998b]{Wiering:98}
Wiering, M.~A. and Schmidhuber, J. (1998b).
\newblock Fast online {Q}($\lambda$).
\newblock {\em Machine Learning}, 33(1):105--116.

\bibitem[Wierstra et~al., 2010]{wierstra2010}
Wierstra, D., Foerster, A., Peters, J., and Schmidhuber, J. (2010).
\newblock Recurrent policy gradients.
\newblock {\em Logic Journal of IGPL}, 18(2):620--634.

\bibitem[Wierstra et~al., 2008]{wierstraCEC08}
Wierstra, D., Schaul, T., Peters, J., and Schmidhuber, J. (2008).
\newblock Natural evolution strategies.
\newblock In {\em Congress of Evolutionary Computation (CEC 2008)}.

\bibitem[Wiesel and Hubel, 1959]{wiesel:1959}
Wiesel, D.~H. and Hubel, T.~N. (1959).
\newblock Receptive fields of single neurones in the cat's striate cortex.
\newblock {\em J. Physiol.}, 148:574--591.

\bibitem[Wiles and Elman, 1995]{wiles95learning}
Wiles, J. and Elman, J. (1995).
\newblock Learning to count without a counter: A case study of dynamics and
  activation landscapes in recurrent networks.
\newblock In {\em In Proceedings of the Seventeenth Annual Conference of the
  Cognitive Science Society}, pages pages 482 -- 487, Cambridge, MA. MIT Press.

\bibitem[Wilkinson, 1965]{Wilkinson:1965}
Wilkinson, J.~H., editor (1965).
\newblock {\em The Algebraic Eigenvalue Problem}.
\newblock Oxford University Press, Inc., New York, NY, USA.

\bibitem[Williams, 1986]{Williams:86}
Williams, R.~J. (1986).
\newblock Reinforcement-learning in connectionist networks: A mathematical
  analysis.
\newblock Technical Report 8605, Institute for Cognitive Science, University of
  California, San Diego.

\bibitem[Williams, 1988]{Williams:88}
Williams, R.~J. (1988).
\newblock Toward a theory of reinforcement-learning connectionist systems.
\newblock Technical Report NU-CCS-88-3, College of Comp. Sci., Northeastern
  University, Boston, MA.

\bibitem[Williams, 1989]{Williams:89}
Williams, R.~J. (1989).
\newblock Complexity of exact gradient computation algorithms for recurrent
  neural networks.
\newblock Technical Report Technical Report NU-CCS-89-27, Boston: Northeastern
  University, College of Computer Science.

\bibitem[Williams, 1992a]{Williams:92}
Williams, R.~J. (1992a).
\newblock Simple statistical gradient-following algorithms for connectionist
  reinforcement learning.
\newblock {\em Machine Learning}, 8:229--256.

\bibitem[Williams, 1992b]{williams1992kalman}
Williams, R.~J. (1992b).
\newblock Training recurrent networks using the extended {Kalman} filter.
\newblock In {\em International Joint Conference on Neural Networks (IJCNN)},
  volume~4, pages 241--246. IEEE.

\bibitem[Williams and Peng, 1990]{WilliamsPeng:90}
Williams, R.~J. and Peng, J. (1990).
\newblock An efficient gradient-based algorithm for on-line training of
  recurrent network trajectories.
\newblock {\em Neural Computation}, 4:491--501.

\bibitem[Williams and Zipser, 1988]{WilliamsZipser:88}
Williams, R.~J. and Zipser, D. (1988).
\newblock A learning algorithm for continually running fully recurrent
  networks.
\newblock Technical Report ICS Report 8805, Univ. of California, San Diego, La
  Jolla.

\bibitem[Williams and Zipser, 1989a]{WilliamsZipser:89cs}
Williams, R.~J. and Zipser, D. (1989a).
\newblock Experimental analysis of the real-time recurrent learning algorithm.
\newblock {\em Connection Science}, 1(1):87--111.

\bibitem[Williams and Zipser, 1989b]{WilliamsZipser:89nc}
Williams, R.~J. and Zipser, D. (1989b).
\newblock A learning algorithm for continually running fully recurrent
  networks.
\newblock {\em Neural Computation}, 1(2):270--280.

\bibitem[Willshaw and von~der Malsburg, 1976]{WillshawMalsburg:76}
Willshaw, D.~J. and von~der Malsburg, C. (1976).
\newblock How patterned neural connections can be set up by self-organization.
\newblock {\em Proc. R. Soc. London B}, 194:431--445.

\bibitem[Windisch, 2005]{windisch2005}
Windisch, D. (2005).
\newblock Loading deep networks is hard: The pyramidal case.
\newblock {\em Neural Computation}, 17(2):487--502.

\bibitem[Wiskott and Sejnowski, 2002]{WisSej2002}
Wiskott, L. and Sejnowski, T. (2002).
\newblock Slow feature analysis: Unsupervised learning of invariances.
\newblock {\em Neural Computation}, 14(4):715--770.

\bibitem[Witczak et~al., 2006]{witczak2006}
Witczak, M., Korbicz, J., Mrugalski, M., and Patton, R.~J. (2006).
\newblock A {GMDH} neural network-based approach to robust fault diagnosis:
  Application to the {DAMADICS} benchmark problem.
\newblock {\em Control Engineering Practice}, 14(6):671--683.

\bibitem[W\"{o}llmer et~al., 2011]{woellmer2011}
W\"{o}llmer, M., Blaschke, C., Schindl, T., Schuller, B., F\"{a}rber, B.,
  Mayer, S., and Trefflich, B. (2011).
\newblock On-line driver distraction detection using {Long Short-Term Memory}.
\newblock {\em IEEE Transactions on Intelligent Transportation Systems (TITS)},
  12(2):574--582.

\bibitem[W\"{o}llmer et~al., 2013]{woellmer2013}
W\"{o}llmer, M., Schuller, B., and Rigoll, G. (2013).
\newblock Keyword spotting exploiting {Long Short-Term Memory}.
\newblock {\em Speech Communication}, 55(2):252--265.

\bibitem[Wolpert, 1992]{wolpert:92stacked}
Wolpert, D.~H. (1992).
\newblock Stacked generalization.
\newblock {\em Neural Networks}, 5(2):241--259.

\bibitem[Wolpert, 1994]{Wolpert:94b}
Wolpert, D.~H. (1994).
\newblock Bayesian backpropagation over i-o functions rather than weights.
\newblock In Cowan, J.~D., Tesauro, G., and Alspector, J., editors, {\em
  Advances in Neural Information Processing Systems (NIPS) 6}, pages 200--207.
  Morgan Kaufmann.

\bibitem[Wu and Shao, 2014]{diwu2014}
Wu, D. and Shao, L. (2014).
\newblock Leveraging hierarchical parametric networks for skeletal joints based
  action segmentation and recognition.
\newblock In {\em Proc. Conference on Computer Vision and Pattern Recognition
  (CVPR)}.

\bibitem[Wu and Baldi, 2008]{wu2008go}
Wu, L. and Baldi, P. (2008).
\newblock Learning to play {Go} using recursive neural networks.
\newblock {\em Neural Networks}, 21(9):1392--1400.

\bibitem[Wyatte et~al., 2012]{wyatte2012b}
Wyatte, D., Curran, T., and O'Reilly, R. (2012).
\newblock The limits of feedforward vision: Recurrent processing promotes
  robust object recognition when objects are degraded.
\newblock {\em Journal of Cognitive Neuroscience}, 24(11):2248--2261.

\bibitem[Wysoski et~al., 2010]{wysoski2010}
Wysoski, S.~G., Benuskova, L., and Kasabov, N. (2010).
\newblock Evolving spiking neural networks for audiovisual information
  processing.
\newblock {\em Neural Networks}, 23(7):819--835.

\bibitem[Yamauchi and Beer, 1994]{yamauchi94sequential}
Yamauchi, B.~M. and Beer, R.~D. (1994).
\newblock Sequential behavior and learning in evolved dynamical neural
  networks.
\newblock {\em Adaptive Behavior}, 2(3):219--246.

\bibitem[Yamins et~al., 2013]{Yamins2013}
Yamins, D., Hong, H., Cadieu, C., and DiCarlo, J.~J. (2013).
\newblock Hierarchical modular optimization of convolutional networks achieves
  representations similar to macaque {IT} and human ventral stream.
\newblock {\em Advances in Neural Information Processing Systems (NIPS)}, pages
  1--9.

\bibitem[Yang et~al., 2009]{trecvid2009}
Yang, M., Ji, S., Xu, W., Wang, J., Lv, F., Yu, K., Gong, Y., Dikmen, M., Lin,
  D.~J., and Huang, T.~S. (2009).
\newblock Detecting human actions in surveillance videos.
\newblock In {\em TREC Video Retrieval Evaluation Workshop}.

\bibitem[Yao, 1993]{yao:review93}
Yao, X. (1993).
\newblock A review of evolutionary artificial neural networks.
\newblock {\em International Journal of Intelligent Systems}, 4:203--222.

\bibitem[Yin et~al., 2013]{icdar2013}
Yin, F., Wang, Q.-F., Zhang, X.-Y., and Liu, C.-L. (2013).
\newblock {ICDAR} 2013 {Chinese} handwriting recognition competition.
\newblock In {\em 12th International Conference on Document Analysis and
  Recognition (ICDAR)}, pages 1464--1470.

\bibitem[Yin et~al., 2012]{yin2012}
Yin, J., Meng, Y., and Jin, Y. (2012).
\newblock A developmental approach to structural self-organization in reservoir
  computing.
\newblock {\em IEEE Transactions on Autonomous Mental Development},
  4(4):273--289.

\bibitem[Young et~al., 2014]{young2014}
Young, S., Davis, A., Mishtal, A., and Arel, I. (2014).
\newblock Hierarchical spatiotemporal feature extraction using recurrent online
  clustering.
\newblock {\em Pattern Recognition Letters}, 37:115--123.

\bibitem[Yu et~al., 1995]{Yu:1995}
Yu, X.-H., Chen, G.-A., and Cheng, S.-X. (1995).
\newblock Dynamic learning rate optimization of the backpropagation algorithm.
\newblock {\em IEEE Transactions on Neural Networks}, 6(3):669--677.

\bibitem[Zamora-Martínez et~al., 2014]{zamora2014}
Zamora-Martínez, F., Frinken, V., España-Boquera, S., Castro-Bleda, M.,
  Fischer, A., and Bunke, H. (2014).
\newblock Neural network language models for off-line handwriting recognition.
\newblock {\em Pattern Recognition}, 47(4):1642--1652.

\bibitem[Zeiler, 2012]{zeiler2012}
Zeiler, M.~D. (2012).
\newblock {ADADELTA: An Adaptive Learning Rate Method}.
\newblock {\em CoRR}, abs/1212.5701.

\bibitem[Zeiler and Fergus, 2013]{zeiler2013}
Zeiler, M.~D. and Fergus, R. (2013).
\newblock Visualizing and understanding convolutional networks.
\newblock Technical Report arXiv:1311.2901 [cs.CV], NYU.

\bibitem[Zemel, 1993]{Zemel:93}
Zemel, R.~S. (1993).
\newblock {\em A minimum description length framework for unsupervised
  learning}.
\newblock PhD thesis, University of Toronto.

\bibitem[Zemel and Hinton, 1994]{Zemel:94nips}
Zemel, R.~S. and Hinton, G.~E. (1994).
\newblock Developing population codes by minimizing description length.
\newblock In Cowan, J.~D., Tesauro, G., and Alspector, J., editors, {\em
  Advances in Neural Information Processing Systems 6}, pages 11--18. Morgan
  Kaufmann.

\bibitem[Zeng et~al., 1994]{zeng94discrete}
Zeng, Z., Goodman, R., and Smyth, P. (1994).
\newblock Discrete recurrent neural networks for grammatical inference.
\newblock {\em IEEE Transactions on Neural Networks}, 5(2).

\bibitem[Zimmermann et~al., 2012]{DBLP:series/lncs/ZimmermannTG12}
Zimmermann, H.-G., Tietz, C., and Grothmann, R. (2012).
\newblock Forecasting with recurrent neural networks: 12 tricks.
\newblock In Montavon, G., Orr, G.~B., and M{\"u}ller, K.-R., editors, {\em
  Neural Networks: Tricks of the Trade (2nd ed.)}, volume 7700 of {\em Lecture
  Notes in Computer Science}, pages 687--707. Springer.

\bibitem[Zipser et~al., 1993]{zipser1993}
Zipser, D., Kehoe, B., Littlewort, G., and Fuster, J. (1993).
\newblock A spiking network model of short-term active memory.
\newblock {\em The Journal of Neuroscience}, 13(8):3406--3420.

\end{thebibliography}
\bibliographystyle{apalike}
\end{document}